# Scalable Bayesian Optimization with Sparse Gaussian Process Models

by

**Ang Yang**

**Submitted in fulfillment of the requirements for the degree of Doctor of Philosophy**

**Deakin University**

**April 2020**

# DEAKIN UNIVERSITY
# ACCESS TO THESIS - A

I am the author of the thesis entitled **Scalable Bayesian Optimization with Sparse Gaussian Process Models,** submitted for the degree of **Doctor of Philosophy**

This thesis may be made available for consultation, loan and limited copying in accordance with the Copyright Act 1968.

*'I certify that I am the student named below and that the information provided in the form is correct'*

**Full Name: Ang Yang**

**Signed:**

**Date: 16/04/2020**

# DEAKIN UNIVERSITY
# CANDIDATE DECLARATION

I certify the following about the thesis entitled (10 word maximum)

**Scalable Bayesian Optimization with Sparse Gaussian Process Models**

submitted for the degree of **Doctor of Philosophy**

a. I am the creator of all or part of the whole work(s) (including content and layout) and that where reference is made to the work of others, due acknowledgment is given.

b. The work(s) are not in any way a violation or infringement of any copyright, trademark, patent, or other rights whatsoever of any person.

c. That if the work(s) have been commissioned, sponsored or supported by any organisation, I have fulfilled all of the obligations required by such contract or agreement.

d. That any material in the thesis which has been accepted for a degree or diploma by any university or institution is identified in the text.

e. All research integrity requirements have been complied with.

*'I certify that I am the student named below and that the information provided in the form is correct'*

**Full Name: Ang Yang**

**Signed:**

**Date: 16/04/2020**

This thesis is dedicated to my wife
Mei Liu and our parents
for their unfailing love and support.

# Preface

Four years ago, before I started my Ph.D. study journey, I used to love baking cookies. There are six ingredients to make my favorite cookie, and I have experienced that cookies taste different if I use different portions of each ingredient. Besides, the choice of oven temperature and time of baking are crucial. At that time, I had a question in my mind: How to make the best cookies in the world? The most straightforward way is to try every possible combination but my whole life will end up with making cookies, which is not something I can afford.

However, I was lucky. It was not far from I got this hard question in my mind that I had a chance to join my current research group, which is working on optimization problems, more specifically, Bayesian optimization. After I did the literature reviews on this optimization strategy I realized that the Bayesian optimization would be the perfect method to answer my question: how to make the best cookies in the world. It operates by expressing a belief over all possible objective functions through a prior. As data is observed, the likelihood of these observations is combined with the prior knowledge to updating the posterior distribution. Then the posterior is used to determine what is the next point for sampling. The algorithm can filter out a huge non-important search spaces and only take samples from important regions. Moreover, it is a global optimization method i.e. it is also guaranteed to jump out from a local minima in the pursuit of a better local minima elsewhere in the space. Therefore, I may only need to try few 10s or 100s of times, instead of millions of possible combinations to find the recipe for the best cookies.

There are more complicated problems that Bayesian optimization can help, for example, the production of superalloys. The superalloys could be used in manufactur-ing vehicles such as the Tesla Roadster. As a result, the vehicles can be lighter, faster and can go farther on a single charge. However, manufacturing superalloys is not as simple as baking cookies. The production processes have a large number of alloying elements and many other processing parameters that need to consider. Moreover, the cost of one experiment is very expensive, in terms of both time and materials. Therefore, the standard Bayesian optimization falls short in such scenarios.

My story with Bayesian optimization starts from here, a journey to advance the state of the art of Bayesian optimization.

# Abstract

Bayesian optimization forms a set of powerful tools that allows efficient black-box optimization and has been applied in a large variety of fields. In this thesis we first seek to advance Bayesian optimization by using estimated derivative observations. Later, we seek to tackle down the issues in Bayesian optimization when a large number of derivative observations and/or function observations are present.

We start to describe our motivations in Chapter 1. We then give a broad review of Bayesian optimization in Chapter 2, where we start by covering the history of Bayesian optimization and its components. We also discuss open questions, espe-cially when a large set of observations are present. We end the chapter with a review on current open-source libraries for Bayesian optimization. In Chapter 3, we introduce the mathematical background for this thesis in depth.

Although Bayesian optimization is an efficient method, the current Gaussian process based approaches cater to functions with arbitrary smoothness, and do not explicitly model the fact that most of the real world optimization problems are well-behaved functions with only a few peaks. In Chapter 4, we incorporate such shape constraint through the use of a derivative meta-model. The derivative meta-model is built using a Gaussian process with a polynomial kernel and derivative samples extract from this meta-model are used as extra observations, which are then adapted to the standard Bayesian optimization procedure. We also provide a Bayesian framework to infer the degree of the polynomial kernel. Experiments on both benchmark functions and hyperparameter tuning problems demonstrate the efficiency of our approach over baselines.

However, a practical limitation of the Gaussian process is that its computational complexity increases cubically with the size of the training set. In Chapter 5, we therefore propose a sparse Gaussian process model to approximate full Gaussian process when both a large number of function observations and derivative obser-vations are available. By introducing a small number of inducing points to the full Gaussian process, the computational complexity can be reduced significantly. Experiments demonstrate the efficiency of our approach on both regression and Bayesian optimization tasks.

In Chapter 6, we propose a novel sparse spectrum approximation of Gaussian pro-cess tailored for Bayesian optimization. Whilst the current sparse spectrum methods provide good approximations for regression problems, it is observed that this partic-ular form of sparse approximations generates an overconfident Gaussian process, i.e. it produces less epistemic uncertainty than a full-scale Gaussian process would have. Since the balance between the predictive mean and variance is the key determinant to the success of Bayesian optimization, the existing sparse spectrum methods are less suitable for Bayesian optimization. To fix this overconfidence issue, particularly for Bayesian optimization, we derive a new regularized marginal likelihood for find-ing the optimal frequencies. The regularizer trades off the accuracy in the model fitting with the predictive variance of the resultant Gaussian process. Specifically, we use the entropy of the global maximum distribution from the posterior Gaussian process as the regularizer that needs to be maximized. Since the the global max-imum distribution cannot be calculated analytically, we first propose a Thompson sampling based approach and then a more efficient sequential Monte Carlo based approach to estimate it. Later, we also show that the expected improvement ac-quisition function can be used as a proxy for it, thus making the calculation further efficient. Experiments show a considerable improvement to Bayesian optimization convergence rate over the vanilla sparse spectrum methods.

# Acknowledgements

It would not have been possible to write this doctoral thesis without the help and support of the kind people around me, to only some of whom it is possible to give a particular mention here.

Above all, I would like to express my deep and sincere gratitude to my research supervisor, Dr. Cheng Li, Prof. Santu Rana, Prof. Sunil Gupta, and Prof. Svetha Venkatesh. Thank you for allowing me to do research and providing invaluable guid-ance throughout this research and it was my great privilege and honor to work and study under their guidance. I could not have imagined having better advisors and mentors for my Ph.D. study. I would also grateful for colleagues who have presented related works during seminars, meetings and have shared ideas and opinions during lunch discussions.

This research was funded by the Australian Government through the Australian Laureate Fellowship and Deakin University. I would like to acknowledge their fin-ancial support to make this happen.

Last but not the least, I would like to thank my wife Merry for sacrificing her career in the US and coming to Australia to offer her support and great patience at all times. Our parents and friends also have given me their unequivocal support throughout, as always, for which my mere expression of thanks likewise does not suffice.

# Relevant Publications

Part of this thesis has been published as manuscripts as detailed below:

**Chapter 4:**

- **Yang. A.**, Li, C., Rana, S., Gupta, S. and Venkatesh, S., 2018, August. Efficient Bayesian optimisation using derivative meta-model. In Pacific Rim International Conference on Artificial Intelligence (pp. 256-264). Springer, Cham [Yang *et al.*2018a].

**Chapter 5:**

- **Yang. A.**, Li, C., Rana, S., Gupta, S. and Venkatesh, S., 2018, December. Sparse Approximation for Gaussian Process with Derivative Observations. In Australasian Joint Conference on Artificial Intelligence (pp. 507-518). Springer, Cham [Yang *et al.*2018b].

**Chapter 6:**

- **Yang. A.**, Li, C., Rana, S., Gupta, S. and Venkatesh, S., 2019. Sparse Spec-trum Gaussian Process for Bayesian Optimisation. arXiv preprint arXiv:1906.-08898 [Yang *et al.*2019]. This work has been submitted to IJCAI-2020 and is currently under full review.

# List of Symbols

| Symbols | Description |
|---|---|
| $f(\cdot)$ | Objective function |
| $\mathbb{R}^d$ | $d$ dimensional real vector space |
| $\mathcal{X}$ | A compact subspace of $\mathbb{R}^d$ |
| $\boldsymbol{x}$ | A vector from $\mathbb{R}^d$ real vector space |
| $\mathcal{N}$ | A normal distribution |
| $\xi$ | Noise distribution $\xi \sim \mathcal{N}(\mathbf{0}, \sigma_{noise}^2)$ |
| $\boldsymbol{y}$ | Observations vector with noise $\boldsymbol{y} \sim \mathcal{N}(f(\boldsymbol{x}), \sigma_{noise}^2)$ |
| $\{\boldsymbol{x}, y\}$ | Observation pair |
| $d$ | Dimension of the observations |
| $k(\cdot)$ | Kernel function |
| $\mu(\cdot)$ | Mean function |
| $\alpha_{EI}(\boldsymbol{x})$ | EI acquisition function |
| r | Gradient operator |
| $\mathcal{D}_t$ | A set of $t$ function observations |
| $d_i$ | $i_{th}$ dimension |
| $t$ | Number of observations |
| $t^0$ | Number of derivative observations |
| $\boldsymbol{X}_{1:t}$ | The sequence $\{\boldsymbol{x}_1 \ldots \boldsymbol{x}_t\}$ |
| $\boldsymbol{x}_{t+1}$ | Next point to evaluate |
| $\boldsymbol{f}_{1:t}$ | The sequence $\{f(\boldsymbol{x}_1) \ldots f(\boldsymbol{x}_t)\}$ |
| $\boldsymbol{x}^+$ | Current maximum |
| $\boldsymbol{x}^*$ | Global maximum |
| $p(\boldsymbol{x}^*)$ | Distribution of the global maximum |
| $q(\boldsymbol{x}^*)$ | Approximated distribution of the global maximum |
| $r_t$ | Simple regret |
| $\mathcal{O}$ | Time complexity |
| $m$ | Number of inducing points |
| $\mathcal{D}_m$ | A set of $m$ inducing points |
| $\boldsymbol{X}_u$ | Inducing inputs (Time domain) |

| Symbols | Description |
|---|---|
| $\mathbf{u}$ | Function values at $\boldsymbol{X}_{\mathrm{u}}$ |
| $\{\boldsymbol{X}_{\mathrm{u}}, \mathbf{u}\}$ | Inducing observation pairs |
| $\rho_l$ | Isotropic lengthscale of the SE kernel |
| $\sigma_f$ | Kernel variance |
| $\boldsymbol{s}_r$ | $r^{th}$ spectral frequencies |
| $\boldsymbol{\Theta}$ | All the model parameters |
| $D_{KL}(\cdot)$ | KL divergence |
| $\mathrm{H}(\cdot)$ | Entropy of a distribution |
| $\varphi(\boldsymbol{x})$ | A set of random Fourier basis |
| $\underline{\varphi}_i$ | A random Fourier basis |
| $\boldsymbol{\theta}_i$ | Corresponding posterior weight of the random Fourier basis |
| $n_p$ | Number of particles used in Particle filtering |
| $n_c$ | Number of challenger particles |
| $\boldsymbol{x}^{-i}$ | $i^{th}$ particle |
| $\boldsymbol{X}_{c_i}$ | $i^{th}$ challenger particle |
| $\lambda$ | The trade-off parameter for regularizer |

# Abbreviations

| Abbreviation | Description |
|---|---|
| BO | Bayesian optimization |
| GP | Gaussian process |
| EI | Expected improvement |
| PI | Probability of improvement |
| GP-UCB | Gaussian process upper confidence bound |
| CDF | Cumulative distribution function |
| PDF | Probability density function |
| SE | Squared exponential kernel |
| GPPK | Gaussian process with polynomial kernel |
| GPD | Gaussian process with derivatives |
| SGP | Sparse Gaussian process |
| FITC | Fully independent training conditional |
| FIC | Fully independent conditional |
| SGPD | Sparse Gaussian process with derivatives |
| SSGP | Sparse spectrum Gaussian process |
| VFF | Variational Fourier features |
| GMD | Global maximum distribution |
| TS | Thompson sampling |
| MC | Monte Carlo |
| RSSGP | Regularized sparse spectrum Gaussian process |
| BODMM | Bayesian Optimization using Derivative Meta-model |
| BOSGPD | Bayesian optimization using sparse Gaussian process with derivatives |

# Contents

# List of Figures

# List of Algorithms

# Chapter 1

# Introduction

Experiments are always conducted to investigate the behavior of a system under different settings, or to simply reach a target behaviour. The design of experiments determines the allocation of resources and methods of experimentation to satisfy such objectives [Jacquez1998]. Sequential design of experiments is useful to when we want to reach a target outcome. To achieve the target, we will perform a sequence of experiments, and utilize the knowledge from the experiments completed so far at each time. We expect to reach a specific objective by the shortest number of exper-iments. Further, in the sequential design of experiments, the decision of progress of the experiments is evaluated sequentially from the results [Chernoff1959, Box and Hunter1963] and we try to find a good set of the facts that maximizes the result of the experiments within the budget constraints [Chernoff1973]. The budget con-straints can be wall clock time, cost of the experiment, etc. Such techniques can be used to make a product of specified property via finding the proper composition of the raw materials.

For example, in superalloy design process, the metallurgists aim to design an alloy with a micro-structure that presents as much strength as possible. The search space is usually a 10 to 20 dimensional combination of the elements. For each composition, the alloy presents the level of strength respectively. It is impossible to deal with such an overwhelming number of composition manually.

Bayesian optimization (BO) using Gaussian process (GP) is one of the most popular sequential designs method for global optimization for expensive black-box functions. It has been successfully applied to problems in materials science, robotics, syn-thetic gene design and recommendation systems, etc [Garnett *et al.*2010, González *et al.*2015, Li *et al.*2017a]. Recently, Bayesian optimization has also found popularity in tuning hyperparameters for machine learning algorithms [Bergstra *et al.*2011, Joy *et al.*2019]. Bayesian optimization has two main components, the first one is to model the unknown function and the other one is to select the next point where to perform the experiment based on this model. It operates by expressing a belief over all possible objective functions through a prior. As data is observed, the posterior distribution is computed by combing the likelihood of these observations and GP prior. Then a utility function which combines the mean and variance of posterior GP is used to determine the next point for evaluating the black-box function. The utility function is called acquisition function and it is cheap to evaluate. The role of the acquisition function is to balance the need to sample next to the already seen good values of the objective function (*exploitation*) and the need to *explore* away from the already known region.While the former is required to reach the peak of a local optima, the latter is required to move away from a local optima towards the global optima. The next point to evaluate is then obtained by maximizing the acquisition function.

Bayesian optimization has sublinear rate of growth in cumulative regret with number of samples. This convergence guarantee holds for any arbitrary smooth functions with arbitrarily large number of local peaks. However, most real world functions, resulting either from physical experiments or hyperparameter tuning, are well be-haved. They are smooth and have a small number of local peaks. If such knowledge can be harnessed, then BO can converge faster. Therefore, in this thesis we study this problem as the first step to enter the space, which is later followed by a deep research in large-scale Bayesian optimization since in many instances incorporating extra knowledge in GP is achieved by adding extra observations. Next, we detail our motivations and provide a sketch of the approaches.

## 1.1 Motivations and approaches

### 1.1.1 Derivative meta-model for Bayesian optimization

Previously, several researchers looked at introducing derivative observations in Gaus-sian Process [Solak *et al.*2003, Jauch and Peña2016] and incorporating derivative information in Bayesian optimization [Siivola *et al.*2017, Wu *et al.*2017]. However, in reality, since we are optimizing a black box function, we do not have the access to the derivative information of the latent function. Driven by this we have identified a promising research direction where derivative can be obtained from a subsidiary meta-model of a known form. As a result, it is possible to sample derivative in-formation from this meta-model. We can then combine the actual observations and this syntactic derivative information to build a GP and this GP can then be used to perform Bayesian optimization. The usefulness of such a derivative meta-model is realized in Chapter 4 by building the meta-model enforcing complex shape in-formation. The derivative meta-model is built using a polynomial function. To maintain the Bayesian flavor and the ability of estimating the meta-model from a few observations, we use a Gaussian process with polynomial kernel (GPPK) for the meta-model. Based on the observed data we fit the GPPK and then sample derivative values for the use in the main GP for the BO. In effect, the main GP is built based on a trade-off between the flexible model induced by the stationery kernel and the structure induced by the derivative information based on GPPK. We refrain from using the samples of the function values from GPPK because we only want to pass the shape information through derivatives, while keeping the function values guided mostly by the main GP. A crucial question in this scheme is how to set the degree of the polynomial kernel. We use a Bayesian formulation to estimate the degree from the observed data. We then use a truncated geometric prior, cut-off at degree of 10 as the prior information. The posterior is then computed based on the marginal likelihood of the GPPK on the observed data. The mode of the posterior is then used as the degree for our derivative meta-model.

### 1.1.2 Sparse Gaussian process for function and derivative observations

Although incorporating derivative observations can empirically accelerate the convergence rate of Bayesian optimization, it also introduces computational challenges. In higher dimension, derivatives is a vector and each entry is a separate observation for GP. So the number of observations also scale with the dimension. Posterior computation of GP entails matrix inversion, which is cubic to the size of the mat-rix. Therefore, it becomes very difficult to apply GP with derivative observations to large-scale problems. The focus of Chapter 5 is on deriving an efficient sparse GP model to approximate the full GP with derivative observations. We use a set of $m$ inducing variables, where $m \leq (t + dt^0)$ and $t$ is the number of function observa-tions, $t^0$ is the number of derivative observations and $d$ is the dimensionality of the space. We first assume that the function observations and derivative observations are conditionally independent given the set of inducing variables. Then the condi-tional distribution for a test point can be obtained by integrating out the inducing variables. To further decrease the costly computation, we introduce the fully inde-pendence that function observations, derivative observations also test points have no any deterministic relation on the inducing variables so we can ignore the covariance between them. The resultant sparse GP model provides a cheaper approximation to the full GP model as the computation for the sparse model depends primarily on the number of inducing points instead of the total number of observations. The sparse Gaussian process although useful for Bayesian optimization, is still not espe-cially tailored for Bayesian optimization, resulting in less than expected behaviour in the experiments. Therefore, in the next we focus on sparse Gaussian process for large-scale Bayesian optimization.

### 1.1.3 Scalable Bayesian optimization

The scalability issue for Bayesian optimization has been previously addressed in two main ways: 1) by replacing GP with a more scalable Bayesian model, *e.g.* using Bayesian neural network [Snoek *et al.*2015] or random forest [Hutter *et al.*2011a], or 2) by making sparse approximation of the full GP. The latter is often desirable as it still maintains the principled Bayesian formalism of GP. There are many sparse

GP models in the literature, however, all of them including the one presented in Chapter 4, suffer from either variance underestimation (i.e., overconfidence) [Snelson and Ghahramani2006, Lazaro Gredilla *et al.*2010] or overestimation [Titsias2009] and thus may hamper BO as the balance between predictive mean and variance is important to the success of BO.

Therefore, in Chapter 6, we aim to develop a sparse GP model tailored for Bayesian optimization. We propose a novel modification to the sparse spectrum Gaussian process approach to make it more suitable for BO tasks by doing targeted fixing of variance misestimation. The main intuition that drives our solution is that while being overconfident at some regions is not very critical to BO when those regions have both low predictive value and low predictive variance. However, being overconfident in the regions where either predictive mean or predictive variance is high would be quite detrimental to BO. Hence, a targeted fixing may be enough to make the sparse models suitable for BO. An overall measure of goodness of GP approximation for BO would be to look at the global maximum distribution (GMD) [Hennig and Schuler2012, Hernández *et al.*2014] from the posterior GP and check its difference to that of the full GP. Fixing overconfidence in the important regions may be enough to make the GMD of the sparse GP closer to that of the full GP. The base method in our work (SSGP) is known to underestimate variance, which is why we need maximizing the entropy of GMD. Following this idea, we add the entropy of the GMD as a new regularizer that is to be maximized in conjunction with the marginal likelihood so the optimal sparse set of the frequencies not only benefits for model fitting, but also fixes the overconfidence issue from the perspectives of the Bayesian optimization.

## 1.2 Outline of the thesis

The rest of the thesis is organized as follows.

In Chapter 2, we present a board review of Bayesian optimization. We also identify open questions in Bayesian optimization and discuss the related works that under-pins the research in this thesis.

In Chapter 3, we provide a detailed mathematical background on Bayesian optim-ization, which includes descriptions on Gaussian process and various acquisition functions. We also provide details on Gaussian process with gradients and two different sparse Gaussian process models.

In Chapter 4, we propose a novel method for Bayesian optimization for well-behaved functions with small numbers of peaks. We incorporate this information through a derivative meta-model. The derivative meta-model is based on a Gaussian process with a polynomial kernel. By controlling the degree of the polynomial we control the shape of the main Gaussian process which is built using the SE kernel and the covariance matrix is computed by using both the observed function value and the derivative values sampled from the meta-model. We also provide a Bayesian way to estimate the degree of the polynomial based on a truncated geometric prior. We demonstrate the efficiency of the proposed method on the benchmark test functions and the hyperparameter tuning of popular machine learning models.

In Chapter 5, we propose an extended framework of FIC [Quinonero-Candela and Rasmussen2005] based sparse approximation to incorporate derivative observations. The novel framework can significantly speed up the Gaussian process with derivat-ives from $O((t + dt^0)^3)$ to $O((t + dt^0)m^2)$. We first analyse our method on synthetic regression tasks. We then investigate the usability of our method in large-scale Bayesian optimization for benchmark functions and tuning hyperparmenters of two machine learning algorithms. In all experiments, our proposed approach closely approximates the full Gaussian process with derivative observations.

In Chapter 6, we propose a regularized sparse spectrum Gaussian process method to make it more suitable for Bayesian optimization applications. The original formulation results in an over-confident Gaussian process. Therefore, we propose a modification to the original marginal likelihood based estimation by adding the entropy of the global maximum distribution induced by the posterior Gaussian process as a regularizer. By maximizing the entropy of that distribution along with the marginal likelihood, we aim to obtain a sparse approximation that is more aligned with the goal of Bayesian optimization. We show that an efficient formulation can be obtained by using a sequential Monte Carlo approach to approximate the global maximum distribution. We also experiment with the expected improvement acquisition function as a proxy to the global maximum distribution. We evaluate the model on two benchmark functions and two real-world problems. The results demonstrate the superiority of our approach over the naive sparse spectrum Gaussian process method at all times and even better than the usual full Gaussian process based approach at certain scenarios.

In Chapter 7, we conclude our contributions and discuss the possible future devel-opments that align with this thesis.

# Chapter 2

# Literature Review

In this chapter, we provide a detailed background on components of Bayesian opti-mization and the related works on this thesis. We start to introduce the history of Bayesian optimization and discuss the open questions of this research field in Sec-tion 2.1. Next, we review the existing methods for scalable Bayesian optimization in Section 2.2. Last, we provide a review of current open-source libraries for Bayesian optimization in Section 2.3.

## 2.1 Bayesian optimization

Bayesian optimization (BO) is a leading sequential design strategy for global optimization of expensive black-box functions [Jones *et al.*1998, Brochu *et al.*2010, Shahriari *et al.*2015]. It is widely used in hyperparameter tuning of machine learning algorithms such as support vector machines [Snoek *et al.*2012, Gardner *et al.*2014, Joy *et al.*2016], massive neural networks [Snoek *et al.*2012, Springenberg *et al.*2016] etc., for which training time can be high. It has also been used for optimization of physical products and processes [Li *et al.*2017a, Nguyen *et al.*2016] where one ex-periment can take days, and experiments can also be expensive in terms of material cost and personnel time.

Briefly, Bayesian optimization includes two main components. It first employs a probabilistic model, typically a Gaussian process (GP), to model the latent function and then constructs an acquisition function that determines where to sample next. A GP is an extension of the multivariate Gaussian distribution to an infinite dimensional stochastic process for which any finite combination of dimensions will be a Gaussian distribution. The main advantage of GP modeling is that its posterior can be computed in closed-form. The role of the acquisition function is to guide us to reach the optimum of the underlying function by carefully trading off exploitation and exploration based on GP posterior mean and variance, respectively.

### 2.1.1 History of Bayesian optimization and its components

Studies in the Bayesian optimization dates back to the work mentioned in [Kushner1964] which used Wiener process for unconstrained one-dimensional problems. This work has a similar flavor of the modern Bayesian optimization, since both of them share the concept of using a parameter that controls the trade off between exploration and exploitation. Later, the works by [Stuckman1988, Elder1992] extended the Wiener process to multidimensional cases by projecting Wiener process between sample points and interpolating in the Delauney triangulation of the space. Meanwhile, a multidimensional Bayesian optimization method using linear combinations of Wiener fields was first proposed in [Močkus1975]. It describes the notion of using an acquisition function that is based on expected improvement of the posterior. This acquisition function, later, evolves to the well-known Expected Improvement (EI) acquisition function, as illustrated in [Mockus1994].

More recently, Bayesian optimization has been known widely as efficient global op-timization in the experimental design literature due to the work of [Schonlau *et al.*1998]. At the same time, Gaussian process has been successfully applied to Bayesian optimization to model the objective function [Storn and Price1997] and this technique is also referred to as Gaussian process bandits [Srinivas *et al.*2010]. More-over, the theoretical convergence of the Bayesian optimization algorithm using mul-tivariate Gaussian process has been well established in [Srinivas *et al.*2010, Vazquez and Bect2010, Bull2011].

Today, Bayesian optimization using Gaussian process has become one of the most popular and efficient framework for derivative free global optimization [Osborne *et al.*2009, Brochu *et al.*2010, Snoek *et al.*2012, Marchant and Ramos2012, Souza *et al.*2014, González *et al.*2015].

#### 2.1.1.1 Gaussian process and other surrogate models

The operation of Bayesian optimization requires expressing a belief over all possible objective functions as a prior. The Gaussian process [Rasmussen and Williams2006] is the most popular choice of the prior from the very early history of Bayesian op-timization [Mockus1994, O'Hagan and Kingman1978, Žilinskas1980], since its pos-terior can be computed in closed-form.

Without loss of generality a GP can be specified by a prior mean of zero function and a covariance function. The choice of covariance function determines what kind of functions (e.g. smoothness, periodicity etc.) have a higher prior probability. There are several covariance functions [Rasmussen and Williams2006] in the liter-ature and Squared Exponential (SE) kernel has been a popular choice [Brochu *et al.*2010, Rasmussen and Williams2006]. The SE holds an assumption that the func-tion being modeled is infinitely differentiable. Another popular covariance function is the Mat´ern 5/2 kernel [Matérn2013]. In contrast to the SE, Mat´ern 5/2 ker-nel induces a high prior probability to rougher functions, whilst still ensuring that derivatives up to fourth order still exist. These kernels have parameters in their formulations, for example, the SE has hyperparameters such as length-scale, which determines the smoothness of the function and kernel variance which determines the magnitudes of the function. The hyperparameters can be learned directly from the training data by maximizing the the likelihood of the evidence [Rasmussen and Williams2006]. Alternatively, one can place a hyper-prior on these hyperparame-ters and learn the hyperparameters through a Bayesian paradigm (i.e. the posterior of hyperparameters) [Frean and Boyle2008, Lizotte2008]. More recently, one work [Snoek *et al.*2012] demonstrates that by using slice sampling, one can compute the integrated acquisition function without using the maximum a posterior (MAP) es-timation of the hyperparameters. Moreover, a linear combination of many kernels can also be used for more complex problems [Gönen and Alpaydın2011].

Recently, random forest has been proposed as an alternate model to Gaussian process in the context of sequential model based algorithm configuration (SMAC) [Hutter *et al.*2011b, Criminisi *et al.*2012]. Random forest is an ensemble method which uses de-cision trees learned on the random samples of data as weak learners [Breiman2001]. The prediction is then computed based on the average response from these trees. Random forest proposes a heuristic based method to estimate the variance, result-ing in some drawbacks in practice. For example, random forest fails miserably in the prediction of the points far from the training data. Moreover, when random forest is used in conjunction with Bayesian optimization, the resulting acquisition function becomes non-differentiable, making it unsuitable to apply gradient-based optimization methods.

More recently, a Bayesian version of feed-forward neural network has been used to specify a distribution over the objective functions [Snoek *et al.*2015]. It achieves the goal by performing adaptive basis function regression with a neural network as the parametric form [Snoek *et al.*2015]. Similarly, multi-layer neural networks have also been used as posterior models to easily scale to large-scale data points.[Schilling *et al.*2015, Springenberg *et al.*2016]. Moreover, neural network has also been used when embedding is employed to optimize over non-vector data e.g. graph [Zhang *et al.*2019b] where one uses BO in the latent space of variational autoencoder. However, all neural network based methods suffer from similar limitations as Random Forest in terms of posterior in-consistency, and lack of differentiability.

#### 2.1.1.2 Acquisition functions

Acquisition functions are the utility functions which trades off between exploration and exploitation of promising values. There are different criteria to define acqui-sition functions. A simple criteria is the Probability of Improvement (PI) which measures the probability of a location leading to improvement over the current best location [Parzen1962]. However, the key limitation with PI is that it can at times be extremely greedy, resulting in the aggressive exploitation [Jones2001]. For exam-ple, if the search falls into a particularly low basin, then exploring an infinitesimal distance from the current best point can yield a high PI. Therefore, the search will simply continue to explore near the current best location rather than exploring ar-eas with higher uncertainty. Alternatively, one can measure Expected Improvement (EI) to remedy such behavior up to certrain extent. The EI was first discussed in [Močkus1975], which incorporates the magnitude of improvement to the utility func-tion. EI has been empirically demonstrated to be an effective acquisition function [Jones *et al.*1998, Lizotte2008, Snoek *et al.*2012]. The extension, which has addressed the parallelization issues and greatly improves the success of EI, has been derived in [Ginsbourger *et al.*2010] with the rate of convergence discussed in [Bull2011].

Gaussian process Upper Confidence Bound (GP-UCB) has been proposed as an ac-quisition function which offers sub-linear growth rate for cumulative regret [Srinivas *et al.*2010]. A more recent study in [De Freitas *et al.*2012] proves that by applying a branch and bound algorithm on GP-UCB one can receive an exponentially van-ishing regret in the noiseless cases. The idea of the branch and bound algorithm is to sample the search space with progressively higher density at targeted regions and shrink this space as much as that the Upper Confidence Bound is lower than the Lower Confidence Bound (LCB). Later, the work of [Marchant and Ramos2012] uses a modified GP-UCB acquisition function that incorporates knowledge about the previously sampled locations to a specific application of environmental monitoring, resulting in a tighter regret bound for this use case.

Unlike the acquisition functions above, there is a separate set of acquisition functions which operate based on the measurement of information about the global optimum. Entropy Search (ES) [Villemonteix *et al.*2009] is one of them. The ES tries to learn a distribution over the location of the function optimum through Expectation Prop-agation (EP) toward an efficient analytic approximation. It aims to choose the next point that brings the maximum information gain from each evaluation [Villemon-teix *et al.*2009, Hennig and Schuler2012, Hernández *et al.*2014]. But this method is computationally expensive due to complex approximations, such as discretization and EP approximation. Alternatively, Using the symmetry of mutual information, Predictive Entropy Search (PES), proposed by [Hernández *et al.*2014], has an alter-native formulation of the entropy search where expected gain is computed using the predictive distribution of the posterior. This formulation avoids discretization, and through some reasonable approximations, the acquisition function can be computed in an efficient manner. Empirically, PES provides highly competitive convergence performance, although theoretical convergence analysis is still missing.

Optimization of acquisition functions is a crucial aspect in Bayesian optimization.

Acquisition functions are generally non-convex and can be quite challenging to optimize. DIRECT [Jones *et al.*1993, Finkel2003], a method that is developed to optimize Lipschitzian functions without the knowledge of the Lipschitz constant, is a popular choice in the BO community. DIRECT offers an anytime algorithm, i.e. it can be stopped at anytime and the best result so far is the best result one would have obtained given that amount of time. Since, many acquisition functions (e.g. EI, GP-UCB etc.) offers gradient computation in closed-form, the use of multi-start [Hutter *et al.*2011b] local optimizer (e.g. BFGS) is also common.

#### 2.1.1.3 High dimensional Bayesian optimization

In spite of many successes, Bayesian optimization typically only works well in low to moderate dimensional search spaces. It is because that the number of evaluations for finding global optimum in the search spaces can grow exponentially with the number of dimensions and it is difficult to optimize high dimensional acquisition functions [Shahriari *et al.*2015]. There has been some existing studies to deal with high dimensional problems. One approach is to use a compressed sensing strat-egy to attack problems with a high degree of sparsity [Carpentier and Munos2012]. Alternatively, one can apply a two-stage strategy that first selects active dimen-sions [Chen *et al.*2012] and then applys the GP-UCB to optimize over the active dimensions. Another approach is to make an assumption that the function lies on a lower dimensional manifold. i.e. the function has an intrinsic dimensionality much smaller than the dimensionality of the search space. As a result, one can use ran-dom linear projection from a high dimensional space to a low dimensional space and performing optimization in there [Wang *et al.*2013]. These methods may not work if all dimensions in the high dimensional function are similarly effective. Recently, a work [Kandasamy *et al.*2015] assumes the true function is a sum of low dimensional functions and introduces an additive model for Bayesian optimization, resulting in a linear dependence on dimensionality. However, practically, it is difficult to know the decomposition of functions in advance, especially for non-separable functions. More recently, there is a push to develop methods that can efficiently optimize the acquisition function in high dimension [Rana *et al.*2017, Li *et al.*2017b]. As a result, one does not rely on the assumptions that the objective functions depends on a lower dimensional manifold or objective function decomposed in additive forms. Although this direction helps optimizers to obtain optimum in acquisition functions, it does not directly improve the scalability of Bayesian optimization. Most recently, some new methods [Zhang *et al.*2019a, Kirschner *et al.*2019, Tran-The *et al.*2019] have been proposed based on subspace approach, which advance the high dimensional Bayesian optimization significantly. Among them, MS-UCB [Tran-The *et al.*2019] is the most efficient one without marking any restrictive assumption on the func-tion structures. This work also provides a complete convergence analysis. High dimensional Bayesian optimization also naturally needs to sample a large number of samples to reach closer to the optimum, making it related to the work in this thesis.

#### 2.1.1.4 Bayesian optimization for multiple tasks

Many tasks in Bayesian optimization may share information to other related tasks. For instance, when tuning the hyperparameters of a machine learning algorithm on certain dataset, it is unlikely that the hyperparameters will change a lot if a new data is updated to that dataset, especially if the new data only represents a small fraction of the total amount [Wang2016]. On the other hand, many datasets may share similarities in terms of size, number of features and variety, so it may be useful to share hyperparameters from a completed task to a new one. For example, if one were to train one algorithm for classifying Melbourne airport dataset, then the hyperparameter settings are likely to be good on classifying Sydney airport dataset as these two airports share a high degree of similarities in their operations. As a result, one question comes out as whether it is possible for these tasks to share information such as hyperparameters among each other and individually be more sample efficient.

In the literature, there have been attempts to exploit this question in the context of Bayesian optimization [Feurer *et al.*2015, Yogatama and Mann2014, Bardenet *et al.*2013, Swersky *et al.*2013, Hutter *et al.*2011a, Krause and Ong2011]. A general idea is to make an assumption that the given set of tasks can be defined by a set of correlated functions where we are interested in optimizing a subset of them. One way to share information between tasks in a Bayesian optimization routine is to modify the underlying Gaussian process model. More specifically, these GP models include a valid covariance over the training and tasks pairs [Bornn *et al.*2012, Seeger *et al.*2005, Goovaerts and others1997]. Therefore, the tasks are effectively jointly embedded in a latent space that is invariant to potentially different output scales across tasks. Furthermore, these tasks may come with some context features or additional side information. Therefore, one can define a joint model to utilize these information [Hutter *et al.*2011a, Feurer *et al.*2015]. For example, we can use task features to find similar tasks before we start a new task, and then find the best inputs from the most similar tasks that to be used in the initial design for the new task [Feurer *et al.*2015]. The work of [Morales-Enciso and Branke2015] proposes three different ways to reuse information gathered on previous objective functions to speed up optimization of the current objective function. For example, one can use the previous posterior mean function as the prior mean function in the new task, and treat the previous samples as noisy observations. The latter one is also used by [Poloczek *et al.*2016] to warm start Bayesian optimization. Later, the authors in [Pearce and Branke2018] propose two sequential sampling methods for collecting performance measurements of task-parameter pairs such that one can construct a mapping that maximizes the total expected performance across all tasks. Since the pooling the data from existing tasks has a tendency to generate a large number of observations and thus is related to the work of this thesis.

### 2.1.2 Open questions in Bayesian optimization

#### 2.1.2.1 Bayesian optimization with gradient information

In many applications of Gaussian process which involve modeling a black box objective from observed data, the model accuracy could be improved by adding derivative observations [Solak *et al.*2003]. Since differentiation is a linear operation, the derivative of a Gaussian process remains a Gaussian process. The theoretical analysis of using derivative observations is described in [Rasmussen and Williams2006]. This approach has been successfully applied on the real world applications. In the work of [Koistinen *et al.*2016], Gaussian process with derivative observations is used to find minimum energy paths of atomic rearrangements. More specifically, it uses the results of previous calculations [Koistinen *et al.*2016] to construct an approximate energy surface and then it converges to the minimum energy path on that surface in each Gaussian process iteration. As a result, the number of energy evaluations is reduced significantly compared with regular nudged elastic band calculations.

Unlike most optimization methods, traditional Bayesian optimization does not use derivative information. The work of [Siivola *et al.*2017] proposes fully Bayesian optimization procedures that use derivative observations to improve the conditioning of the Gaussian process covariance matrix. However, full gradients (i.e. derivative of all dimensions) are used in this work, which may raise computational issues in high dimension problems. There is another work of [Ahmed *et al.*2016] proposes a simple strategy of using directional derivatives instead of full gradients. By doing this, the memory and computational requirements of existing first-order Bayesian optimization methods have been significantly reduced. The directional derivatives can be obtained from analytic functions even when gradient is not available.

Recent works have looked at ways in which GP can accommodate shape constraints [Riihimäki and Vehtari2010, Lin and Dunson2014, Wang and Berger2016] as empir-ically the functions are known to be in a shape-constrained class, such as the class of monotonic or convex functions. Therefore, we can obtain approximate shape constraints on the original process by imposing conditions on the partial deriva-tives at a sufficient number of points. In the mean time, one work of [Jauch and Peña2016] proposed a methodology for incorporating a variety of shape constraints within Bayesian optimization framework and presented positive results from simple applications, which suggest that Bayesian optimization with shape constraints is a promising topic for further research.

However, these existing methods assume either the availability of the analytic form of functions [Solak *et al.*2003, Ahmed *et al.*2016, Siivola *et al.*2017] or the availability of specific type of information such as monotonicity directions [Riihimäki and Vehtari2010, Lin and Dunson2014, Jauch and Peña2016]. Bayesian optimization methods for functions with more general shape properties such as incorporating the knowledge that the function has only a few peaks has not been addressed before, and thus remains an open problem. In Chapter 4, we will introduce our derivative meta-model based method to fill this gap.

#### 2.1.2.2 Scalability

Another challenge is the scalability of Bayesian optimization. BO using Gaussian process becomes computationally challenging once more than a few hundred data points are collected and it is due to the cubic complexity of GP posterior compu-tation concerning the number of data points [Gelbart2015]. The problem can be more severe if one wishes to include gradient information. It is noted that each gradient adds $d$ separate derivative observations, leading to a more rapid increase in the Kernel matrix in presence of gradients. Other Bayesian models such as deep neural network [Snoek *et al.*2015] and random forest [Hutter *et al.*2011a] have been proposed to replace GP for modeling the objective functions, resulting in a scalable Bayesian optimization. However, these models are not naturally applicable to BO due to their formulation, which can sacrifice the efficiency of modeling the obser-vations. We will explore more details about scalable Bayesian optimization in the next section.

## 2.2 Scalable Bayesian optimization

Bayesian optimization is a leading method for global optimization for expensive black-box functions, where function evaluations are very costly. However, there could be scenarios when a large number of observations become naturally available. For example, in transfer learning, where many algorithms [Pan and Yang2009, Long *et al.*2013] pool existing observations from source tasks together for use in the op-timization of a target task. Then even though the target function is expensive, the number of observations can be large if the number of source tasks is large and/or the number of observations from each source is large. Another scenario where we may have a large number of observations is when we deal with optimization of ob-jective functions which are not very costly. For example, consider the cases when BO is performed using simulation software. They are often used in the early stage of a product design process to reduce a massive search space to a manageable one before real products are made. Evaluation of a few thousand may be feasible, but millions are not because each evaluation can still take from several minutes to hours. Such problems cannot be handled by the traditional global optimizers which often require more than thousands of evaluations. Bayesian optimization is useful in such scenarios, but it has to be able to handle more than few hundreds observations.

The scalability issue for Bayesian optimization has been previously addressed in two main ways: 1) by replacing GP with a more scalable Bayesian model, *e.g.* using Bayesian neural network [Snoek *et al.*2015] or random forest [Hutter *et al.*2011a], or 2) by using scalable form of Gaussian process methods. The latter is often desirable as it still maintains the principled Bayesian formalism of GP. Therefore, we use a scalable Gaussian process heavily in this thesis and the current research in this topic is reviewed in the following subsections.

## 2.2.1 Scalable Gaussian process

### 2.2.1.1 Sparse Gaussian process

The Sparse Gaussian process is one of the most important approaches for scalable Gaussian processes [Hensman *et al.*2013] which employs a low-rank approximation of the kernel matrix. Generally, one can conduct eigen-decomposition and choose the first $m$ eigenvalues to approximate the full-rank ($t$) kernel matrix. As a re-sult, it brings down the computation complexity to $O(tm^2)$. However, the eigen-decomposition is of limited interest since it is an $O(t^3)$ operation. Therefore, as an alternative, one can approximate the eigen-functions using $m$ data points, result-ing in the Nystro¨m approximation that greatly improves large scale kernel learning [Gittens and Mahoney2016]. Nevertheless, this scalable GP may produce negative predictive variances [Williams *et al.*2002] as 1) it cannot guarantee the Positive Semi-Definite (PSD) [Gneiting2002, Melkumyan and Ramos2009] of kernel matrix and 2) it is not a complete generative probabilistic model as the Nystro¨m approximation is only imposed on the training data. Inspired by the Nystro¨m approximation, sparse approximations build a generative probabilistic model, which enables the sparsity by using $m$ inducing points (also known as a sparse set or sparse features) to optimally summarize the dependency of the entire training data.

There are three main categories in sparse approximations [Liu *et al.*2018]:

- prior approximations which approximate the prior but perform exact inference;
- posterior approximations which approximate the inference but keep the exact prior; and
- structured sparse approximations which exploit specific structures in the kernel matrix.

**Prior approximations** Prior approximations use the independence assumption that the dependencies between training and testing points are only induced through inducing variables [Quinonero-Candela and Rasmussen2005] so that we can modify the joint prior. By using this modification, the computational complexity has been substantially reduced to O($tm^2$). A good case in point is the Subset of Regressors (SoR) [Smola and Bartlett2001], also known as Deterministic Inducing Conditional (DIC). The SoR imposes deterministic training and testing conditionals making it equivalent to applying Nystro¨m approximation to both training and testing data, resulting in a degenerate (kernel with a finite number of non-zero eigenvalues) GP. However, due to the restrictive assumptions of this approach it produces overconfi-dent prediction variances when leaving the training data. To rectify the uncertainty behavior of SoR we can choose an alternative way to impose more informative as-sumptions on training and testing conditionals. For example, the Deterministic Training Conditional (DTC) [Csató and Opper2002, Herbrich *et al.*2003], only im-poses the deterministic training conditional but retains the exact testing conditional (i.e. it only applies Nystro¨m approximation on training data). Therefore, the pre-dictive variance of DTC is more accurate than that of SoR, while the prediction mean is the same as that of SoR. Although the DTC can produce a larger predic-tive variance, it is not an exact GP due to the inconsistent conditionals. Moreover, both SoR and DTC do not perform well in practice since the restrictive training conditional is assumed.

As an alternative, one can use the Fully Independent Training Conditional (FITC) [Quinonero-Candela and Rasmussen2005, Snelson and Ghahramani2006], which im-poses a fully independence assumption to remove the dependency among training data, resulting in a matrix that only keeps exact covariance on diagonal. However, the testing conditional retains exactly in FITC. Therefore, not like SoR and DTC which ignore the uncertainty in prior approximation, FITC retains the uncertainty in test data. Thus, FITC performs a closer approximation to the prior distribution and holds identical computational complexity to that of SoR and DTC. Further-more, one can extend the fully independence assumption to testing data to derive the Fully Independent Conditional (FIC) [Quinonero-Candela and Rasmussen2005]. To further improve the sparse approximation, the work of [Quinonero-Candela and Rasmussen2005] proposes to utilize the Partially Independent Training Conditional (PITC) that partitions the training data into *m* blocks and enforce independence between these *m* blocks. By retaining the block diagonal covariance matrix, the PITC keeps more information than that in FITC, so that it performs slightly bet-ter [Snelson and Ghahramani2007]. Similar to FITC, one can extend the partially independence assumption to testing data to receive the Partially Independence Con-ditional (PIC) [Snelson and Ghahramani2007].

There is another line of work that one can interpret the FITC from the spectral representation view. It is known that any stationary covariance function can be represented as the Fourier transform of some finite measure and the GP using such kernel can be approximated by a finite basis function expansion. One of the most popular interpretation is knowing as the sparse spectrum Gaussian process (SSGP) [Lazaro Gredilla *et al.*2010], which uses *m* optimal basis function to approximate the full GP and reduces the computational complexity to $O(tm^2)$. Its variational variants [Gal and Turner2015, Tan *et al.*2016, Hoang *et al.*2017] are also available with a desirable predictive accuracy.

Until now, we have introduced state of the art prior approximations such as FITC and SSGP. These sparse models require a careful choice of inducing variables where we can optimize along with other kernel hyperparameters [Snelson and Ghahra-mani2006]. Although one can faithfully recover the full GP by increasing inducing points, it is not a practical method as it turns the parameter optimization into a high dimensional task [Liu *et al.*2018]. Therefore, choosing the optimal set of induc-ing points is important. Some approaches greedily select inducing variables from training data [Smola and Bartlett2001, Herbrich *et al.*2003, Keerthi and Chu2006], while other methods use fitting optimization (e.g. log marginal likelihood ) [Snel-son and Ghahramani2006, Walder *et al.*2008] with respect to the choice of inducing variables. We use the later one in Chapter 5 and Chapter 6. Next, we will review the posterior approximations.

**Posterior approximations** Posterior approximations [Titsias2009, Hensman *et al.*2013], different from prior approximations, perform approximate inference but retain exact prior. The Variational Free Energy (VFE) [Titsias2009] is the most popular posterior approximation which makes use of variational inference [Blei *et al.*2017]. The VFE directly approximates the posterior distribution rather than modifying the prior. It introduces a variational distribution and then minimizes KL divergence between this distribution and the true posterior distribution. By doing so, it is equivalent to maximize the Evidence Lower Bound (ELBO) so that one can jointly optimize the variational parameters and kernel hyperparameters.

However, it has been noted in [Cao *et al.*2013] that in VFE, the similarity of inducing points measured in the Euclidean space is inconsistent to that measured by the GP kernel function. Therefore, the work of [Zhe2017] derived a regularized bound for choosing good inducing points in the kernel space. Moreover, in [Matthews *et al.*2016, Matthews2017], the researchers proposed their methods to bridge the gap between the variational inducing points framework and the more general KL divergence between stochastic process. Once this gap has been filled, one can approximate both posterior and evidence directly by minimizing the KL divergence. Hence, the VFE and FITC are interpreted jointly as a hybrid approximation and this often produces better predictions [Bui *et al.*2017].

To further increase the scalability of VFE, Hensman *et al*. [Hensman *et al.*2013] proposed to employ the Stochastic Variational Inference (SVI) [Hoffman *et al.*2013] using natural gradients [Salimbeni *et al.*2018] to optimize variational parameters. As a result, the model stochastic variational inference Gaussian process (SVIGP) can reach a remarkable complexity of $O(m^3)$. More specifically, in each iteration, the SVIGP trains a sparse GP with a small subset of the training data. Despite desirable approximation and high scalability, the SVIGP also comes with certain drawbacks such as 1) the bound of ELBO is less tight than that in VFE and 2) the approximation requires optimization of a large number of variational parameters and thus expensive in practice.

Inspired by the idea of SVIGP, asynchronous distributed variational Gaussian process (ADVGP) [Peng *et al.*2017] derived a similar factorized variational bound for GP in the spectrum domain. Through this, one can use flexible basis functions to incorporate various low-rank structures and speed up the efficiency of the existing variational methods as the ADVGP provides a composite non-convex bound. Thus, it is the first model that can scale GP to fit billions of training data. Later, VFE and SVIGP have been further improved by using Bayesian treatment of hyperpa-rameters [AUEB and Lázaro-Gredilla2013, Hensman *et al.*2015, Yu *et al.*2017] rather than traditional point estimation which suffers from the risk of overconfidence when the number of hyperparameters is small.

Recently, the work of [Burt *et al.*2019] provides a concrete theoretical analysis that demonstrates GP posteriors can be approximated cheaply for large scale data. More importantly, it shows that with a high probability the KL divergence can be made arbitrarily small by growing $m$ more slowly than $t$. A rule for how to increase $m$ in continual learning scenarios is also provided.

**Structured sparse approximations** Apart from prior and posterior approximations, many studies also exploit specific structures in the kernel matrix to scale the GP. For example, the Matrix Vector Multiplication (MVM) [Shen *et al.*2006, Morariu *et al.*2009] is an efficient method to directly accelerate the inverse of the covariance matrix in generic GP. It allows us to iteratively solve the linear system using Conjugate Gradients (CG) within a small number of iterations. However, there are issues in original MVM such as the determination of iterations is not well studied and it does not protect from ill-conditioned kernel matrix. Therefore, one can employ the preconditioned CG (PCG) [Cutajar *et al.*2016] where a preconditioning matrix is been used to improve the conditioning of the covariance matrix and speedup the convergence of conjugate gradients.

It is discovered that the MVM provides massive scalability when the kernel matrix itself has some algebraic structure. A good case in point is the Kronecker methods [Saatçi2012, Gilboa *et al.*2013], which decompose the kernel matrix to a Kronecker product, resulting in a significantly reduced time complexity. However, these methods are limited to grid inputs, so the general arbitrary data points cannot be applied. Alternatively, one can use the Structured Kernel Interpolation (SKI) [Wilson and Nickisch2015] to handle arbitrary data while keeping the efficient Kronecker structure. But, the number of inducing variables grows exponentially with dimensionality in SKI, so it is difficult to extend it to multi-dimension and high dimension problems. To address this issue, dimensionality reduction has been considered as a solution where one can map the inducing variables into a low dimensional latent space [Wilson *et al.*2015]. Moreover, due to the use of SoR framework, the SKI may suffer from overconfidence on variance prediction. To address this issue, Dong *et al.* [Dong *et al.*2017, Evans and Nair2018] proposed to construct a diagonal correlation as that of FITC, resulting in a more accurate uncertainty estimate.

However, our literature review failed to find a sparse approximation for Gaussian process that can include derivative observations from any category. Therefore, we derived an efficient sparse Gaussian process model to approximate the full GP with derivative observations in Chapter 5 and applied to Bayesian optimization.

#### 2.2.1.2 Other scalable Gaussian process

**Distributed Gaussian process** Unlike the sparse Gaussian process approximations, the distributed Gaussian process [Deisenroth and Ng2015] does not rely on inducing or variational parameters. The key idea of distributed GP is to operate on the full data set but recursively distribute computations to independent computational units and subsequently recombine them to form an overall result. This hierarchical model includes the Bayesian committee machine (BCM) [Tresp2000] and the generalized Product-of-GP-Experts (PoE-GP) [Cao and Fleet2014] as special cases. As a result, the distributed GP can consistently approximate a large scale full GP. However, the main drawback of this line of work is the Kolmogorov inconsistency [Samo and Roberts2016]. It is caused by the separation of training and prediction, resulting in a non-unifying probabilistic framework and producing unreliable predictive mean when leaving training data.

**Exact Gaussian process on a million data points** Recently, there has been a push to scale GP inference to millions of data points using modern hardware (e.g. GPUs) [Wang *et al.*2019]. However, it still remains computationally demanding and vulnerable to kernel matrix ill-conditioning, which is a common issue in the presence of a large number of data points in GP. Hence, it is infeasible for practical use. In contrast, sparse GP provides a recourse, which is fast, cheap and resilient to kernel matrix ill-conditioning.

#### 2.2.1.3 Sparse Gaussian process for Bayesian optimization

To increase the scalability of Bayesian optimization, we can directly apply the sparse Gaussian process models for Bayesian optimization by replacing the full GP. How-ever, it is noted that these sparse models suffer from either variance underestimation (i.e. overconfidence) [Snelson and Ghahramani2006, Lazaro Gredilla *et al.*2010] or overestimation, compared to the full GP [Titsias2009] and thus may hamper BO as the balance between predictive mean and variance is important to the success of BO. Recently, the work in [Hensman *et al.*2017] has proposed a method named variational Fourier features (VFF), which combines variational approximation and spectral representation of GP together and plausibly can approximate both mean and variance well. However, it is difficult to extend VFF to multiple dimensional problems. It is because a) the number of inducing variables grows exponentially with dimensions if the Kronecker kernel is used, and b) the correlation between dimensions would be ignored if an additive kernel is used.

To improve the usability of the sparse Gaussian process in Bayesian optimization, Mcintire *et al.* [McIntire *et al.*2016] proposed the weighted-update online Gaussian process (WOGP) that utilizes a weighted measure of divergence between the pre-dictive distribution of the sparse GP and that of the full GP. As a result, we can have better control of allocating the limited modeling resources to the goal of opti-mization (i.e. allocate more resources to important regions). However, although the WOGP wisely allocates more resources to high function value regions, it does not consider the case that the regions of the posterior function value are lower but the uncertainty are higher (i.e. under-explored regions). To address the research gap, in Chapter 6, we developed a sparse Gaussian process model tailored for Bayesian optimization.

## 2.3 Bayesian optimization libraries

There are several libraries available to the public for incorporating Bayesian opti-mization with state of the art programmings.

- GPflowOpt [De G. Matthews *et al.*2017] is written in Python and uses TensorFlow [Abadi *et al.*2015] to accelerate computation on GPU hardware. It supports multi-objective acquisition functions and black-box constraints. The source files are available from https://github.com/GPflow/GPflowOpt.
- GPyOpt [authors2016] is a Bayesian optimization framework, written in Python and supporting parallel optimization, mixed factor types (continuous, discrete, and categorical), and inequality constraints. The source files are available from https://github.com/SheffieldML/GPyOpt.
- BayesOpt [Martinez-Cantin2014] is written in C++ and includes common in-terfaces for C, C++, Python, Matlab, and Octave. The source files are avail-able from https://github.com/rmcantin/bayesopt.
- SigOpt [Hayes *et al.*2019] offers Bayesian optimization as a web service. The implementation is based on MOE but includes some enhancements such as mixed factor types (continuous, discrete, categorical), and automatic hyper-parameter tuning. The source files are available from http://sigopt.com.
- DiceOptim is a BO package written in R. Mixed equality and inequality constraints are implemented using the method of [Picheny *et al.*2016], and parallel optimization is via multi-point EI [Ginsbourger *et al.*2008], however parallel and constraints cannot be mixed in a single optimization. The source files are available from https://cran.r-project.org/web/packages/DiceOptim/index. html.

## 2.4 Summary

This chapter provided a comprehensive review on Bayesian optimization and its components. We have identified the open questions of Bayesian optimization with gradient information and the scalability of Bayesian optimization. We have also reviewed the state of the art methods for scalable Bayesian optimization. The popular existing Bayesian optimization libraries have been given at the end of this chapter.

# Chapter 3

# Mathematical Background

This chapter describes the mathematical background on which this thesis is built. In Section 3.1, we provide an overview of the role of Bayesian optimization. In Section 3.2, we include an elaborative description of the generic Gaussian process. Next, we discuss the variety of acquisition functions in Section 3.3. We then introduce the Gaussian process with derivative observations in Section 3.4, follow by the sparse Gaussian process with fully independent conditionals in Section 3.5. Finally, we discuss the sparse spectrum Gaussian process in Section 3.6.

## 3.1 Bayesian optimization

Bayesian optimization is a method for performing global optimization of black-box objective functions that are expensive to evaluate and possibly noisy. In this thesis, we consider the maximization problem of the following kind

$$\boldsymbol{x}^* = \mathrm{argmax}_{\boldsymbol{x} \in \mathcal{X}} f(\boldsymbol{x}), \tag{3.1}$$

where $f : \boldsymbol{x} \to \mathbb{R}^d$ is a smooth function, and $\mathcal{X}$ is a compact subspace in $\mathbb{R}^d$.

Bayesian optimization proceeds by iteratively developing a global statistical model of the objective function. We begin with a prior distribution over objective functions and a likelihood function that encodes our assumptions about the noise presence in our observations of the objective function. Then, at each iteration, a posterior distribution is computed by conditioning on the previous evaluations of the objective function, treating them as observations in a Bayesian nonlinear regression. This is often achieved with Gaussian processes, which are described in Section 3.2. An acquisition function is then used and queried liberally in order to select a high belief candidate for the next function evaluation. Once the function evaluation is complete, the new observation is augmented to the data set and the next iteration follows. Usually, Bayesian optimization is terminated when either the specific budget on the objective function evaluation is reached or a good enough solution has been found. We outline the iterative process of Bayesian optimization in Figure 3.1

In the next few sections, we will go into more detail about the components of the Bayesian optimization framework.

## 3.2 Gaussian process

Gaussian process (GP) can be used as a prior distribution over the space of smooth functions. It is a distribution over functions and the properties of the Gaussian distribution allow us to compute the predictive means and variances in closed form. A GP is specified by its mean function, $\mu(\boldsymbol{x})$ and covariance function, $k(\boldsymbol{x}, \boldsymbol{x}^0)$. A sample from a Gaussian process is a function:

$$f(\boldsymbol{x}) \sim GP\left(\mu(\boldsymbol{x}), k(\boldsymbol{x}, \boldsymbol{x}^0)\right) \tag{3.2}$$

where the value $f(\boldsymbol{x})$ at an arbitrary $\boldsymbol{x}$ is a Gaussian distributed random variable. Without any loss in generality, the prior mean function can be assumed to be a zero function making the Gaussian process fully defined by the covariance function.

Given a set of noise-free observations $D_{1:t} = \{\boldsymbol{x}_i, f_i\}_{i=1}^{t}$, the function values are drawn according to a multivariate normal distribution $N(0, \mathbf{K})$, where the kernel

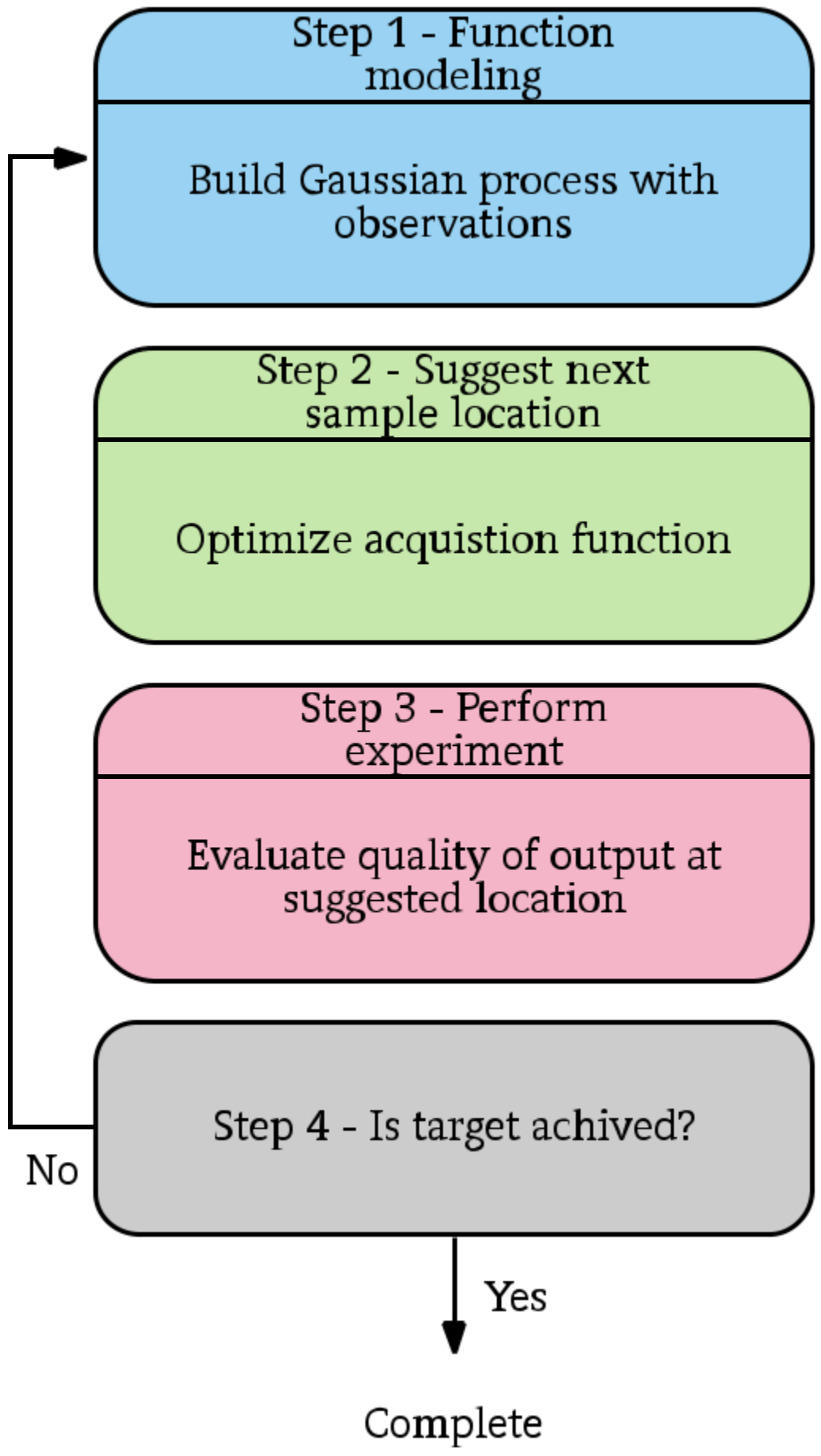

Figure 3.1: The iterative process of Bayesian optimization where the objective func-tion is modeled by a Gaussian process and an optimized acquisition function ex-presses the most promising location for conducting next function evaluation. The model quality improves progressively over time as successive measurements are in-corporated.

matrix is given by

$$\mathbf{K} = \begin{bmatrix} k(\boldsymbol{x}_1, \boldsymbol{x}_1) & \ldots & k(\boldsymbol{x}_1, \boldsymbol{x}_t) \\ \vdots & \ddots & \vdots \\ k(\boldsymbol{x}_t, \boldsymbol{x}_1) & \ldots & k(\boldsymbol{x}_t, \boldsymbol{x}_t) \end{bmatrix}. \tag{3.3}$$

In GP, the joint distribution for any finite set of random variables are multivariate Gaussian distribution. Therefore, the joint distribution of $f_{t+1}$ and $\boldsymbol{f}_{1:t}$ can be written as

$$\begin{bmatrix} \boldsymbol{f}_{1:t} \\ f_{t+1} \end{bmatrix} \sim \mathcal{N}\left(0, \begin{bmatrix} \mathbf{K} & \mathbf{k} \\ \mathbf{k} & k(\boldsymbol{x}_{t+1}, \boldsymbol{x}_{t+1}) \end{bmatrix}\right), \tag{3.4}$$

where $\mathbf{k} = [\, k(\boldsymbol{x}_{t+1}, \boldsymbol{x}_1), \quad k(\boldsymbol{x}_{t+1}, \boldsymbol{x}_2), \quad \ldots, \quad k(\boldsymbol{x}_{t+1}, \boldsymbol{x}_t)]$.

Using Sherman-Morrison-Woodbury formula, the predictive distribution of the func-tion value at a new location $\boldsymbol{x}_{t+1}$ can be written as

$$P\,(f_{t+1}|\mathrm{D}_{1:t}, \boldsymbol{x}_{t+1}) = \mathcal{N}\,(\mu(\boldsymbol{x}_{t+1}), \sigma^2(\boldsymbol{x}_{t+1})), \tag{3.5}$$

where the predicted mean and the variance is given by

$$\mu(\boldsymbol{x}_{t+1}) = \mathbf{k}^T\, \mathbf{K}^{-1}\boldsymbol{f}_{1:t}, \tag{3.6}$$

$$\sigma^2(\boldsymbol{x}_{t+1}) = k(\boldsymbol{x}_{t+1}, \boldsymbol{x}_{t+1}) - \mathbf{k}^T\, \mathbf{K}^{-1}\mathbf{k}. \tag{3.7}$$

If the observation is a noisy estimate of the actual function value as

$$y = f(\boldsymbol{x}) + \xi, \tag{3.8}$$

where $\xi \sim \mathcal{N}\,(\mathbf{0}, \sigma_n^2)$. Then the predicted mean and the variance is given by

$$\mu(\boldsymbol{x}_{t+1}) = \mathbf{k}^T\, [\mathbf{K} + \sigma_n^2\mathbf{I}]^{-1}\boldsymbol{y}_{1:t}, \tag{3.9}$$

$$\sigma^2(\boldsymbol{x}_{t+1}) = k(\boldsymbol{x}_{t+1}, \boldsymbol{x}_{t+1}) - \mathbf{k}^T\, [\mathbf{K} + \sigma_n^2\mathbf{I}]^{-1}\mathbf{k}. \tag{3.10}$$

Figure 3.2 shows some samples of the GP priors and Figure 3.3 shows the associated posteriors given 5 observations. We are going to discuss the kernel functions that we have used in this thesis in the coming sub-sections.

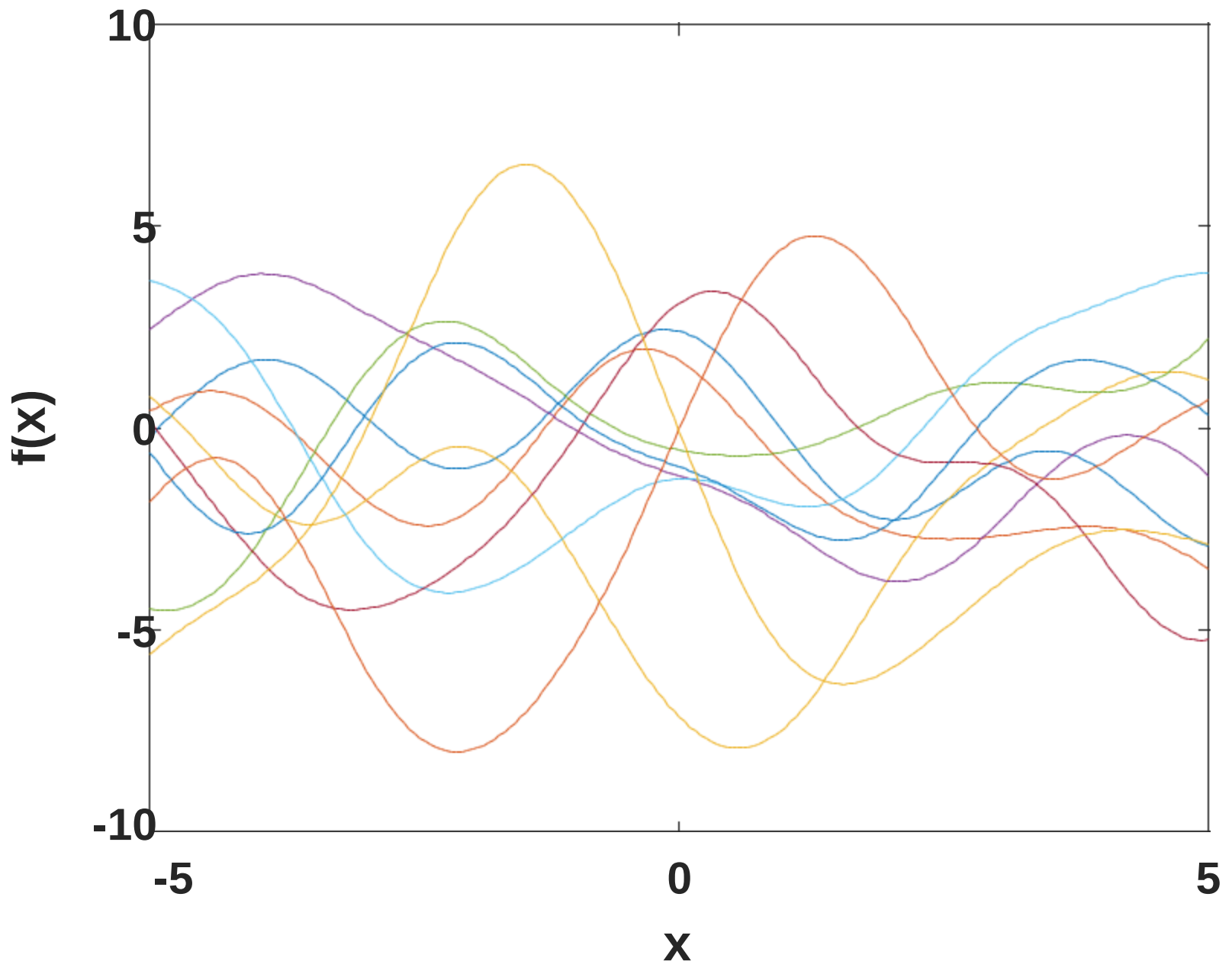

Figure 3.2: Samples of 10 functions from the prior of a 1*d* Gaussian process without any observations. In the absence of any observations, the shape of the sample functions are fully controlled by the covariance function.

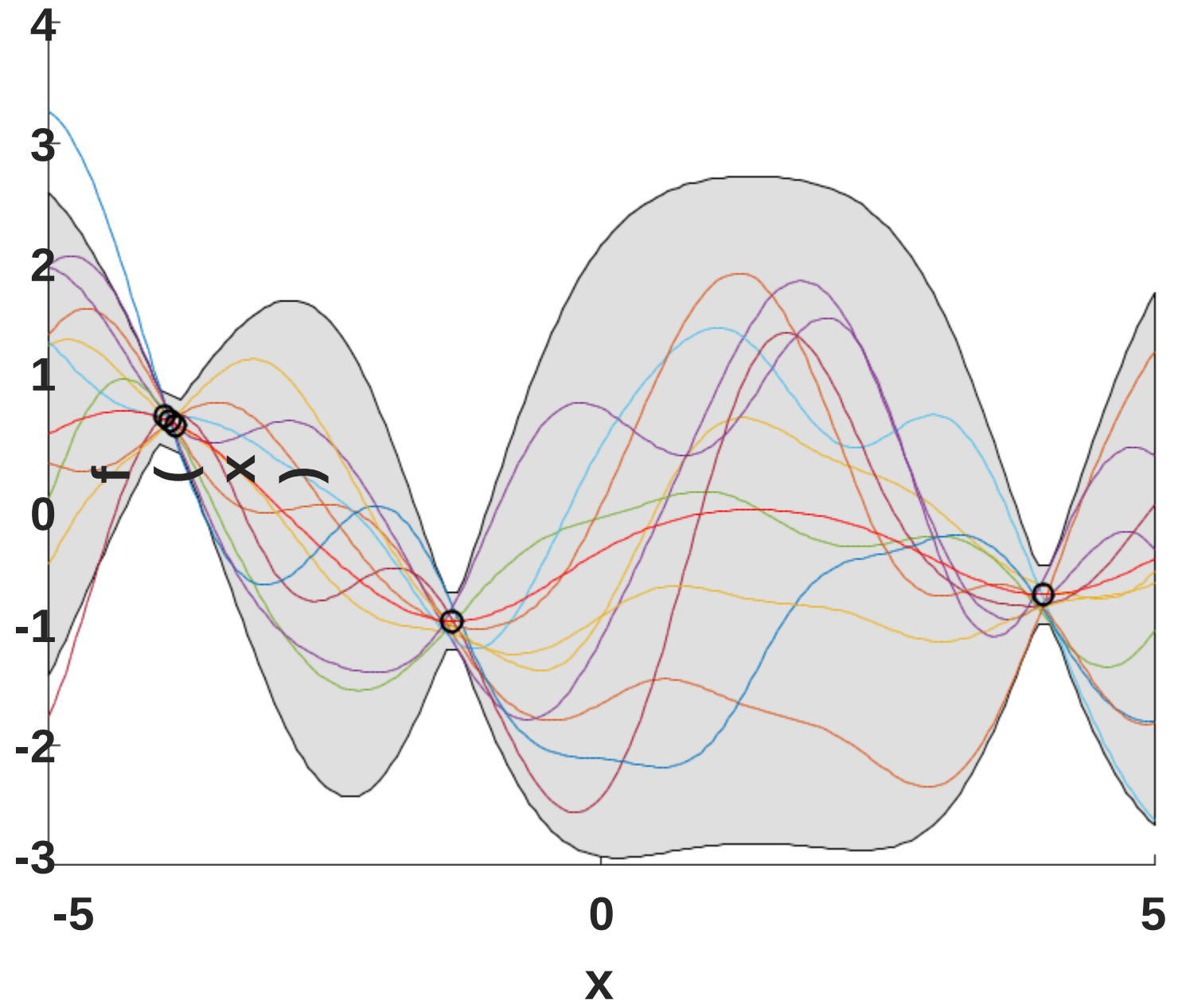

Figure 3.3: Samples of 10 functions from the posterior with the uncertainty of a 1*d* Gaussian Process. The solid red line is the posterior GP mean function with 5 observations, and the shaded area covers the one standard deviation on both sides of the mean.

### 3.2.1 Squared exponential kernel

Squared exponential (also known as Radial Basis Function or Gaussian) kernel is a popular choice of the covariance function, it is defined as

$$k(\boldsymbol{x}, \boldsymbol{x}') = \sigma_f^2 \exp\left(-\frac{1}{2}\frac{\|\boldsymbol{x} - \boldsymbol{x}'\|^2}{\rho_l^2}\right), \tag{3.11}$$

where $\|\cdot\|$ is the $L_2$ vector norm, $\rho_l$ is the characteristic length-scale, and $\sigma_f$ is the signal standard deviation. This is a stationary kernel i.e. the covariance is function of the distance between two points only and the covariance drops off with distance. This choice of covariance function also gives rise to Gaussian process which produces infinitely differentiable functions as samples. If $\rho_l = \infty$, the corresponding dimension is ignored and that is how the automatic relevance determination is achieved.

### 3.2.2 Matérn kernel

The Matérn kernel is another popular choice of kernel for Bayesian optimisation. The Matérn kernels are parameterised by a smoothness factor. When the smoothness factor approaches infinity, it becomes a squared exponential kernel. The commonly used Matérn 3/2 kernel with smoothness factor is given as

$$k(\boldsymbol{x}, \boldsymbol{x}') = \sigma_f^2\left(1 + \frac{\sqrt{3\|\boldsymbol{x} - \boldsymbol{x}'\|^2}}{\sigma_l}\right)\exp\left(-\frac{\sqrt{3\|\boldsymbol{x} - \boldsymbol{x}'\|^2}}{\sigma_l}\right). \tag{3.12}$$

Unlike squared exponential kernel, Matérn 3/2 produces more non-smooth functions for which higher than 4th order derivatives do not exist.

### 3.2.3 Polynomial kernel

Polynomial kernel is a dot product covariance function as it depends only on $\boldsymbol{x}$ and $\boldsymbol{x}'$ through their dot product. The kernel is defined as

$$k(\boldsymbol{x}, \boldsymbol{x}') = (\sigma_0^2 + \boldsymbol{x} \cdot \boldsymbol{x}'), \tag{3.13}$$

where is the degree of the polynomial and $\sigma_0^2$ is a free parameter trading off the influence of higher-order versus lower-order terms in the polynomial, and builds a Bayesian polynomial regression model [Rasmussen and Williams2006].

### 3.2.4 Learning GP hyperparameters

These kernels are parameterized, for example, the SE has hyperparameters such as length-scale which determines the smoothness of the function and kernel variance which determines the magnitudes of the function. Although we can use fixed hy-perparameters, a mismatch of hyperparameters and data may lead to a very poor performance. For example, if the length-scales are set too large, then the GP prior will overlook the higher frequency variations in the true function; on the other hand, if the length-scales are too small, the GP will fail to generalize across meaningful distances. There are many methods for learning hyperparameters in the literature such as the leave one out method in cross-validation criteria and optimizing the marginal likelihood criteria [Rasmussen and Williams2006]. We use the later one in this thesis. The parameters of the kernel and the observation noise then can be trained by maximizing the log-marginal likelihood of the observations log $p(\boldsymbol{y}_{1:t} \mid \Theta)$, where $\Theta$ is the set of all hyperparameters in the kernel function, and the latent func-tion values are marginalized. The expanded log marginal likelihood of a GP can be given as

$$\log p(\boldsymbol{y}_{1:t}|\Theta) = -\frac{1}{2}\boldsymbol{y}^T(\mathbf{K} + \sigma_n^2\mathbf{I})^{-1}\boldsymbol{y} - \frac{1}{2}\log|\mathbf{K} + \sigma_n^2\mathbf{I}| - \frac{t}{2}\log 2\pi. \quad (3.14)$$

Since it is possible to analytically compute the gradient of Eq.6.8, in our experiments we use a gradient based optimization routine to obtain the optimized hyperparam-eters at every iteration.

## 3.3 Acquisition functions

Once the surrogate is modeled using the Gaussian process, the next evaluation point is obtained by constructing a utility function based on the posterior. The utility function, also known as acquisition function, has to be defined in such a way that it carefully trades off exploitation and exploration of function values. Exploitation implies the areas where the posterior mean is high, while exploration implies the areas where posterior uncertainties are high. The maximizing of the acquisition function provides the new evaluation point. Below we discuss a few common acquisition functions, followed by a discussion on the ways to optimize them.

### 3.3.1 Probability of improvement

This acquisition function [Kushner1964] operates by maximizing the probability of improvement (PI) over the current best observation $f(\boldsymbol{x}^+)$. It is defined as

$$P(I(\boldsymbol{x})) = \varphi\left(\frac{\mu(\boldsymbol{x}) - f(\boldsymbol{x}^+)}{\sigma(\boldsymbol{x})}\right), \tag{3.15}$$

where $\varphi(\cdot)$ is the probability distribution function of the standard normal. The goal of PI is to find the maximum probability of improvement. The main drawback of this method is that it is more exploitative in nature. Often, a trade-off parameter is used to increase the exploration by trading off with exploitation as

$$P(I(\boldsymbol{x})) = \varphi\left(\frac{\mu(\boldsymbol{x}) - f(\boldsymbol{x}^+) - \xi}{\sigma(\boldsymbol{x})}\right), \tag{3.16}$$

where $\xi > 0$ is a user defined exploration bias.

### 3.3.2 Expected improvement

Instead of measuring the probability of improvement, one can measure the expected improvement (EI) for a balanced exploration and exploitation. For the maximization problem of Eq.(3.1), the improvement function can be written as

$$I(\boldsymbol{x}) = max\{0, f(\boldsymbol{x}) - f(\boldsymbol{x}^+)\}, \tag{3.17}$$

where $f(\boldsymbol{x}^+)$ is current best solution [Brochu *et al.*2010]. The improvement function will be positive if the predicted value of the random variable $f(\boldsymbol{x})$ is greater than $f(\boldsymbol{x}^+)$, otherwise it is zero. The likelihood of improvement $I$ on a normal posterior distribution characterized by $\mu(\boldsymbol{x})$ and $\sigma^2(\boldsymbol{x})$ can be computed from the normal density function

$$\frac{1}{\sqrt{2\pi}\sigma(\boldsymbol{x})} \exp\left(-\frac{(\mu(\boldsymbol{x}) - f(\boldsymbol{x}^+) - I)^2}{2\sigma^2(\boldsymbol{x})}\right) \tag{3.18}$$

The expected improvement can be computed by integrating function $I$

$$\begin{aligned} \mathbb{E}(I(\boldsymbol{x})) &= \int_{I=0}^{I=\infty} I \frac{1}{\sqrt{2\pi}\sigma(\boldsymbol{x})} \exp\left(-\frac{(\mu(\boldsymbol{x}) - f(\boldsymbol{x}^+) - I)^2}{2\sigma^2(\boldsymbol{x})}\right) dI \\ &= \sigma(\boldsymbol{x}) \left[ \frac{\mu(\boldsymbol{x}) - f(\boldsymbol{x}^+)}{\sigma(\boldsymbol{x})} \Phi\left(\frac{\mu(\boldsymbol{x}) - f(\boldsymbol{x}^+)}{\sigma(\boldsymbol{x})}\right) + \varphi\left(\frac{\mu(\boldsymbol{x}) - f(\boldsymbol{x}^+)}{\sigma(\boldsymbol{x})}\right) \right] \end{aligned} \tag{3.19}$$

The analytic form of $\mathrm{EI}(\boldsymbol{x})$ [Mockus1994] can be obtained as

$$\mathrm{EI}(\boldsymbol{x}) = \begin{cases} \sigma(\boldsymbol{x}) Z \Phi(Z) + \sigma(\boldsymbol{x}) \varphi(Z) & \sigma(\boldsymbol{x}) > 0 \\ 0 & \sigma(\boldsymbol{x}) = 0 \end{cases}, \tag{3.20}$$

where $Z = \frac{\mu(\boldsymbol{x}) - f(\boldsymbol{x}^+)}{\sigma(\boldsymbol{x})}$, $\Phi(\cdot)$ and $\varphi(\cdot)$ are the cumulative distribution function and probability density function of the standard normal distribution, respectively. Since EI can work well without human efforts i.e. without hyperparameters, we choose EI as the acquisition function in all of our Bayesian optimization experiments.

In Figure 3.4 we illustrate the behavior of Bayesian optimization using EI as the acquisition function. The GP posterior is shown at top as solid red line. The dotted blue line is the true function and green crosses are current observations. The lower image shows the acquisition functions for that GP. The maximum of acquisition function is shown with a blue circle marker.

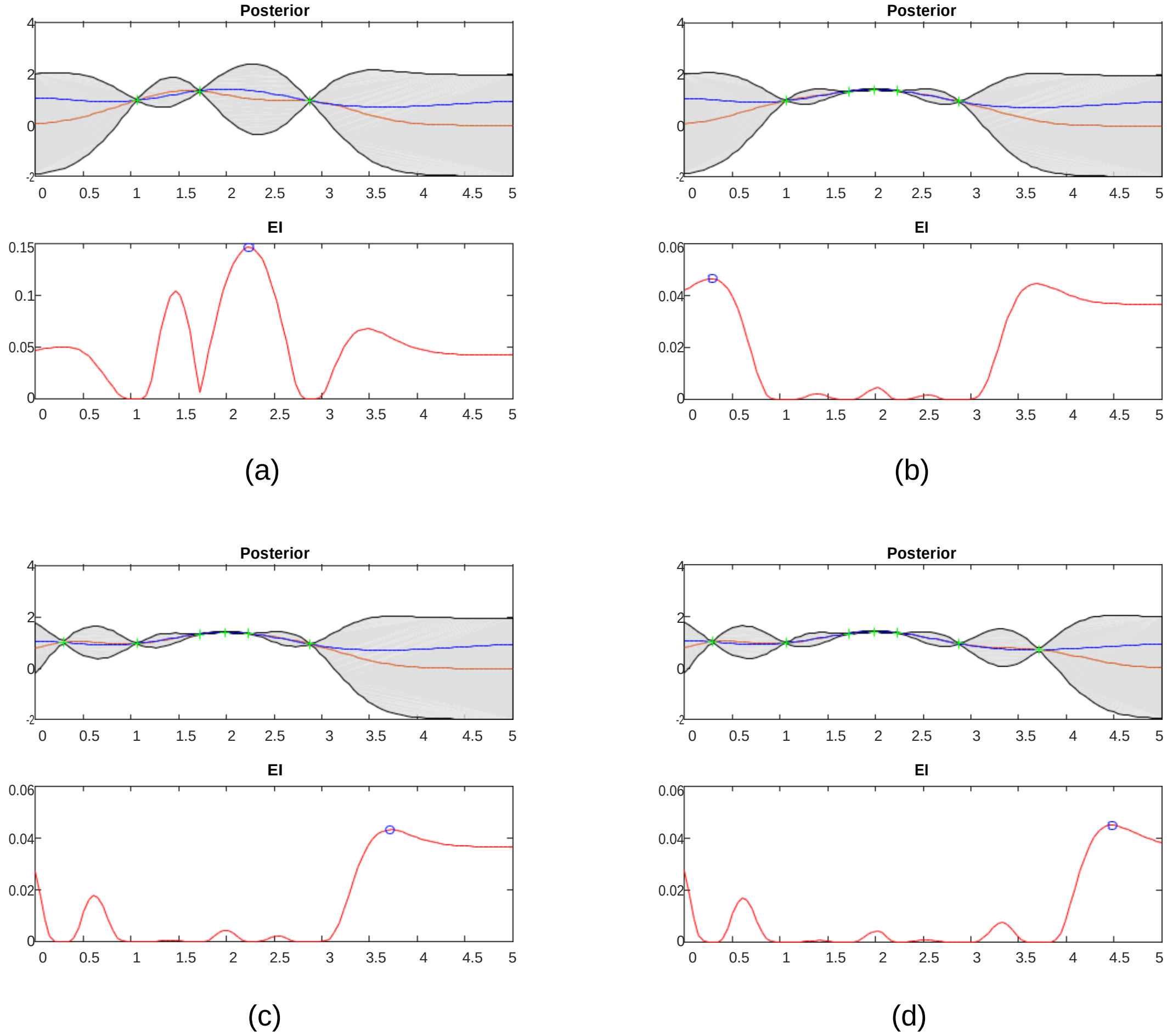

Figure 3.4: Illustration of using EI as acquisition function in Bayesian optimization on a toy 1*d* problem. The GP posterior is shown at the top as a solid red line, the dotted blue line is the true function and the green crosses are current observations. The lower image shows the acquisition functions for that GP. The maximum of the acquisition function is shown with a blue circle marker.

### 3.3.3 Upper confidence bound based

Lai and Robbins[Lai and Robbins1985] suggested using upper confidence bound (UCB) as a way of controlling the exploration and exploitation in multi-armed bandit scenarios. The upper confidence bound is defined as

$$UCB(\boldsymbol{x}) = \mu(\boldsymbol{x}) + \tau\, \sigma(\boldsymbol{x}), \tag{3.21}$$

where $\tau \geq 0$ . Recently, a work [Srinivas *et al.*2010] proposed a modified form for Bayesian optimization by treating it as an infinite-armed bandit problem but with correlated arms. However, in this case $\tau$ grows sub-linearly with $t$. By doing this they were able to prove that $f(\boldsymbol{x}^*) - f(\boldsymbol{x}_t)$ goes to 0 as $t \to \infty$ and $\sum(f(\boldsymbol{x}^*) - f(\boldsymbol{x}_t))$ only grows sub-linearly with $t$. Between EI and GP-UCB, EI is parameter free, whilst GP-UCB has parameters to define the sequence of $\tau$. However, whilst EI convergence requires noise-free setting, GP-UCB converges for noisy scenarios too.

In Figure 3.5 we demonstrate the behavior of Bayesian optimization using UCB as the acquisition function. The GP posterior is shown at top as solid red line. The dotted blue line is the true function and green crosses are current observations. The lower image shows the acquisition functions for that GP. The maximum of acquisition function is shown with a blue circle marker.

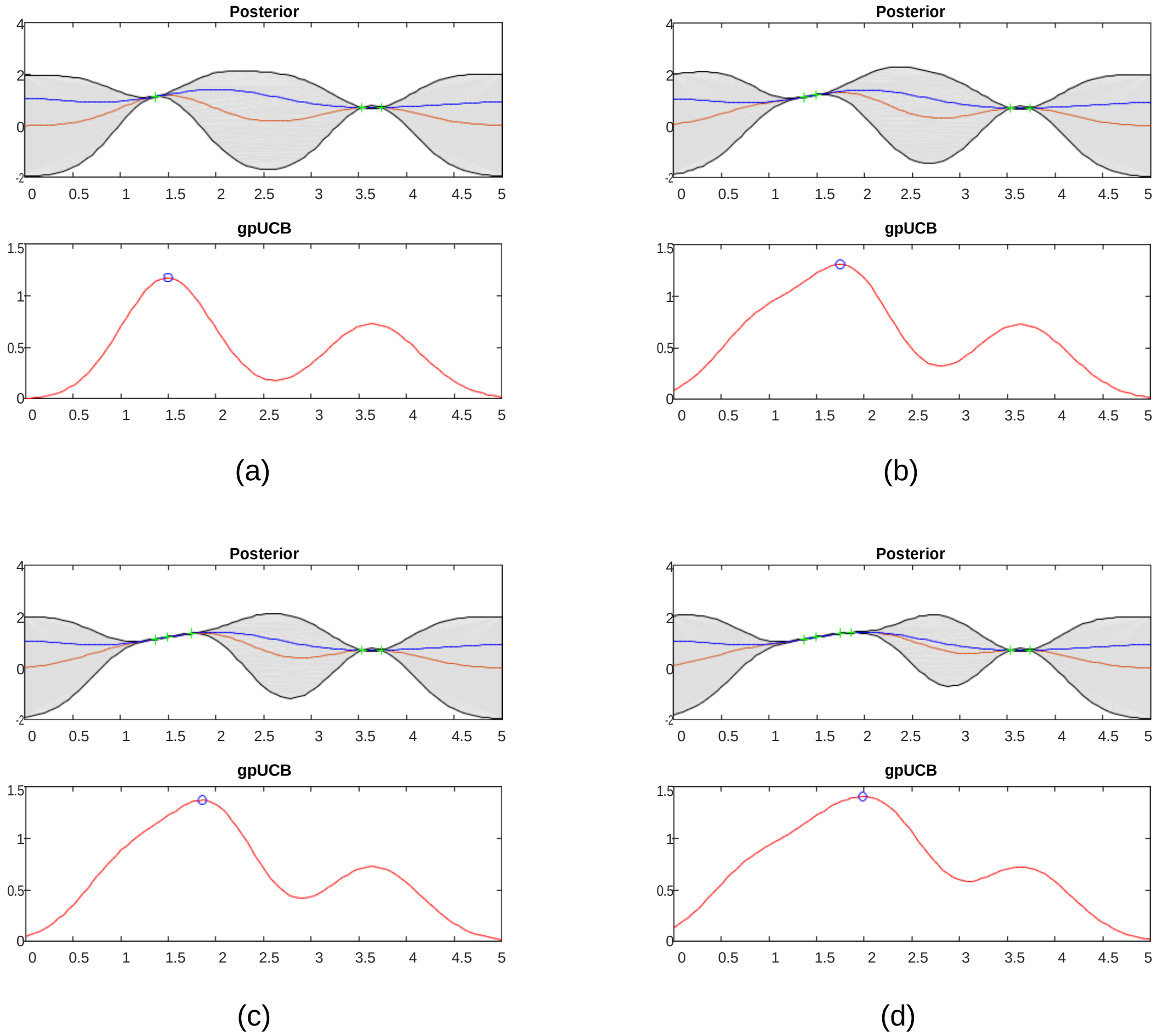

Figure 3.5: Illustration of using GP-UCB as acquisition function in Bayesian optimization on a toy 1*d* problem. The GP posterior is shown at the top as a solid red line. The dotted blue line is the true function and green crosses are current observations. The lower image shows the acquisition functions for that GP. The maximum of the acquisition function is shown with a blue circle marker.

### 3.3.4 Entropy search method

One view of Bayesian optimization is that at each step we are gathering information about where the maximum of the function might be. The entropy search (ES) strategy [Hennig and Schuler2012] quantifies this by considering the posterior distribution over the location of the maximum $p(\boldsymbol{x}^*)$. This distribution is captured implicitly by modeling the distribution over function values. The idea behind entropy search is to choose the point that most reduces our uncertainty over the location of the maximum. We can write the acquisition function as

$$\alpha_{ES}(\boldsymbol{x}) = \mathrm{H}[p(\boldsymbol{x}^* \mid D_t)] - \mathrm{E}_{p(y_{t+1}|D_t,\boldsymbol{x}_{t+1})}\{\mathrm{H}[p(\boldsymbol{x}^* \mid D_t \cup \{\boldsymbol{x}_{t+1}, y_{t+1}\})]\} \tag{3.22}$$

where $p(y_{t+1} \mid D_t, \boldsymbol{x}_{t+1})$ is the probability of observing $y_{t+1}$ at $\boldsymbol{x}_{t+1}$ under this model and $\mathrm{H}[p] = -\int_{\boldsymbol{x}^*} p(\boldsymbol{x}^*) \log p(\boldsymbol{x}^*) d\boldsymbol{x}^*$ is the differential entropy of distribution $p$. That is to say, entropy search seeks to choose the point $\boldsymbol{x}$ that maximizes the information gain in terms of the distribution over the location of the maximum. It does this by first computing the entropy of $p(\boldsymbol{x}^* \mid D_t)$ given the current observations, and then computing the expected entropy after incorporating $\boldsymbol{x}_{t+1}$ into the set of observations.

For continuous spaces, computing the entropy can be difficult. One simple strategy is to only consider the entropy over a discrete grid of points. A useful heuristic is to choose this grid to be local maximizers of expected improvement since these points are likely to contain information about the location of the optimal of the function, especially towards the end of the search.

### 3.3.5 Thompson sampling

Thompson sampling (TS) [Thompson1933], also known as probability matching, is a randomized acquisition strategy in which the probability that $\boldsymbol{x}$ is chosen for the next evaluation is proportional to the probability (according to the model) that $\boldsymbol{x}$ is the utility-maximizing action. However, instead of marginalizing out the uncertainty, TS works by randomly sampling the unknown quantities and then maximizing the utility given this sample. There are two ways to draw a Thompson sample 1) in the spatial domain on a set of fixed grid using the fact that the function values at those grid points and the function values at the observation set follows a multivariate Gaussian distribution, 2) in the frequency domain using a finite set of basis functions, often using random Fourier features. More specifically, given a shift-invariant kernel $k$, e.g. squared exponential kernel, Bochner's theorem [Bochner1959] states that it has a Fourier dual $s(\mathbf{s})$, which is equal to the spectral density of $k$. By normalizing the density as $p(\mathbf{s}) = s(\mathbf{s})/\beta$, the expectation of SE kernel can be written as

$$k(\boldsymbol{x}, \boldsymbol{x}^0) = \beta \mathrm{E}_{p(\mathbf{s})}[e^{-i\mathbf{s}^T(\boldsymbol{x}-\boldsymbol{x}_0)}] = 2\beta \mathrm{E}_{p(\mathbf{s},b)}[\cos(\mathbf{s}^T\boldsymbol{x} + b)\cos(\mathbf{s}^T\boldsymbol{x}^0 + b)], \tag{3.23}$$

where $b \sim U[0, 2\pi]$. Let us draw $L$ random samples of $p(\mathbf{s}, b)$ to generate stacked samples $\mathbf{S}$ and $\mathbf{b}$, and we let $\phi(\boldsymbol{x}) = \sqrt{2\beta/L}\cos(\mathbf{S}\boldsymbol{x} + \mathbf{b})$. Then the kernel $k$ can be approximated by the inner product of these Fourier features as

$$k(\boldsymbol{x}, \boldsymbol{x}') \approx \phi(\boldsymbol{x})^T \phi(\boldsymbol{x}'). \tag{3.24}$$

We can use this feature mapping $\phi(\boldsymbol{x})$ to approximate the Gaussian process prior for $f$ with a linear model $f(\boldsymbol{x}) = \phi(\boldsymbol{x})^T \boldsymbol{\theta}$ where $\boldsymbol{\theta} \sim \mathcal{N}(\mathbf{0}, \mathbf{I})$ is a normal distribution. In the context of Bayesian optimization, TS corresponds to sampling the objective function from the GP posterior and then maximizing this sample. Since the posterior for $\boldsymbol{\theta}$ is also a multivariate Gaussian by conditioning on $D_t$. Therefore, we have

$$\boldsymbol{\theta} \mid D_{1:t} \sim \mathcal{N}(\mathbf{A}^{-1}\mathbf{\Phi}\boldsymbol{y}_{1:t}, \sigma_n^2\mathbf{A}^{-1}), \tag{3.25}$$

where $\mathbf{A} = \mathbf{\Phi}\mathbf{\Phi}^T + \sigma_n^2\mathbf{I}$ and $\mathbf{\Phi} = [\phi(\boldsymbol{x}_1), \ldots, \phi(\boldsymbol{x}_t)]$. The predictive mean can be viewed as an approximation of the objective function as

$$f(\boldsymbol{x}_{t+1}) \approx \phi(\boldsymbol{x}_{t+1})^T \mathbf{A}^{-1}\mathbf{\Phi}\boldsymbol{y}_{1:t}. \tag{3.26}$$

We can use a global optimization routine to obtain the $\boldsymbol{x}^*$ from $f(\boldsymbol{x}_{t+1})$, which is approximately distributed according to $p(\boldsymbol{x}^* \mid D_{1:t})$. Thompson sampling has been shown to work well in practice [Chapelle and Li2011, Hernández *et al.*2014] and it is rather simple to implement.

### 3.3.6 Optimizing the acquisition function

So far we have discussed various acquisition functions that are typically used in Bayesian optimization, however each of these strategies themselves must be optimized in order to select a new evaluation point. In fact, theoretical convergence results currently require that these acquisition functions need to be globally optimized [Bull2011]. Since these acquisition functions yield non-convex shapes, this can be a difficult task. In practice, Bayesian optimization systems typically use some form of local search [Hutter *et al.*2011a] or other global optimization techniques, such as CMA-ES [Hansen2006], and DIRECT [Finkel2003]. We use DIRECT throughout this thesis.

Now we can present the Generic Bayesian optimization algorithm in Algorithm 3.1:

**Algorithm 3.1** Generic Bayesian Optimization

1: **for** $n = 1, 2, ...t$ **do**
2: Find $\boldsymbol{x}_{t+1}$ by optimizing the acquisition function over the GP:
$\boldsymbol{x}_{t+1} = \text{argmax}_{\boldsymbol{x}} \alpha(\boldsymbol{x}|D_t)$
3: Evaluate the function value: $y_{t+1}$
4: Augment the observation set $D_t = D_t \cup (\boldsymbol{x}_{t+1}, y_{t+1})$and update the GP.
5: **end for**

All the acquisition functions can be used for any kind of probabilistic models, for example a Gaussian process with derivative observations and a sparse Gaussian process which we will introduce in the following sections.

## 3.4 Gaussian process with derivative observations

Since derivative is a linear operator, so the derivative of a Gaussian process is also a GP. Therefore, GP will also work on making predictions about derivatives and also to make inference based on derivative information.

First of all, a covariance function $k(\cdot, \cdot)$ on function values implies the following (mixed) covariance between function values and partial derivatives [Riihimäki and Vehtari2010]

$$cov\left(f^i, \frac{\partial f^j}{\partial \boldsymbol{x}_d^{(j)}}\right) = \frac{\partial k(\boldsymbol{x}^{(i)}, \boldsymbol{x}^{(j)})}{\partial \boldsymbol{x}_d^{(j)}}, \tag{3.27}$$

where the superscript $i$ indicates the $i$th function observation and $j$ indicates the $j$th derivative observation. The subscript $d$ indicates the $d$th dimension of the observation. Similarly, the covariance between partial derivatives can be given as

$$cov\left(\frac{\partial f^i}{\partial \boldsymbol{x}_d^{(i)}}, \frac{\partial f^j}{\partial \boldsymbol{x}_g^{(j)}}\right) = \frac{\partial^2 k(\boldsymbol{x}^{(i)}, \boldsymbol{x}^{(j)})}{\partial \boldsymbol{x}_d^{(i)} \partial \boldsymbol{x}_g^{(j)}}. \tag{3.28}$$

In terms of Squared Exponential covariance function, the covariance between func-tion values and partial derivatives can be written as

$$cov(f^{i}, \frac{\partial f^{j}}{\partial x_g^{(j)}}) = \sigma_f^2 \exp(-\frac{1}{2}\sum_{d=1}^{D} \rho_d^{-2}(x_d^{(i)} - x_d^{(j)})^2) \times (\rho_g^{-2}(x_g^{(i)} - x_g^{(j)})), \quad (3.29)$$

where $\boldsymbol{\rho} = \{\rho_1, ...\rho_d\}$ is the lengthscale of the GP model.

The covariance between partial derivatives can be written as

$$cov(\frac{\partial f^{i}}{\partial x_g^{(i)}}, \frac{\partial f^{j}}{\partial x_h^{(j)}}) = \sigma_f^2 \exp(-\frac{1}{2}\sum_{d=1}^{D} \rho_d^{-2}(x_d^{(i)} - x_d^{(j)})^2) \times \rho_g^{-2}(\delta_{gh} - \rho_h^{-2}(x_h^{(i)} - x_h^{(j)})(x_g^{(i)} - x_g^{(j)})), \quad (3.30)$$

where $\delta_{gh} = 1$ if $g = h$, and $\delta_{gh} = 0$ if $g \neq h$.

We derive the joint distribution over a function and its derivatives which is also GP. The joint set of random variables $[f(\mathbf{x}), \nabla f(\mathbf{x})]$ follow a multi-variate Gaussian distribution as

$$\begin{bmatrix} f \\ \nabla f \end{bmatrix} \sim \mathcal{N}\left(0, \begin{bmatrix} k_{[f,f]} & k_{[f,\nabla f]} \\ k_{[\nabla f,f]} & k_{[\nabla f,\nabla f]} \end{bmatrix}\right), \quad (3.31)$$

where the first block is the covariance matrix for the function observation, $k_{[f,\nabla f]} = J(\mathbf{x}, \mathbf{x}') = (\frac{\partial k(\mathbf{x},\mathbf{x}')}{\partial x_1'}, \cdots, \frac{\partial k(\mathbf{x},\mathbf{x}')}{\partial x_d'})$ which is a Jacobian matrix and $k_{[\nabla f,\nabla f]} = H(\mathbf{x}, \mathbf{x}')$ which is a $d \times d$ Hessian of $k(\mathbf{x}, \mathbf{x}')$.

Now we can derive the predictive distribution for $f_{t+1}$ by given a set of function observations and its derivative observations. We use $\bar{\mathbf{K}}$ to denote the joint covariance matrix over a set of observations of function values and gradients as above. Then the new joint distribution for $[\mathbf{f}_{1:t}, \nabla \mathbf{f}_{1:t}, f_{t+1}]$ can be written as

$f_{1:t}$
$\nabla \boldsymbol{f}_{1:t}$
$f_{t+1}$

, (3.32)

where $\bar{\mathbf{k}} = [\mathbf{k}^T_{[f_{1:t}, \nabla f]}, \mathbf{k}^T_{[f_{1:t}, \nabla f]}]^T$, $\nabla f_{1:t} = \{\nabla f(\mathbf{x}_i)\}_{i=1}^{t}$, and

$$\bar{\mathbf{K}} = \begin{bmatrix} \mathbf{K} & \mathbf{k}_{[f_{1:t}, \nabla f_{1:t}]} \\ \mathbf{k}_{[\nabla f_{1:t}, f_{1:t}]} & \mathbf{k}_{[\nabla f_{1:t}, \nabla f_{1:t}]} \end{bmatrix}. \tag{3.33}$$

By following Eq.(3.9) and Eq.(3.10), the predictive mean $\bar{\mu}(\mathbf{x}_{t+1})$ and variance $\bar{\sigma}^2(\mathbf{x}_{t+1})$ can be derived as

$$\bar{\mu}(\mathbf{x}_{t+1}) = \bar{\mathbf{k}}^T \bar{\mathbf{K}}^{-1} [\mathbf{f}_{1:t}, \nabla \mathbf{f}_{1:t}]^T, \tag{3.34}$$

$$\bar{\sigma}^2(\mathbf{x}_{t+1}) = k(\mathbf{x}_{t+1}, \mathbf{x}_{t+1}) - \bar{\mathbf{k}}^T \bar{\mathbf{K}}^{-1} \bar{\mathbf{k}}. \tag{3.35}$$

**Gaussian process with derivative observations** In Figure 3.6 we demonstrate how the posterior mean and variance are influenced by using derivative observations. We can see from Figure 3.6b that by adding one derivative observations, the posterior mean and variance are more close to the true function than that in Figure 3.6a. We then add another derivative observation in our dataset and find the approximation is further improved in Figure 3.6c.

We also examine the behavior of GP when the locations of derivatives are same as function observations. We receive similar result with the previous setting and demonstrate them in Figure 3.7.

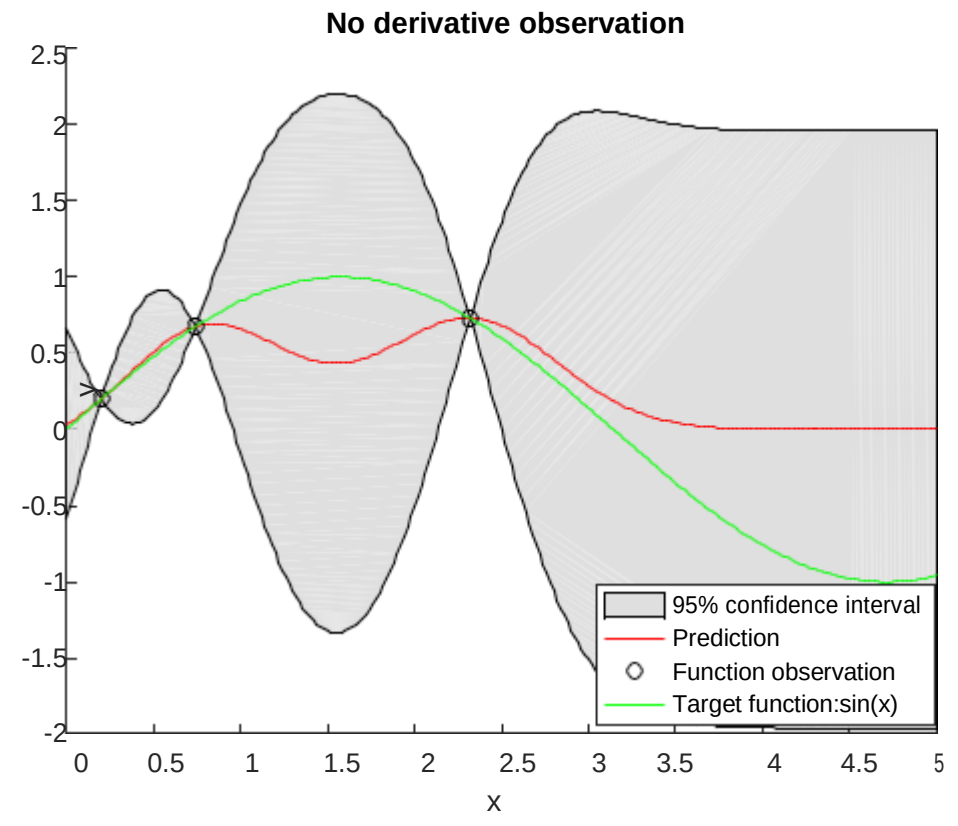

(a)

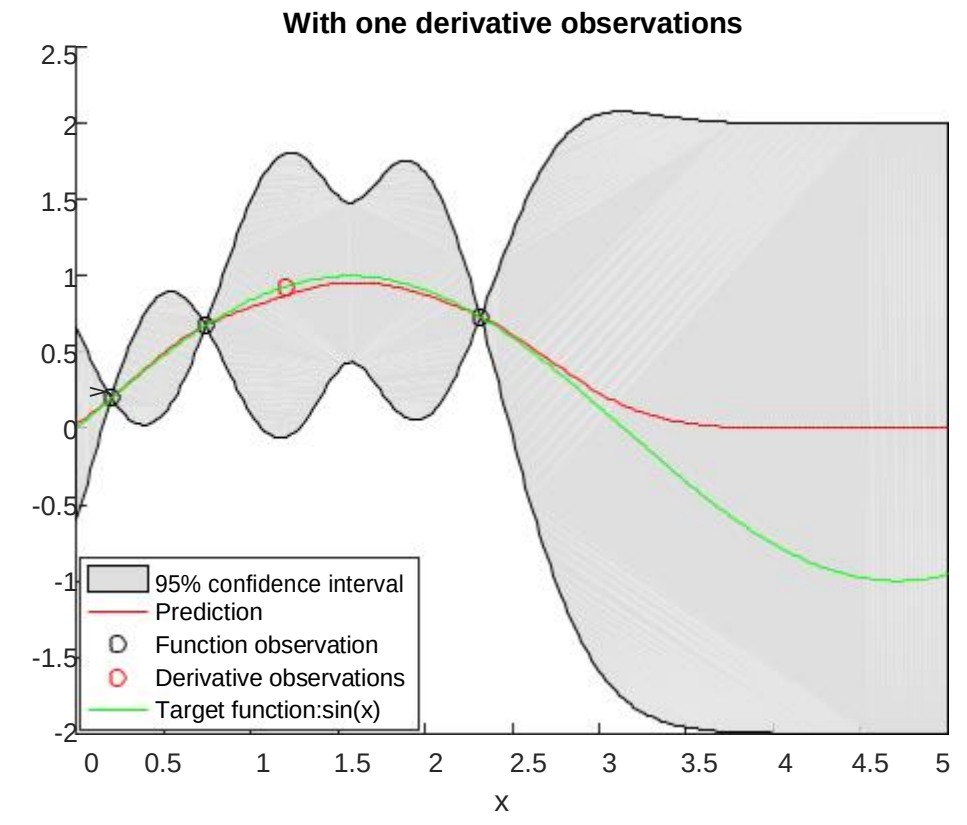

(b)

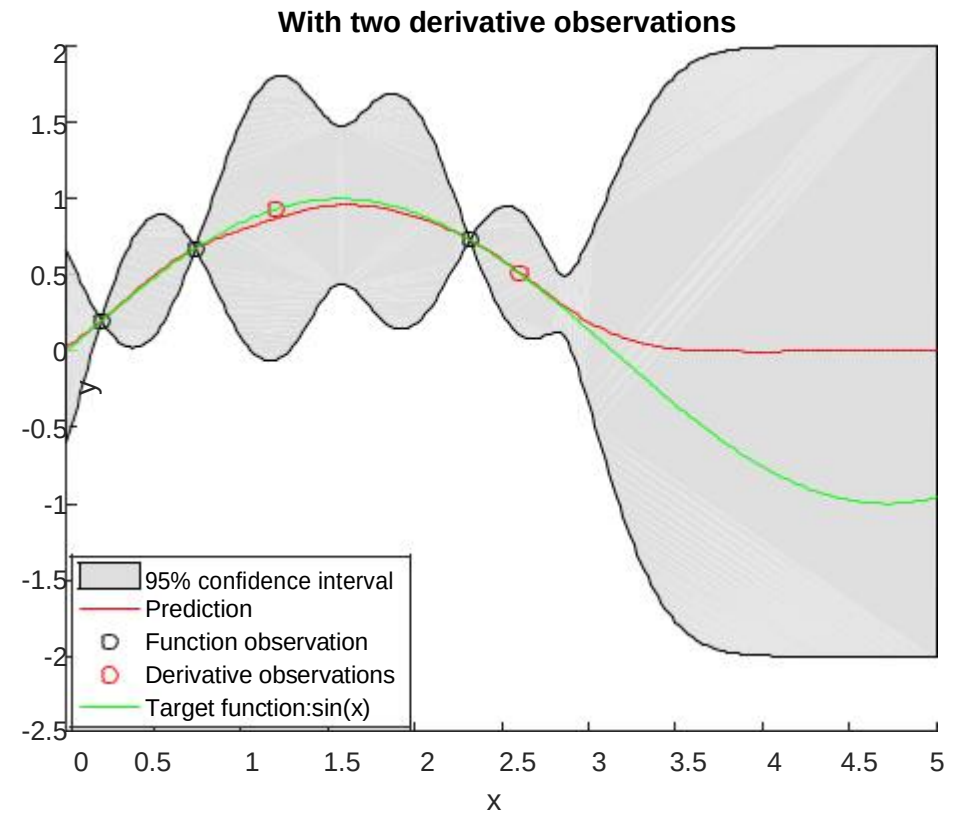

(c)

Figure 3.6: Posterior mean and variance of a GP (a) without derivative observation, (b) with one derivative observation and (c) two derivative observations, The green solid line indicates the true function of *sin*(***x***) and the red solid line is the posterior mean function of the GP. The gray area demonstrates the 95% confidence interval.

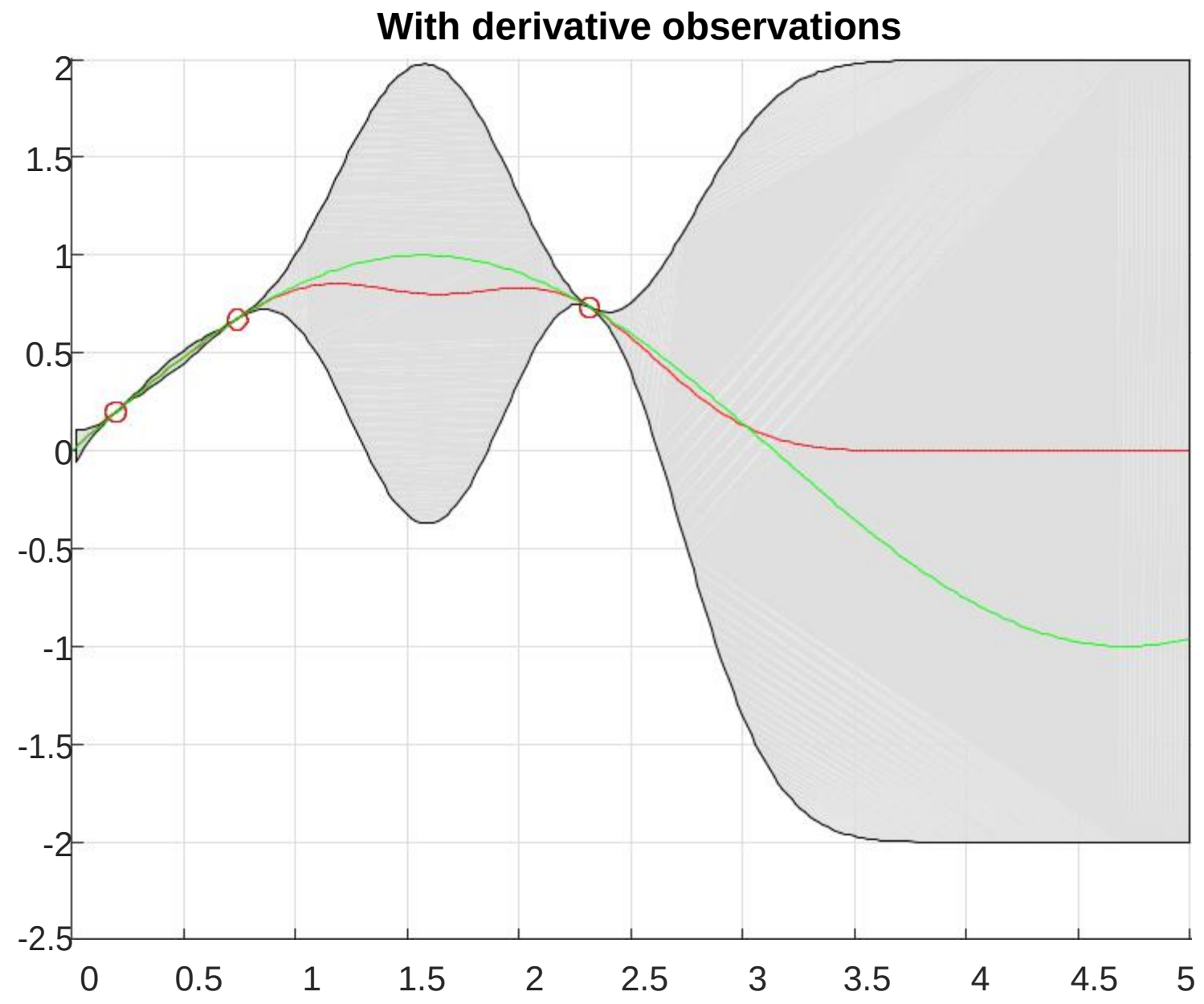

Figure 3.7: Location of derivative observations are same as function observations.

**Bayesian optimization with derivative observations** Sine we use GP as prior in Bayesian optimization and the contribution of incorporating derivative observa-tions in GP has been shown to provide a better model for the objective function. Therefore, in the following, we show that use GP with actual derivative observations can accelerate the convergence of Bayesian optimization.

In Figures 3.8, we demonstrate how posterior and acquisition function change iter-atively in Bayesian optimization with and without derivative observations. We use three function observations and two derivative observations, and we use EI as our acquisition function. We can see that with the improved modeling in GP, acquisition function will suggest different location to evaluate in each iteration, and it samples more closer to global maximum ($x$ = 2.0) after 3 iterations. The results of using derivative locations same as our initial function observations are showing in Figure 3.9.

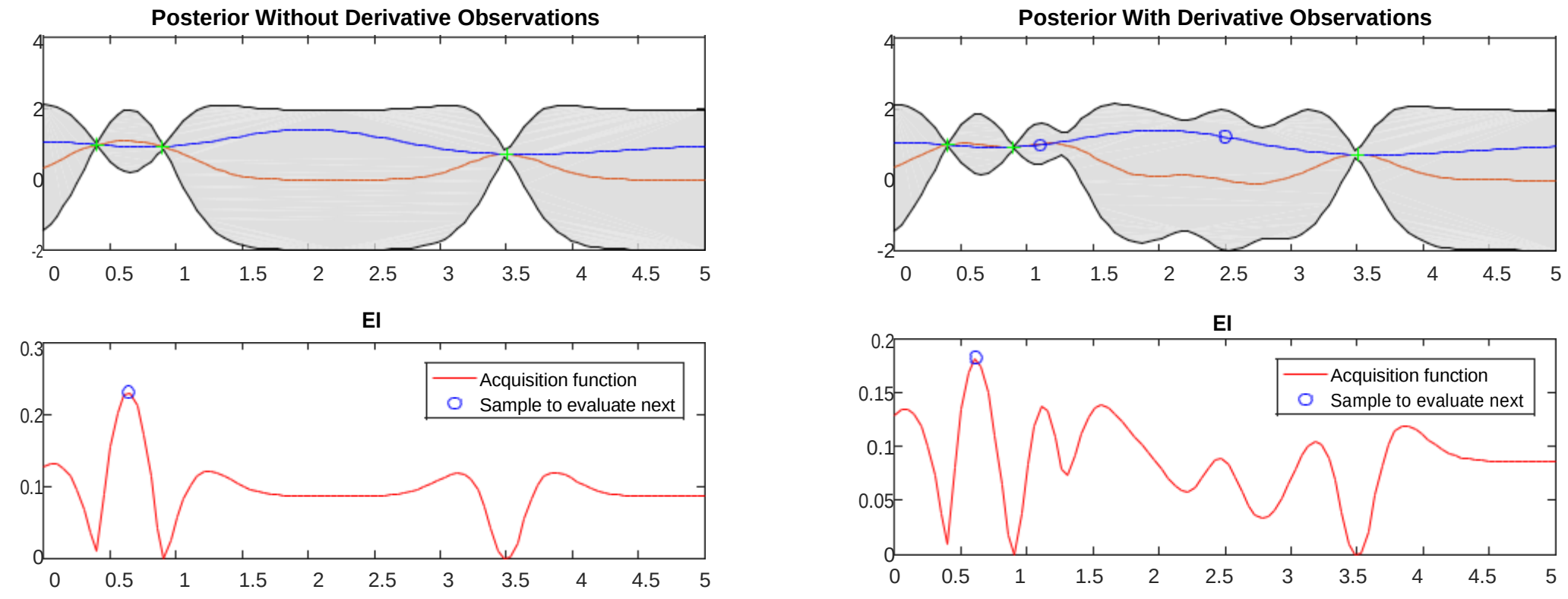

(a) Iteration 1

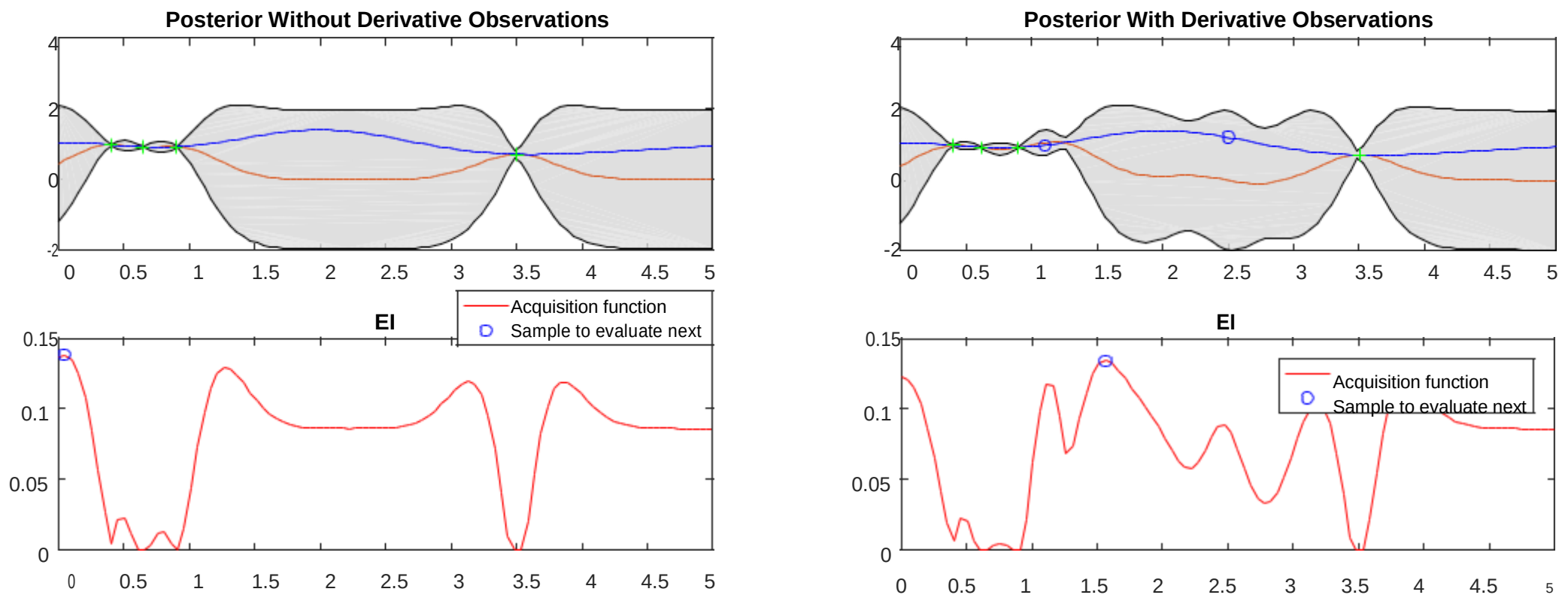

(b) Iteration 2

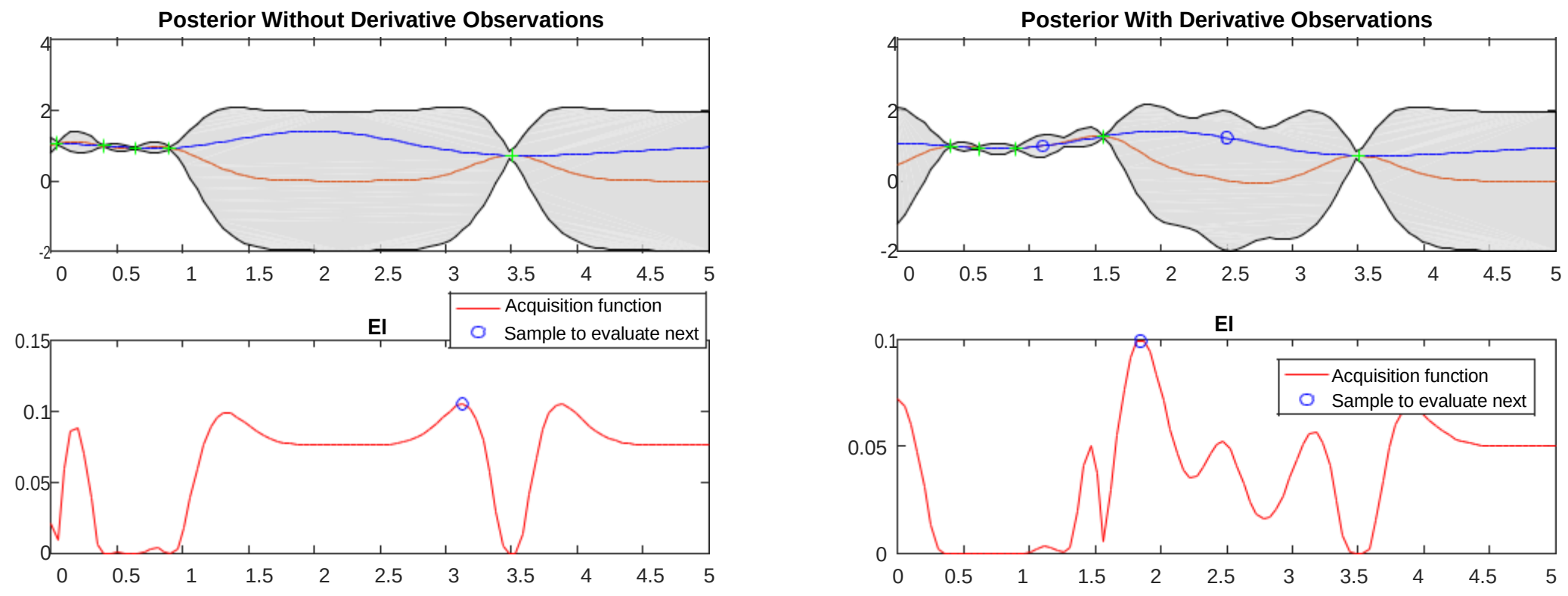

(c) Iteration 3

Figure 3.8: The comparison of the posterior distribution (top sub-figure) and acquisi-tion function (bottom sub-figure) between Bayesian optimization without derivative observations (left) and with derivative observations (right) in 3 iterations. The GP posterior is shown at the top sub-figure as a solid red line. The dotted blue line is the true function and green crosses are current observations. The blue circle indicates derivative observations. The bottom sub-figure shows the acquisition functions for that GP and the next sample which is determined by the acquisition function is shown with a blue circle marker.

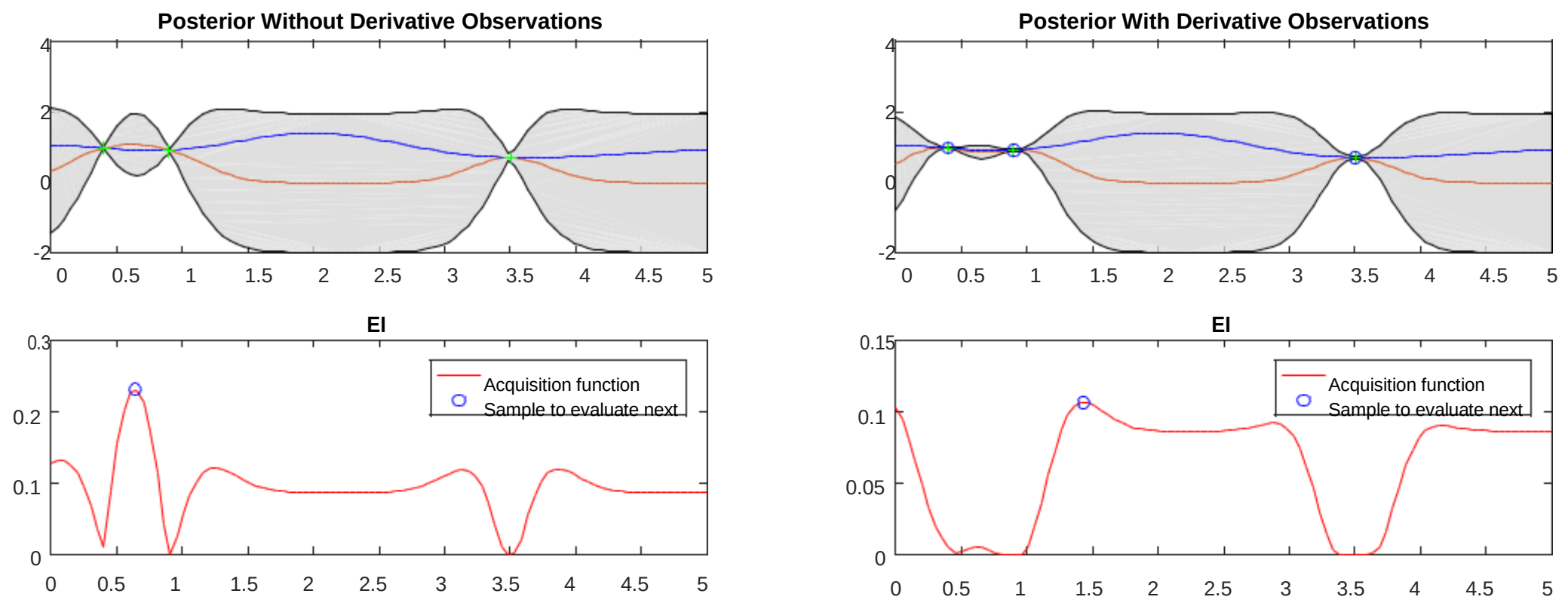

(a) Iteration 1

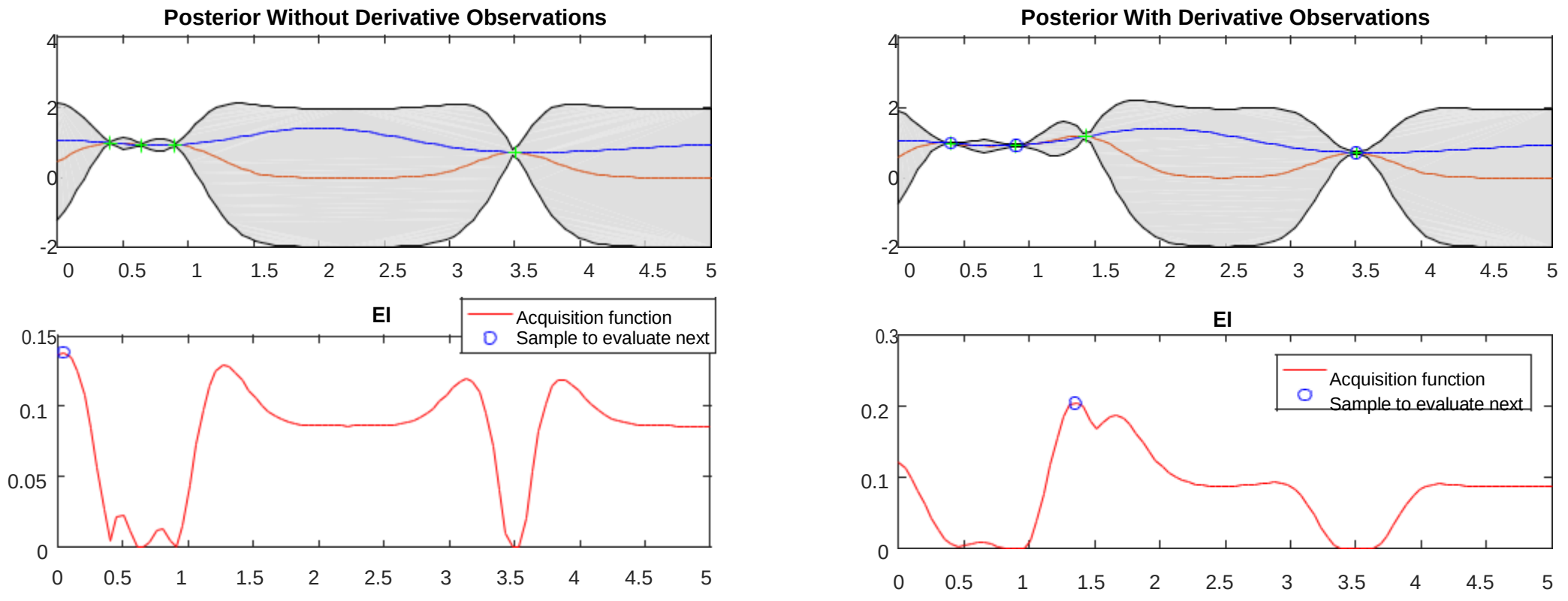

(b) Iteration 2

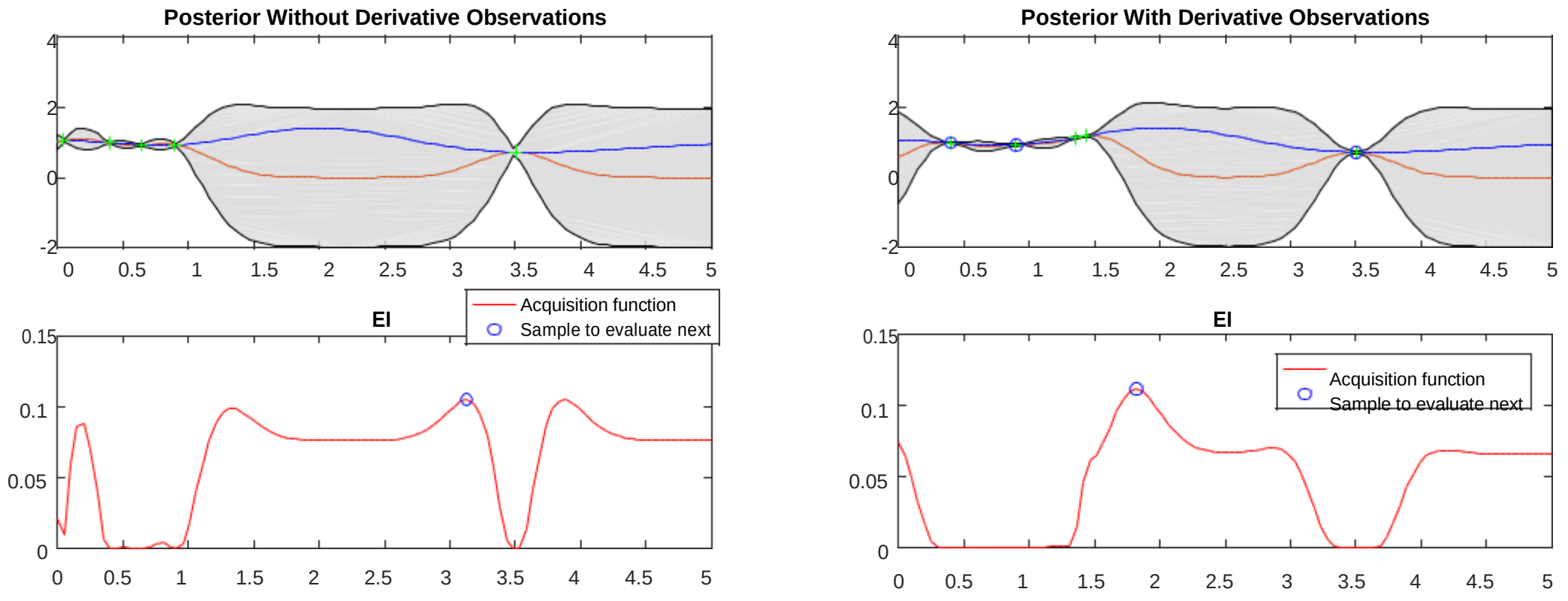

(c) Iteration 3

Figure 3.9: The comparison of the posterior distribution (top sub-figure) and acquisi-tion function (bottom sub-figure) between Bayesian optimization without derivative observations (left) and with derivative observations (right) in 3 iterations. The GP posterior is shown at the top sub-figure as a solid red line. The dotted blue line is the true function and green crosses are current observations. The blue circle indicates derivative observations. The bottom sub-figure shows the acquisition functions for that GP and the next sample which is determined by the acquisition function is shown with a blue circle marker.

The algorithm of Bayesian optimization with actual derivative observations is pre-sented in Algorithm 3.2:

---

**Algorithm 3.2** Bayesian Optimization with Actual Derivative Observations

---

1: **for** $n = 1, 2, ... t$ **do**

2: Build GP with function observations and derivatives of observations

3: Find $\boldsymbol{x}_{t+1}$ by optimizing the acquisition function:

$$\boldsymbol{x}_{t+1} = argmax_{\boldsymbol{x}} \alpha(\boldsymbol{x} | D_t)$$

4: Evaluate the objective function: $y_{t+1} = f(\boldsymbol{x}_{t+1}) + \xi$

5: Augment the observation set $D_t = D_t \cup (\boldsymbol{x}_{t+1}, y_{t+1})$.

6: **end for**

---

## 3.5 Sparse Gaussian process with fully independent conditionals

Gaussian process is a powerful non-parametric model, however, its posterior computation involves inversion of a matrix of the size number of observations. As the matrix inversion is cubic in terms of size, posterior computation becomes infeasible with number of observations more than a few thousands. To overcome the computational limitations, many researchers proposed to use sparse approximations. The common idea is to approximate the function with a much lower number of latent variables than what would be needed for a full GP. Some achieve it using a subset of most useful observations [Snelson and Ghahramani2006], while others do it by approximating the kernel function [Melkumyan and Ramos2009].

We will introduce the fully independent conditional (FIC) approximation [Quinonero-Candela and Rasmussen2005] in this section as it has been heavily used in our Chapter 5.

Many sparse Gaussian process models will introduce a set of inducing observations $D_u = \{\mathbf{M}_i, \mathbf{u}_i\}_{i=1}^m$ where $\mathbf{u} = \{\mathbf{u}_i\}_{i=1}^m$ contains values of the function at the points $\mathbf{M} = \{\mathbf{M}_i\}_{i=1}^m$ and $\mathbf{M}$ can be a subset of $\{\boldsymbol{x}_i\}_{i=1}^t$ which are know as inducing points.

Given prior distribution $p(\mathbf{u}) = \mathcal{N}(\mathbf{u} \mid \mathbf{0}, \boldsymbol{K}_{m,m})$, the training conditional distribu-tion of $\boldsymbol{f}_{1:t}$ given $\mathbf{u}$ can be written as

$$p(\boldsymbol{f}_{1:t} \mid \mathbf{u}) = \mathcal{N}(\boldsymbol{f}_{1:t} \mid \boldsymbol{K}_{t,m}\boldsymbol{K}_{m,m}^{-1}\mathbf{u}, \boldsymbol{K}_{t,t} - \boldsymbol{Q}_{t,t}), \tag{3.36}$$

where $\boldsymbol{K}_{t,m}$ is the covariance matrix between all observations and inducing points and $\boldsymbol{K}_{m,m}$ is the covariance matrix between all the inducing points. A matrix is also introduced as the shorthand notation $\boldsymbol{Q}_{a,b} = \boldsymbol{K}_{a,m}\boldsymbol{K}_{m,m}^{-1}\boldsymbol{K}_{m,b}$. Here, $\mathbf{u}$ plays the role of observations so that the posterior mean is written as $\boldsymbol{K}_{t,m}\boldsymbol{K}_{m,m}^{-1}\mathbf{u}$. The covariance matrix has the form of the $\boldsymbol{K}_{t,t}$ minus a non-negative definite matrix $\boldsymbol{Q}_{t,t}$ which gives the measurement of how much information that $\mathbf{u}$ provides in $\boldsymbol{f}_{1:t}$. The test conditional $p(f_{t+1} \mid \mathbf{u})$ is formed the same way with Eq(3.36) as

$$p(f_{t+1} \mid \mathbf{u}) = \mathcal{N}(f_{t+1} \mid \boldsymbol{K}_{t+1,m}\boldsymbol{K}_{m,m}^{-1}\mathbf{u}, \boldsymbol{K}_{t+1,t+1} - \boldsymbol{Q}_{t+1,t+1}), \tag{3.37}$$

where $f_{t+1}$ is a test point and $\boldsymbol{K}_{t+1,m}$ is the covariance matrix between the test point and inducing points. To recover $p(\boldsymbol{f}_{1:t}, f_{t+1})$ we can simply integrate out $\mathbf{u}$ from the joint GP prior $p(\boldsymbol{f}_{1:t}, f_{t+1}, \mathbf{u})$ as

$$p(\boldsymbol{f}_{1:t}, f_{t+1}) = \int p(\boldsymbol{f}_{1:t}, f_{t+1}, \mathbf{u})\, d\mathbf{u} = \int p(\boldsymbol{f}_{1:t}, f_{t+1} \mid \mathbf{u})p(\mathbf{u})d\mathbf{u}. \tag{3.38}$$

In the first stage of approximation, we can make the assumption that $\boldsymbol{f}_{1:t}$ and $f_{t+1}$ are conditionally independent given $\mathbf{u}$. So the joint distribution of them can be written as

$$p(\boldsymbol{f}_{1:t}, f_{t+1}) \simeq q(\boldsymbol{f}_{1:t}, f_{t+1}) = \int q(\boldsymbol{f}_{1:t} \mid \mathbf{u})q(f_{t+1} \mid \mathbf{u})p(\mathbf{u})d\mathbf{u}, \tag{3.39}$$

and this stage is general to all the prior approximation methods which are making use of inducing points. The specific sparse approximation is then derived by making additional assumptions on the $q(\boldsymbol{f}_{1:t} \mid \mathbf{u})$ and $q(f_{t+1} \mid \mathbf{u})$ (e.g. SoR, FITC and FIC as introduced in Chapter 2).

Here, we assume training points and a test point are fully independent on $\mathbf{u}$, so we can derive the $q(\boldsymbol{f}_{1:t} \mid \mathbf{u})$ under the FIC as

$$q(\boldsymbol{f}_{1:t} \mid \mathbf{u}) = \mathcal{N}(\boldsymbol{K}_{t,m}\boldsymbol{K}_{m,m}^{-1}\mathbf{u}, diag[\boldsymbol{K}_{t,t} - \boldsymbol{Q}_{t,t}]), \tag{3.40}$$

where $diag[\boldsymbol{K}_{t,t} - \boldsymbol{Q}_{t,t}]$ is a diagonal matrix whose elements match the diagonal of $\mathrm{K}_{t,t} - \boldsymbol{Q}_{t,t}$. With this assumption, we only keep the variance information of function observations themselves and ignore covariance between function observations and inducing points. Likewise, we can derive the $q(f_{t+1} \mid \mathbf{u})$ as

$$q(f_{t+1} \mid \mathbf{u}) = \mathrm{N}\,(\boldsymbol{K}_{t+1,m}\boldsymbol{K}_{m,m}^{-1}\mathbf{u},\ diag[\boldsymbol{K}_{t+1,t+1} - \boldsymbol{Q}_{t+1,t+1}]). \tag{3.41}$$

Then we can form the joint distribution of function observations and the test point as

$$q(\boldsymbol{f}_{1:t}, f_{t+1}) = \mathrm{N}\left(0, \begin{bmatrix} \boldsymbol{Q}_{t,t} + diag[\boldsymbol{K}_{t,t} - \boldsymbol{Q}_{t,t}] & \boldsymbol{Q}_{t,t+1} \\ \boldsymbol{Q}_{t+1,t} & \boldsymbol{Q}_{t+1,t+1} + diag[\boldsymbol{K}_{t+1,t+1} - \boldsymbol{Q}_{t+1,t+1}] \end{bmatrix}\right), \tag{3.42}$$

and the posterior mean and variance as

$$\mu(\boldsymbol{x}_{t+1}) = \boldsymbol{K}_{t+1,m}\boldsymbol{\Sigma}\boldsymbol{K}_{m,t}\Lambda^{-1}\boldsymbol{y}, \tag{3.43}$$

$$\sigma^2(\boldsymbol{x}_{t+1}) = diag[\boldsymbol{K}_{t+1,t+1} - \boldsymbol{Q}_{t+1,t+1}] + \boldsymbol{K}_{t+1,m}\boldsymbol{\Sigma}\boldsymbol{K}_{m,t+1}. \tag{3.44}$$

where $\boldsymbol{\Sigma} = (\boldsymbol{K}_{m,m} + \boldsymbol{K}_{m,t}\Lambda^{-1}\boldsymbol{K}_{t,m})^{-1}$ and $\Lambda = diag[\boldsymbol{K}_{t,t} - \boldsymbol{Q}_{t,t} + \sigma_n^2\mathbf{I}]$.

**Complexity analysis** In full GP with derivatives, the computational complexity for training is $O(t)^3$, where $t$ is the number of function observations The FIC method which introduced a set of inducing points $\mathbf{u}$, bring the complexity down to $O(tm^2)$, where $m$ is the number of inducing points and $m$ $t$.

## 3.6 Sparse spectrum Gaussian process

Traditional sparse Gaussian process approximations such as FIC introduce induc-ing points to approximate the posterior mean and variance of full GP whilst sparse spectrum Gaussian process (SSGP) [Lazaro Gredilla *et al.*2010] uses a sparse set of frequencies to approximate the kernel function in spectrum domain. Briefly, accord-ing to the Bochner's theorem [Bochner1959], any stationary covariance function can be represented as the Fourier transform of some finite measure $\sigma_f^2 p(\mathbf{s})$ with $p(\mathbf{s})$ a probability density

$$k(\boldsymbol{x}_i, \boldsymbol{x}_j) = \int_{\mathbb{R}^D} e^{2\pi i \mathbf{s}^T (\boldsymbol{x}_i - \boldsymbol{x}_j)} \sigma_f^2 p(\mathbf{s}) d\mathbf{s}. \tag{3.45}$$

The frequency vector $\mathbf{s}$ has the same length $D$ as the input vector $\boldsymbol{x}$. In other words, a spectral density entirely determines the properties of a stationary kernel. Furthermore, Eq.(6.1) can be computed and approximated as

$$k(\boldsymbol{x}_i, \boldsymbol{x}_j) = \sigma_f^2 \mathbb{E}_{p(\mathbf{s})} \left[ e^{2\pi i \mathbf{s}^T \boldsymbol{x}_i} (e^{2\pi i \mathbf{s}^T \boldsymbol{x}_j})^* \right]$$

$$\approx \frac{\sigma_f^2}{m} \sum_{r=1}^{m} \cos \left[ 2\pi \mathbf{s}_r^T (\boldsymbol{x}_i - \boldsymbol{x}_j) \right] \tag{3.46}$$

$$= \frac{\sigma_f^2}{m} \varphi(\boldsymbol{x}_i)^T \varphi(\boldsymbol{x}_j). \tag{3.47}$$

The Eq.(3.46) can be approximated using Monte Carlo approach with symmetric sets $\{\mathbf{s}_r, -\mathbf{s}_r\}_{r=1}^m$ sampled from $\mathbf{s}_r \sim p(\mathbf{s})$. $m$ is the number of spectral frequencies (features). The Eq.(6.4) can be obtained with the setting

$$\varphi(\boldsymbol{x}) = [\cos(2\pi \mathbf{s}_1^T \boldsymbol{x}), \sin(2\pi \mathbf{s}_1^T \boldsymbol{x}), \cdots, \cos(2\pi \mathbf{s}_m^T \boldsymbol{x}), \sin(2\pi \mathbf{s}_m^T \boldsymbol{x})]^T, \tag{3.48}$$

which is a column vector of length $2m$ containing the evaluation of the $m$ pairs of trigonometric functions at $\boldsymbol{x}$. Then by following the formulas of generic GP as in Eq.(3.9) and Eq.(3.10), it is straightforward to derive the posterior mean and variance of sparse spectrum Gaussian process as

$$\mu(\boldsymbol{x}_{t+1}) = \varphi(\boldsymbol{x}_{t+1})^T \mathbf{A}^{-1} \mathbf{\Phi} \boldsymbol{y} \tag{3.49}$$

$$\sigma^2(\boldsymbol{x}_{t+1}) = \sigma_n^2 + \sigma_n^2 \varphi(\boldsymbol{x}_{t+1})^T \mathbf{A}^{-1} \varphi(\boldsymbol{x}_{t+1}), \tag{3.50}$$

where $\mathbf{\Phi} = [\varphi(\boldsymbol{x}_1), \ldots, \varphi(\boldsymbol{x}_t)]$ is a $2m \times t$ matrix and $\mathbf{A} = \mathbf{\Phi}\mathbf{\Phi}^T + \frac{m\sigma_n^2}{\sigma_f^2} \mathbf{I}_{2m}$. To obtain the $m$ optimal features, we can maximize the log marginal likelihood $\log p(\boldsymbol{y}|\Theta)$, where we collapse all hyperparameters in the kernel function and spectrum features into $\Theta$ and we obtain

$$\log p(\boldsymbol{Y}\,|\Theta) = -\frac{1}{2\sigma_n^2}[\boldsymbol{y}^T \boldsymbol{y} - \boldsymbol{y}^T \boldsymbol{\Phi}^T \mathbf{A}^{-1}\boldsymbol{\Phi}\boldsymbol{y}]$$

$$-\frac{1}{2}\log|\mathbf{A}| + m\log\frac{m\sigma_n^2}{\sigma_f^2} - \frac{t}{2}\log 2\pi\sigma_n^2. \qquad (3.51)$$

The SSGP, therefore, uses $m$ optimal features to approximate full GP, which again brings the computational complexity down to O($tm^2$), and $m$ $t$.

It is noticed that the sparse spectrum Gaussian process and Thompson sampling are both use random Fourier features in their formulations. Therefore, we can directly observe how the $p(\boldsymbol{x}^*)$ changes along with applying different Fourier features. This is the key reason why we choose SSGP as our base sparse method in Chapter 6. Currently, we can not relate the sparse sets from other time domain sparse models such as SoR and FI(T)C to $p(\boldsymbol{x}^*)$ due to insufficient research in this area.

## 3.7 Summary

This chapter provided a brief overview of the mathematical background of the topics discussed and applied in this thesis. It covered Bayesian optimization, Gaussian process, and extended Gaussian process such as Gaussian process with derivative observations and scalable Gaussian process models.

# Chapter 4

# Efficient Bayesian Optimization Using Derivative Meta-model

Most real world functions, such as physical experiments or hyperparameter tuning, are well behaved. That is to say, they are smooth and have a small number of local peaks. If such knowledge can be harnessed, Bayesian optimization may converge even faster. Bayesian optimization algorithms for well-behaved functions have been addressed only in limited contexts when either the function is monotonic [Riihimäki and Vehtari2010, Lin and Dunson2014] or it has a concave/convex shape [Wang and Berger2016, Jauch and Peña2016]. However, Bayesian optimization for functions with more general shape properties such as incorporating the knowledge that the function has only a few peaks has not been addressed before, and thus remains an open problem.

To fill this gap, in this chapter, we propose a new method that can flexibly incorpo-rate the shape of the target function. The shape information can be guided by using a model that only admits certain kinds of shape. However, the information that the function only has few peaks is vague and cannot be modeled by using known func-tions. For example, a unimodal function may not be always quadratic. Thus, we use a combination of strategy where actual function is modeled by a flexible Squared Exponential (SE) kernel based Gaussian process, but it is regularized by synthetic observations from a shape-constrained polynomial model. The shape information is passed using derivative observations sampled from the polynomial model. We could have used a regular polynomial fitting approach, however Gaussian process with polynomial kernel (GPPK) provides a versatile framework, and hence we choose it as our meta-model. Specifically, we use GPPK as the surrogate for function shape from which we can extract derivative information. Next, we use both the sampled derivative values and the usual function observations for the Bayesian optimization. In effect, the main GP combing the sampled derivative values and the function observations, is built based on a trade-off between the flexible model induced by the stationery kernel and the structure induced by the derivative information based on GPPK. We refrain from using the samples of the function values from GPPK because we only want to pass the shape information through derivatives, while keeping the function values guided mostly by the main GP. Since the shape information is the key to success, therefore the crucial in this scheme is setting the degree of the polynomial kernel. We use a Bayesian formulation to estimate the degree from the observed data. Specifically, we use a truncated geometric prior, cut-off at degree of 10 and then normalize. We choose this kind of prior since the objective functions that we are targeting are smooth and only have few peaks. Therefore, we essen-tially prefer to use lower degrees as the prior information. The posterior of degree is then computed based on the product of the prior and the marginal likelihood of the GPPK on the observed data. The mode of the posterior is used as the estimated degree for our derivative meta-model.

We demonstrate our method on three synthetic examples and real world applica-tions on hyperparameter tuning for two machine learning algorithms 1) Support Vector Machines and 2) Elastic Net. We compare with Bayesian optimization with-out derivatives and Bayesian optimization with true derivatives in synthetic exam-ples and only compare Bayesian optimization without derivatives in hyperparameter tuning since true derivatives are not available in this case. In all experiments our proposed method outperforms the baselines.

In the following sections, we introduce our proposed method that incorporates shape information in Bayesian optimization through a derivative meta-model. Section 4.1 discusses the choice of derivative meta-models. Section 4.2 shows an efficient Bayesian optimization model with estimated derivative observations. Section 4.3 provides a derivation of a mechanism to estimate the parameter of the prior shape function through Bayesian inference. Section 4.4 examines the performance on both synthetic functions and real world applications of hyperparameter tuning and Sec-tion 4.5. concludes this chapter.

# 4.1 Meta-model

In the following, we elaborate on two different ways to build our meta-model for shape regularization.

## 4.1.1 Polynomial curve fitting

The polynomial function of a single indeterminate is defined as $g(\boldsymbol{x}) = w_0 + w_1\boldsymbol{x} + w_2\boldsymbol{x}^2 + .. + w_d\boldsymbol{x} = \sum_{i=0} w_i\boldsymbol{x}^i$, where is the degree of the polynomial and $w_i$ is the weight. We could have used the least squares approach where we sum the error over all observations. An example has been shown in Figure 4.1. However, we envisage that whilst simple polynomial models can easily be developed for low dimensional problems, they may not be suitable for high dimensional objective functions as the number of observations required to fit those models statistically grows exponentially. Hence, we propose to use Gaussian process with polynomial kernel as the meta-model for derivative estimation.

## 4.1.2 Gaussian process with polynomial kernel

Recall from the Chapter 3 that Gaussian process (GP) is specified by its mean function $\mu(\boldsymbol{x})$ and covariance function $k(\boldsymbol{x}, \boldsymbol{x}^0)$. A sample from a Gaussian process is a function given by

$$f(\boldsymbol{x}) \sim \mathcal{N}(\mu(\boldsymbol{x}), k(\boldsymbol{x}, \boldsymbol{x}^0)), \tag{4.1}$$

where $\mathcal{N}$ is a Gaussian distribution and $\boldsymbol{x}$ denotes a $d$-dimensional covariate vector. Without any loss in generality, the prior mean function can be assumed to be a zero function making the Gaussian process fully defined by the covariance function. Suppose we have a set of noise free observations $\mathcal{D} = \{\boldsymbol{x}_{1:t}, \boldsymbol{f}_{1:t}\}$, where $\boldsymbol{f}_{1:t} = \{f(\boldsymbol{x}_i)\}_{i=1}^{t}$. The joint distribution of observations $\{\boldsymbol{x}_{1:t}, \boldsymbol{f}_{1:t}\}$ and a new point

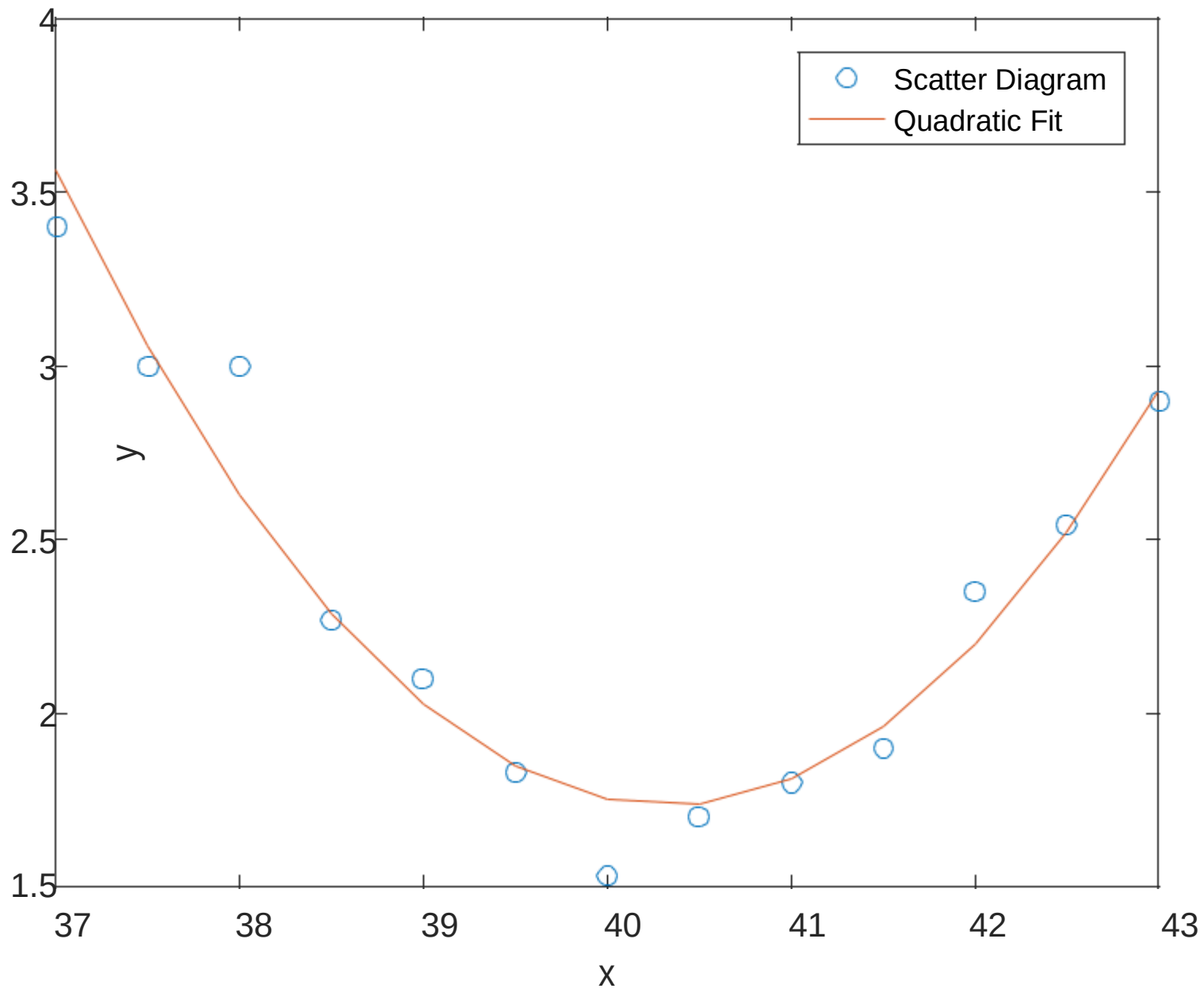

Figure 4.1: An example of polynomial curve fitting by using the least squares ap-proach

$\{\mathbf{x}_{t+1}, f_{t+1}\}$ in GP is presented as

$$\begin{bmatrix} \mathbf{f}_{1:t} \\ f_{t+1} \end{bmatrix} \sim \mathcal{N}\left(0, \begin{bmatrix} \mathbf{K} & \mathbf{k} \\ \mathbf{k} & k(\mathbf{x}_{t+1}, \mathbf{x}_{t+1}) \end{bmatrix}\right), \quad (4.2)$$

where $\mathbf{k} = [\, k(\mathbf{x}_{t+1}, \mathbf{x}_1),\ k(\mathbf{x}_{t+1}, \mathbf{x}_2),\ \ldots,\ k(\mathbf{x}_{t+1}, \mathbf{x}_t)\, ]$, and $\mathbf{K}$ is the Gram ma-trix given by

$$\mathbf{K} = \begin{bmatrix} k(\mathbf{x}_1, \mathbf{x}_1) & \ldots & k(\mathbf{x}_1, \mathbf{x}_t) \\ \vdots & \ddots & \vdots \\ k(\mathbf{x}_t, \mathbf{x}_1) & \ldots & k(\mathbf{x}_t, \mathbf{x}_t) \end{bmatrix}. \quad (4.3)$$

A Gaussian process allows us to compute the predictive mean and variance in closed form which is fully defined by the covariance function. A Gaussian process with polynomial kernel uses kernels of the form

$$k(\mathbf{x}, \mathbf{x}^0) = (\sigma_0^2 + \mathbf{x} \cdot \mathbf{x}^0)\ , \quad (4.4)$$

where is the degree of the polynomial and $\sigma_0{}^2$ is a free parameter trading off the influence of higher-order versus lower-order terms in the polynomial, and builds a Bayesian polynomial regression model. The posterior mean function of GPPK can be computed as

$$\mu(\boldsymbol{x}) = \mathbf{k}\mathbf{K}^{-1}\mathbf{y}, \tag{4.5}$$

where $\mathbf{k}$ =[ $(\sigma^2 + \boldsymbol{x} \cdot \boldsymbol{x}_1)$ , $(\sigma^2 + \boldsymbol{x} \cdot \boldsymbol{x}_2)$ , . . . , $(\sigma^2 + \boldsymbol{x} \cdot \boldsymbol{x}_t)$ ], and the covariance matrix $\mathbf{K}$ can be computed by substituting Eq.(4.4) into Eq.(4.3). The GPPK works the role of our derivative meta-model from which we then extract derivative information from.

### 4.1.3 Derivative estimation

Now we can estimate the derivative value at $\boldsymbol{x}$ by differentiating the posterior mean function Eq.(4.5) as

$$\nabla f^e = \frac{\partial}{\partial \boldsymbol{x}} \mu(\boldsymbol{x}) = \mathbf{k}^0\mathbf{K}^{-1}\mathbf{y}, \tag{4.6}$$

where $\nabla f^e$ denotes the derivative value, and $\mathbf{k}^0$=[ $(\sigma^2 + \boldsymbol{x} \cdot \boldsymbol{x}_1)^{-1} \cdot \boldsymbol{x}_1$, $(\sigma^2 + \boldsymbol{x} \cdot \boldsymbol{x}_2)^{-1} \cdot \boldsymbol{x}_2$, . . ., $(\sigma^2 + \boldsymbol{x} \cdot \boldsymbol{x}_t)^{-1} \cdot \boldsymbol{x}_t$].

## 4.2 BO with estimated derivatives

Once the estimated derivatives are available, we incorporate them into Bayesian optimization. Recall from the Chapter 3 that the derivatives of a Gaussian Process is still a GP, the joint distribution of function values and estimated derivatives is analytically tractable given as

$r_{1:t}$ $\sim N$

$$0, \mathbf{K}, \qquad (4.7)$$

where $\nabla \mathbf{f}^{e}_{1:t} = \{ \frac{\partial}{\partial \mathbf{x}_i} f(\mathbf{x}_i)\}_{i=1}^{t}$ and

$$\bar{\mathbf{K}} = \begin{bmatrix} \mathbf{K} & k_{[\mathbf{f}_{1:t}, \nabla \mathbf{f}^{e}_{1:t}]} \\ k_{[\nabla \mathbf{f}^{e}_{1:t}, \mathbf{f}_{1:t}]} & k_{[\nabla \mathbf{f}^{e}_{1:t}, \nabla \mathbf{f}^{e}_{1:t}]} \end{bmatrix}, \tag{4.8}$$

which denotes the joint covariance matrix between a set of observations of function values and the estimated derivatives and the $\mathbf{K}$ is defined in Eq.(4.3).

Now we can derive the posterior over a new function value $f_{t+1}$ at $\mathbf{x}_{t+1}$ when given a set of observations of the function values and a set of derivative information. The joint distribution for $[\mathbf{f}_{1:t}, \nabla \mathbf{f}^{e}_{1:t}, f_{t+1}]$ also follows a multi-variate normal distribution as

$$\begin{bmatrix} \mathbf{f}_{1:t} \\ \nabla \mathbf{f}^{e}_{1:t} \\ f_{t+1} \end{bmatrix} \sim \mathcal{N}\left(0, \begin{bmatrix} \bar{\mathbf{K}} & \bar{\mathbf{k}} \\ \bar{\mathbf{k}}^T & k(\mathbf{x}_{t+1}, \mathbf{x}_{t+1}) \end{bmatrix}\right), \tag{4.9}$$

where $\bar{\mathbf{k}} = [\, k_{[\mathbf{f}_{1:t}, f_{t+1}]}^T, k_{[f_{t+1}, \nabla \mathbf{f}^{e}_{1:t}]} \,]^T$, and the predictive distribution on $\mathbf{x}_{t+1}$ is a normal distribution $\mathcal{N}(\bar{\mu}, \bar{\sigma}^2)$ where $\bar{\mu}(\mathbf{x}_{t+1})$ and $\bar{\sigma}^2(\mathbf{x}_{t+1})$ are given as

$$\bar{\mu}(\mathbf{x}_{t+1}) = \bar{\mathbf{k}}^T \bar{\mathbf{K}}^{-1} [\mathbf{f}_{1:t}, \nabla \mathbf{f}^{e}_{1:t}]^T, \tag{4.10}$$

$$\bar{\sigma}^2(\mathbf{x}_{t+1}) = k(\mathbf{x}_{t+1}, \mathbf{x}_{t+1}) - \bar{\mathbf{k}}^T \bar{\mathbf{K}}^{-1} \bar{\mathbf{k}}. \tag{4.11}$$

We then use the Eq.(4.10) and Eq.(4.11) to construct acquisition function and perform Bayesian optimization. The proposed algorithm is described in Algorithm 4.1:

**Algorithm 4.1** Bayesian Optimization using Derivative Meta-model (**BODMM**)

**for** $n = 1, 2, ... t$ **do**
  Fit the data $D$ using GPPK
  Estimate derivative values $\nabla \mathbf{f}^{e}_{1:t}$ from GPPK via Eq.(4.6)
  Build GP with function observations and estimated derivatives of obser-vations
  Find $\mathbf{x}_{t+1}$ by optimizing the acquisition function:
    $\mathbf{x}_{t+1} = \text{argmax}_{\mathbf{x}} \alpha_{\text{EI}}(\mathbf{x}|D)$
  Evaluate the objective function: $y_{t+1} = f(\mathbf{x}_{t+1})$
  Augment the observation set $D = D \cup (\mathbf{x}_{t+1}, y_{t+1})$.
**end for**

## 4.3 Degree estimation

Given the degree d = 2 in the polynomial kernel, the posterior mean function in GP can be maximum quadratic. While a quadratic meta-model may be sufficient for functions having only one peak, for functions with more than one peak, we provide a Bayesian paradigm to estimate the degree of the polynomial.

In Bayesian treatment, the posterior probability of a variable is proportional to the product of its prior and the likelihood. In our case, the posterior of the degree d is computed as

$$p(\ \mid X, \boldsymbol{y}) \propto p(\ )\, p(\boldsymbol{y} \mid X,\ ), \tag{4.12}$$

where $X = \{\boldsymbol{x}_i\}^t_{i=1}$. The prior $p$(d) represents our belief on the degree. Since the degree is discrete, we choose the geometric distribution as our prior although other distributions are alternatives. The geometric distribution presents the probability that the first occurrence of success requires $m$ independent trials, each with success probability $q$,

$$p(\ = m) = (1 - q)^{m-1} q, \tag{4.13}$$

where $m = 1, 2, 3 \dots$. In practice, we do not expect that the d is over than a high value such as 10 and then we can use the truncated geometric distribution with re-normalization. Figure 4.2 illustrates the geometric distribution with different $q$ truncated at d = 10. The likelihood $p(\boldsymbol{y} \mid X,\ )$ in Eq.(4.12) is the marginal likelihood of GP with polynomial kernel. We compute it as following

$$\log p(\boldsymbol{y} \mid X,\ ) = -\frac{1}{2}\boldsymbol{y}^T (\mathbf{K} + \sigma^2 I)^{-1}\mathbf{y} - \frac{1}{2} \log |\mathbf{K} + \sigma^2 I| - \frac{n}{2} \log 2\pi. \tag{4.14}$$

Given the Eq.(4.13) and Eq.(4.14), we can compute the posterior as Eq.(4.12), and then use the mode of the posterior as the estimated degree. Once we infer the , we directly apply it into the derivative estimation as Eq.(4.6).

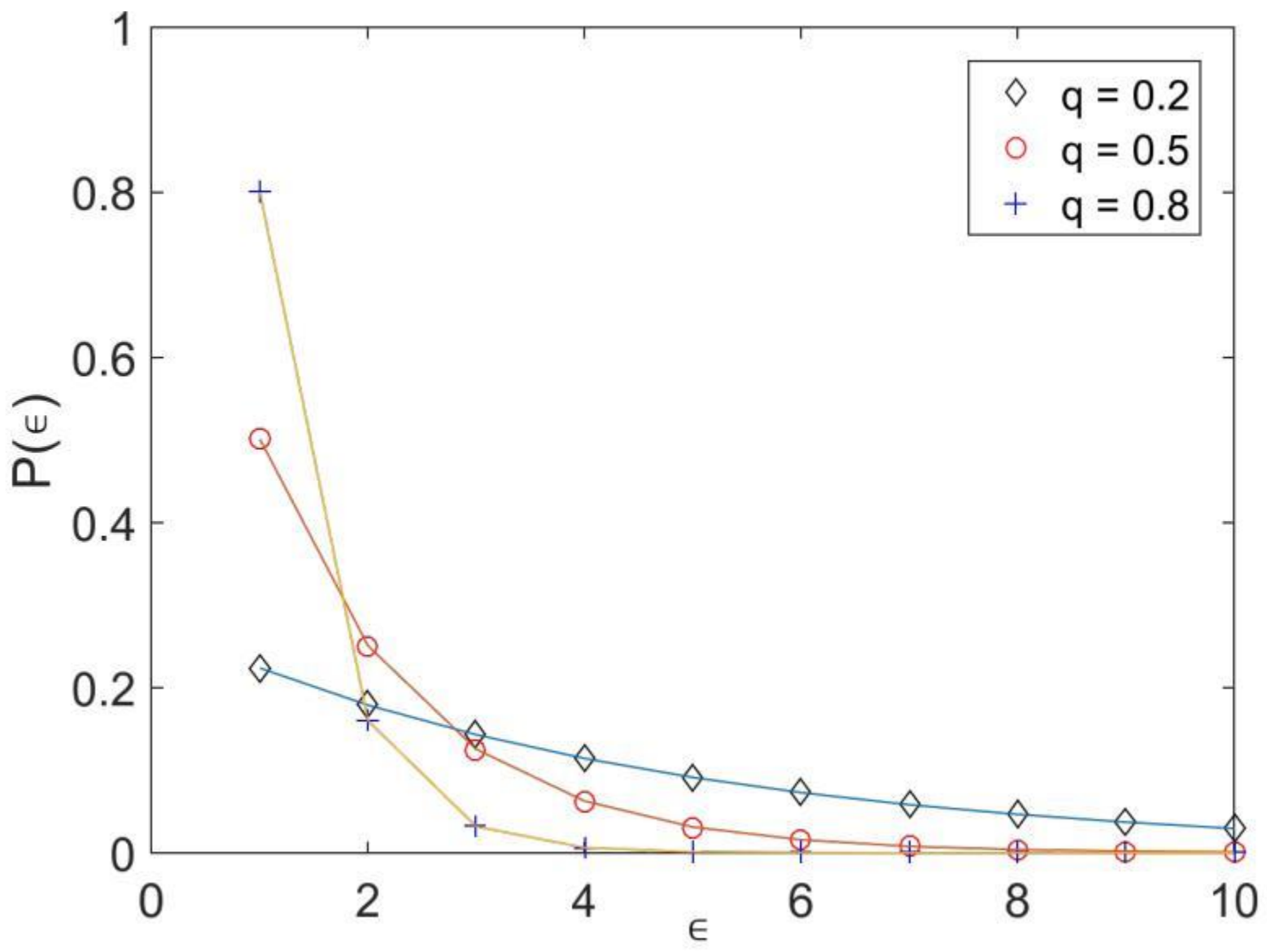

Figure 4.2: Truncated geometric distribution(truncated at = 10) for $q$ = 0.2, 0.5 and 0.8.

## 4.4 Experiments

We firstly demonstrate the GPPK can effectively capture function shape. Secondly, we investigate the performance of GPPK on optimizing synthetic functions. We then evaluate our method on three different benchmark functions and real world applications on hyperparameter tuning for two machine learning algorithms. We compare the proposed method BODMM with the following baselines for benchmark functions:

- Bayesian Optimization without derivative observations (Standard BO).
- Bayesian Optimization using true derivative values (BOTD).

As the true derivative values are not available in real applications, so we only com-pare with Standard BO in this scenario.

In all experiments, we use EI as the acquisition function and the SE kernel as the covariance function. We use DIRECT [Finkel2003] to optimize the acquisition function. In terms of kernel parameters, we use the isotropic length-scale $\rho_l$ = 0.1, signal variance $\sigma_f^2$ = 1 and noise variance $\sigma_n^2$ = $(0.01)^2$. In GPPK, we use $\sigma$ = 0.5 as kernel offset and $q = 0.5$ in truncated geometric distribution. The number of initial observations are set as dimensions+1. The computer used is a Xeon Quad-core PC running at 2.6 GHz, with 16 GB of RAM. We run each algorithm 100 trials with different initialization and report the simple regret and standard errors for benchmark functions while reporting accuracy for hyperparameter tuning tasks. Simple regret is defined as

$$r_t = f(\mathbf{x}^*) - f(\mathbf{x}_t^+), \tag{4.15}$$

where $f(\mathbf{x}^*)$ is the global optimum and $f(\mathbf{x}^+_t) = max_{\mathbf{x}\ \{\mathbf{x}_{1:t}\}} f(\mathbf{x})$ which is the current best value.

### 4.4.1 Experiment on GPPK shape

We visualize GPPK on two benchmark functions as below:

1. $1d$ function with multiple local optima $y = \exp(-(x-2)^2)+\exp(-(x-6)^2/10)+1/(x^2+1)$. The global maximum is $f(x^*) = 1.40$ at $x^* = 2.00$ where search space is in $[1, 4]$;

2. $2d$ Branin's function (Branin-2D). The global minimum is $f(\mathbf{x}^*) = 0.397887$ at $\mathbf{x}^* = (\pi, 2.275)$ where search space is in $[0, 4]$.

For the $1d$ function, we compare the true function (green solid line) with mean function of the GPPK (magenta dot line) in Figure 4.3a. The graphs from left to right illustrates the estimated mean function of GPPK by using 3 randomly initial observations only, after 2 iterations and 5 iterations respectively. The results show that GPPK can approximately capture the shape of the $1d$ objective function. Figure 4.3b demonstrates the results for $2d$ Branin's function. We can see that the mean function of GPPK is close to the true function after 15 iterations.

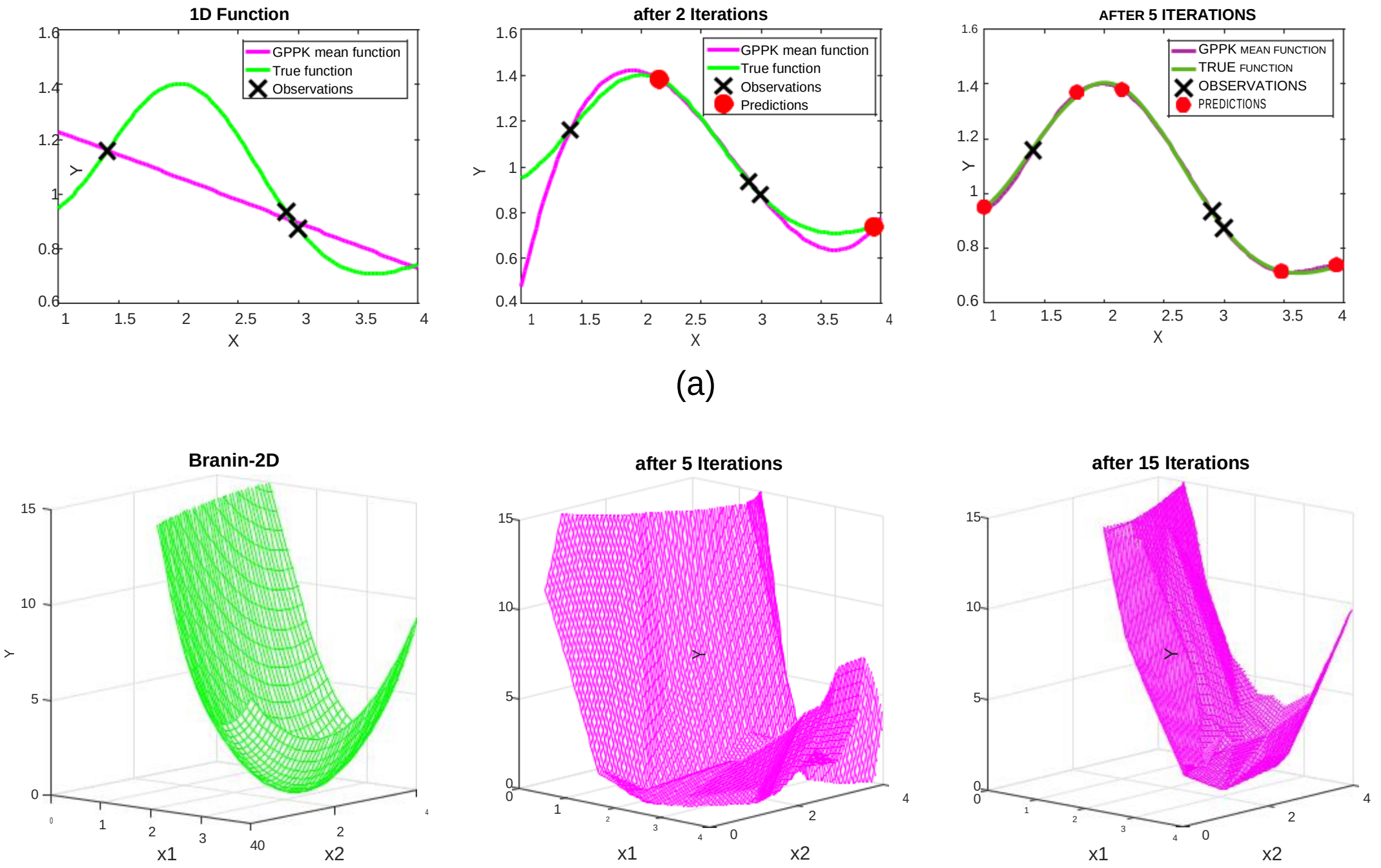

(a)

(b)

Figure 4.3: The visualization of the mean functions of GPPK and the true functions (a) illustrates the mean functions of GPPK at initial observations, 2 iterations and 5 iterations for 1D function (b) demonstrates the mean functions of GPPK at 5 iterations and 15 iterations for 2D Branin's function.

### 4.4.2 Experiment on GPPK performance of optimizing benchmark functions

We also consider to use GPPK as our main GP for Bayesian optimization. To investigate the feasibility, we compare the performance of Bayesian optimization using GPPK and Gaussian process with squared exponential (SE) on optimizing synthetic functions. We apply the comparison on two functions:

1. Unnormalized 3$d$ Gaussian PDF (Gaussian PDF-3D). The global maximum is $f(\mathbf{x}^*) = 1$ at $\mathbf{x}^* = [1, 1, 1]$ where search space is $[0, 2]$ for each dimension.
2. 3$d$ Hartmann (Hartmann-3D). The global minimum is $f(\mathbf{x}^*) = -3.86278$ at $\mathbf{x}^* = [0.114614, 0.555649, 0.852547]$ where search space is in $[0, 1]$ for each dimension;

We start to conduct the examination on Gaussian PDF-3D. Figure 4.4 demonstrates the simple regret vs iteration for GPPK and SE on optimizing this function. We use estimated degree in GPPK and we can see from Figure 4.4a that the GPPK slightly outperforms SE in this experiment. We then apply GPPK and SE on optimizing Hartmann-3D, which is a more complex function and we illustrate the results in Figure 4.4b . Although the GPPK still performs as desired, but SE is significantly better in this case. Base on the examinations, we find the SE is more capable to handle a variety of problems well. Therefore, in our model, we only want to pass the shape information through estimated derivative by using GPPK, but still use squared exponential in the main GP for Bayesian optimization.

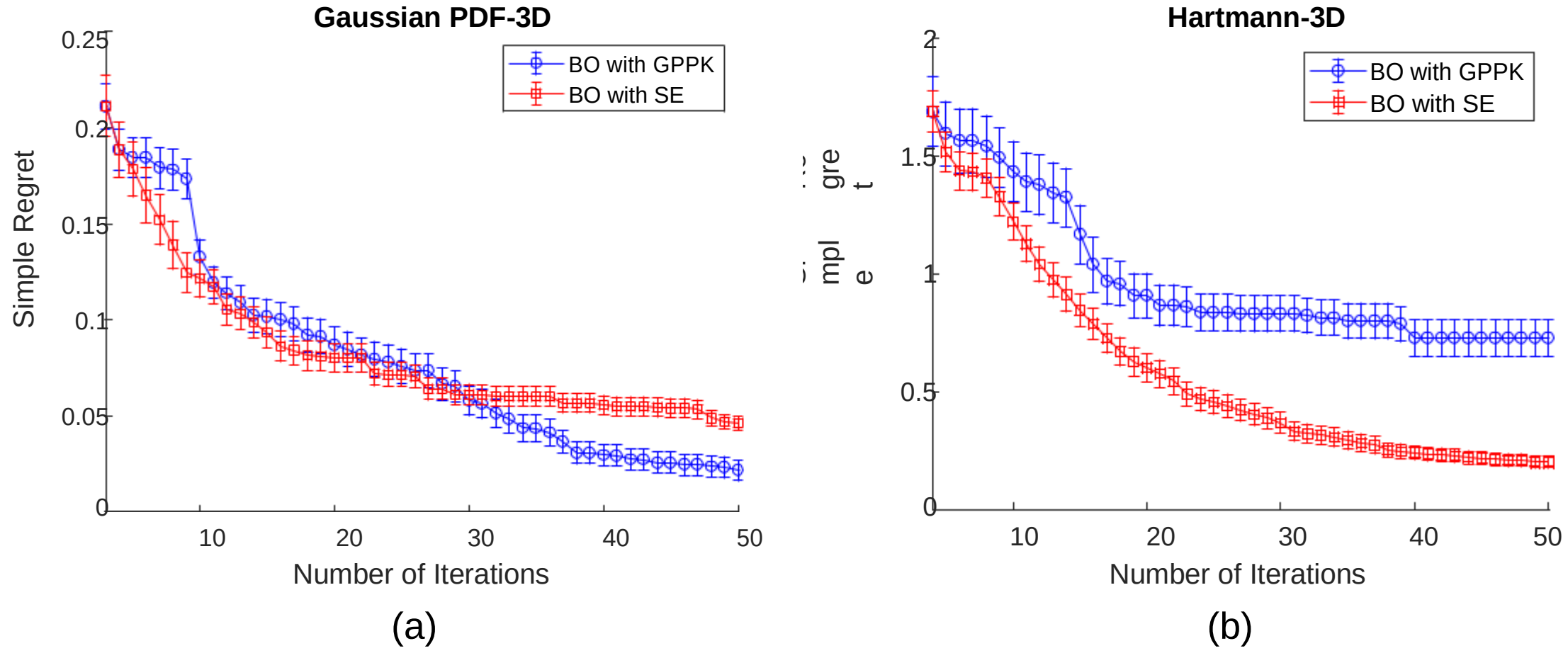

Figure 4.4: Comparison of simple regret vs iterations for GPPK and SE on (a) GaussianPDF-5D, (b) Hartmann-3D function.

### 4.4.3 Experiment with benchmark functions

We test our algorithm on 2D Branin's function and other two benchmark functions as below:

1. 2*d* Branin's function (Branin-2D)
2. 3*d* Hartmann (Hartmann-3D).
3. Unnormalized 5*d* Gaussian PDF (Gaussian PDF-5D). The global maximum is $f(\boldsymbol{x}^*) = 1$ at $\boldsymbol{x}^* = [1, 1, 1, 1, 1]$ where search space is $[0, 2]$ for each dimension.

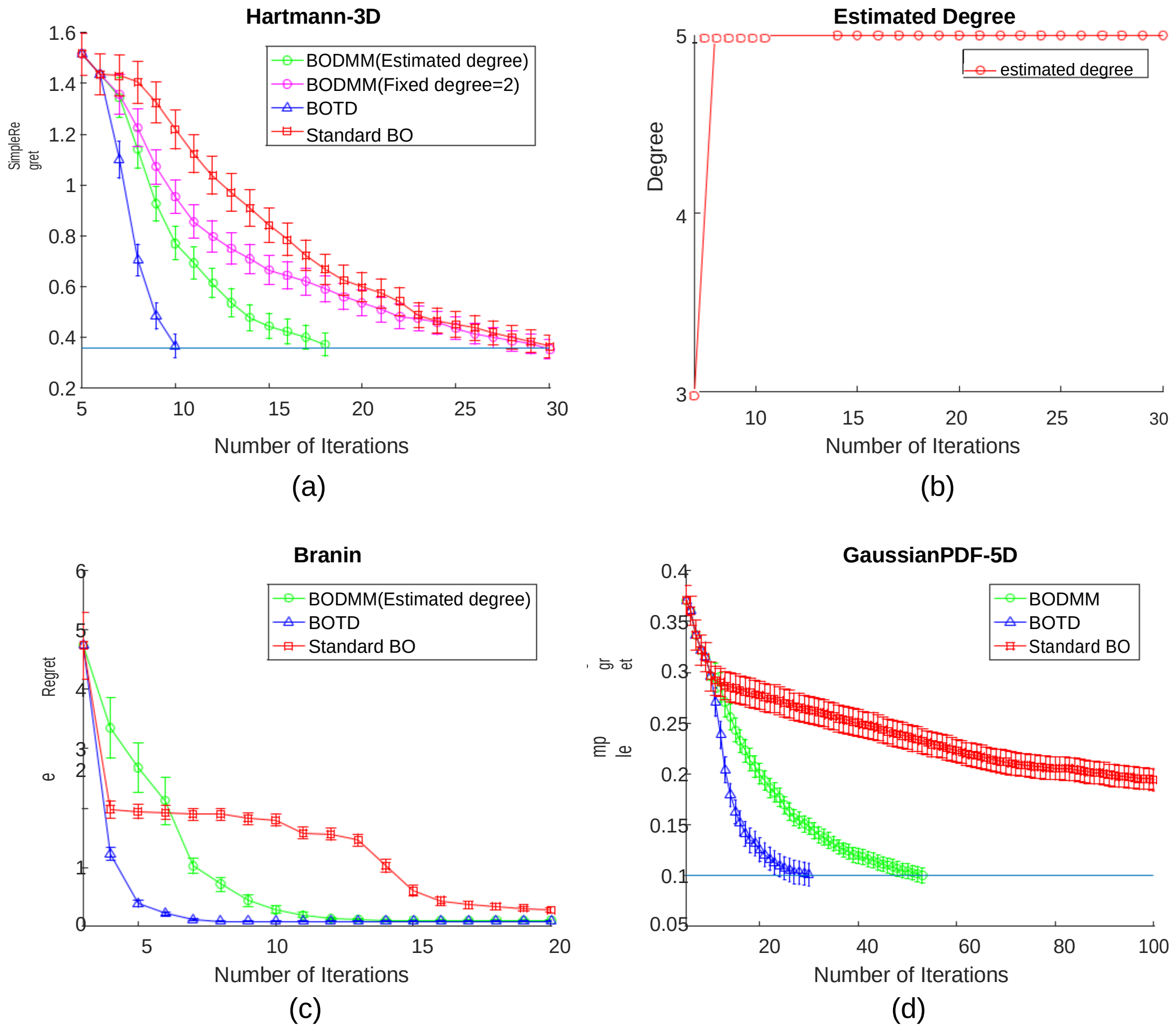

Figure 4.5: Simple regret vs iterations for (a) Hartmann-3D function (b)Estimated degree *d* in the GPPK for optimising Hartmann-3D function (c) Branin's Function (d) GaussianPDF-5D.

- convergence threshold: reaching 10% to the optimum values. We did not apply the threshold on Branin's function since the BO for it can converge fast.

We start to experiment our algorithm on Hartmann-3D. Figure 4.5a plots the simple regret vs iteration for three different algorithms. For the proposed BODMM we run it with fixed degree and estimated degree, respectively. Bayesian optimization using true derivative values performs the best in all three algorithms and converges after 10 iterations. It is easy to explain since true derivatives have been incorporated in this algorithm. Our algorithms with fixed degree and estimated degree outperform Standard BO. The setting with estimated degree performs better than that of fixed degree. Fig 4.5b demonstrates the estimated degree for Hartmann-3D case at each iteration. We also receive positive results from other test functions. Results of Branin's function and Gaussian PDF-5D have been illustrated in Fig 4.5c and Fig 4.5d respectively.

### 4.4.4 Hyperparamter tuning

We experiment with three real world datasets for tuning hyperparameters of two classifiers: Support Vector Machines (SVM) and Elastic Net. In SVM we optimize two hyperparameters which are the cost parameter ($C$) and the width of the RBF kernel ($\gamma$). The search bounds for the two hyperparameters are $C = 10^{\lambda}$ where $\lambda \in [-3, 3]$ and $\gamma = 10^{\omega}$ where $\omega \in [-5, 0]$ correspondingly. To make our search bounds manageable, we optimize for $\lambda$ and $\omega$ (i.e. we optimize in the exponent space.). In Elastic Net, the hyperparameters are the $l_1$ and $l_2$ penalty weights. The search bound for both of them is $[10^{-5}, 10^{-2}]$. We optimize in the range of exponents ($[-5, -2]$). For each dataset we randomly sample 70% of the data for training, and the rest as validation. All three datasets: BreastCancer, LiverDisorders and Mushrooms are publicly available from UCI data repository [Dheeru and Karra Taniskidou2017].

The results for the hyperparameter tuning of SVM and Elastic Net on the different datasets are shown in Fig 4.6. In all cases our approach BODMM performs better than Standard BO. For example in the right graphic of Fig 4.6a, Standard BO achieves **0.89** after 40 iterations while our algorithm achieves **0.97**.

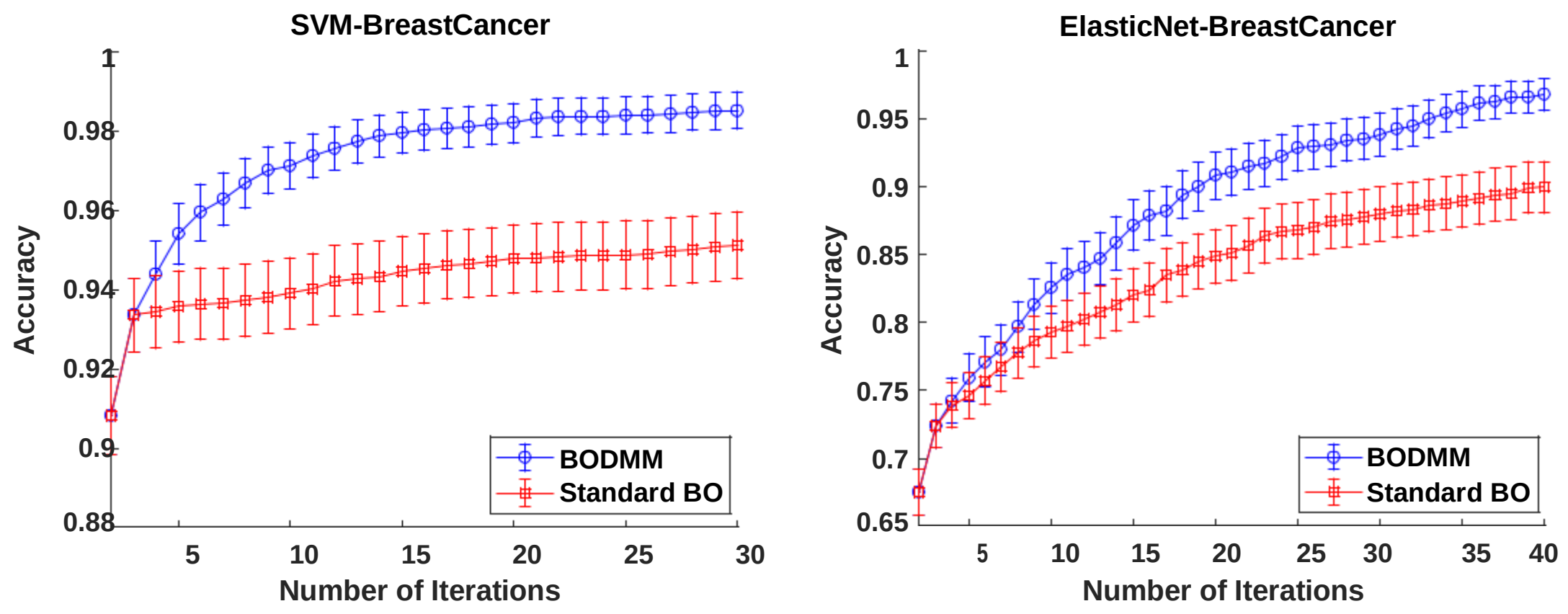

(a) BreastCancer

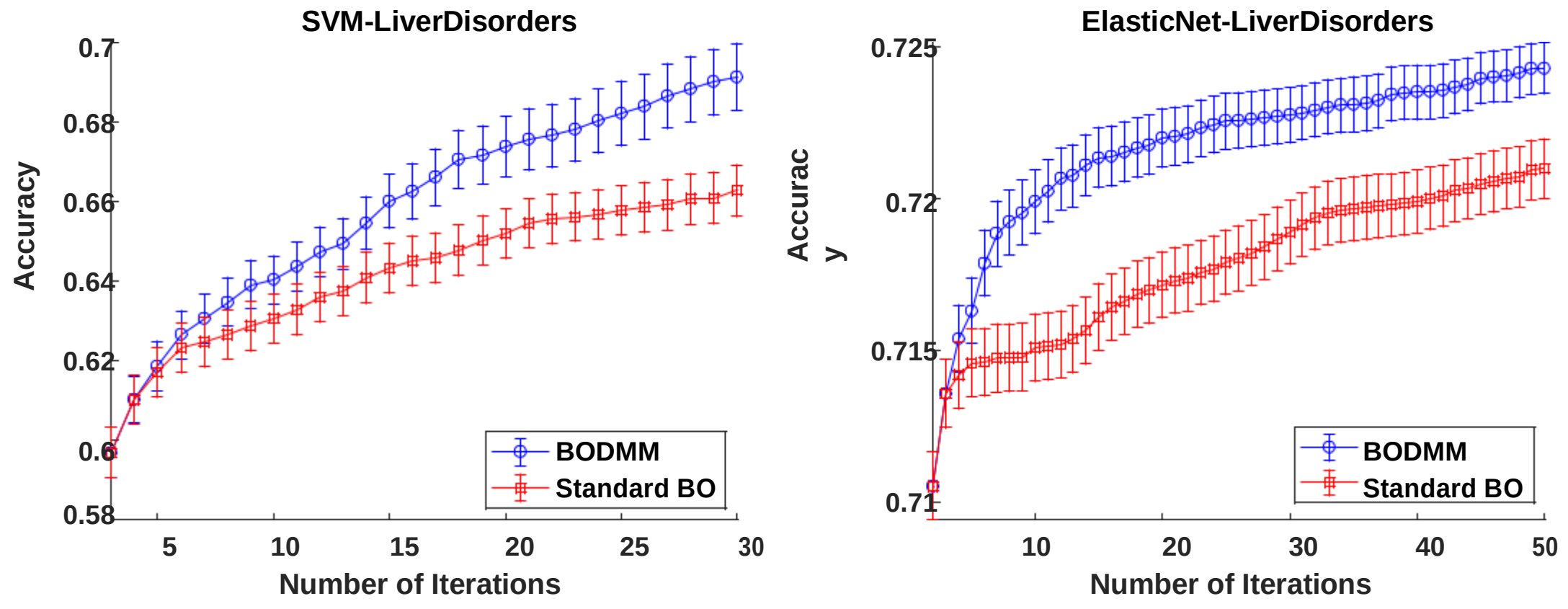

(b) LiverDisorders

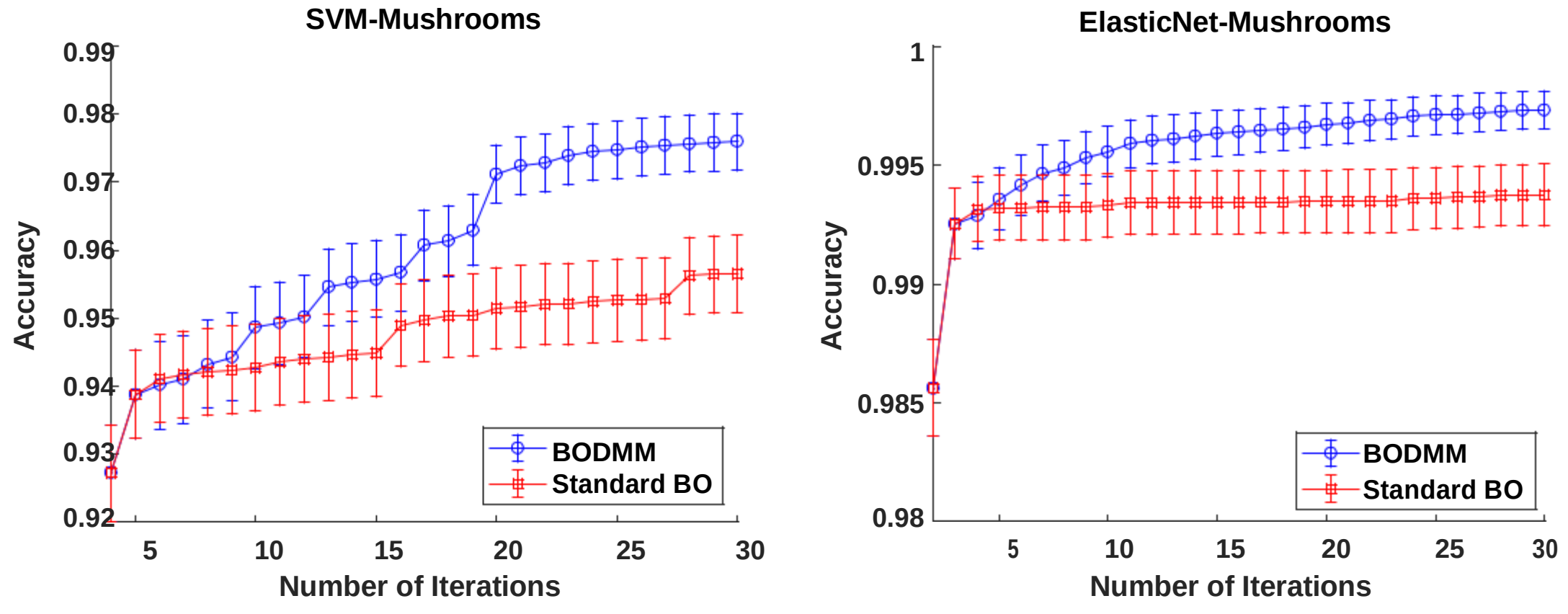

(c) Mushrooms

Figure 4.6: Accuracy vs iterations for hyperparameter tuning of SVM (left) and Elastic Net (right) on (a) BreastCancer, (b)LiverDisorders and (c) Mushrooms.

## 4.5 Summary

We propose a novel method for Bayesian optimization for well-behaved functions with a small number of peaks. We incorporate this information through a derivative meta-model. The derivative meta-model is based on a Gaussian process with polynomial kernel. By controlling the degree of the polynomial we control the shape of the main Gaussian process where covariance matrix is computed by using both the observed function value and the derivative values estimated from the meta-model. We refrain from using the samples of the function values from GPPK because we only want to pass the shape information through derivatives, while keeping the func-tion values guided mostly by the main GP. We also provide a Bayesian paradigm to estimate the degree of the polynomial based on placing a truncated geometric prior. In experiments, both on benchmark function optimization and the hyperparameter tuning of machine learning models, our proposed model converges faster than the baselines.

# Chapter 5

# Sparse Approximation for Gaussian Process with Derivative Observations

In Chapter 4 we have proposed a meta-model to estimate derivatives for the latent function and then incorporated them into Gaussian process (GP) to accelerate the convergence of Bayesian optimization. However, a practical limitation of a generic Gaussian process is that its computational complexity increases rapidly with the size of the training set. Specifically, the standard Gaussian process requires $O(t^3)$ time to compute the Cholesky decomposition of a covariance matrix, where $t$ is the number of function observations. The same happens to our BODMM model in Chapter 4 which incorporates both function observations and derivative observations. We shall recall that each derivative observation is a vector and each entry is a separate observation for GP. So the number of total observations also scales with the dimension. Hence, the computational complexity of GP with full derivative observations is $O((t+dt^0)^3)$, where $t^0$ is the number of derivative observations and $d$ is the dimensionality of the space. As a result, it becomes very difficult to apply GP to large scale derivative observations.

Our literature review failed to find a sparse approximation for Gaussian process that can include derivative observations. Therefore, the aim of this chapter is to develop an efficient sparse GP model to approximate the full GP with the presence of derivative observations while preserving its predictive accuracy. There are many sparse Gaussian process methods and they can be categorized into three groups as 1) prior approximations, 2) posterior approximations and 3) structured sparse approx-imations. Among them, the prior approximations use the independence assumption that the dependencies between training and testing points are only induced through inducing variables [Quinonero-Candela and Rasmussen2005] so that we can modify the joint prior and easily to reduce the computational complexity to $O(tm^2)$, where $m$ is the number of inducing points. Therefore, we choose this sparse method and use a set of inducing variables, which contains the number of $m$ inducing points in our framework and $m \leq t$. Firstly, we assume that the function observations and derivative observations are conditionally independent given the set of inducing variables. This assumption is common in other methods [Quinonero-Candela and Rasmussen2005, Snelson and Ghahramani2006]. Then the conditional distribution for a test point can be obtained by integrating out the inducing variables. To fur-ther decrease the costly computation, we introduce the fully independent conditional approximation where function observations, derivative observations and test points do not have any deterministic relation on the inducing variables, so that we can ignore the covariance but retain the variance. This work is successful in developing a sparse GP model to approximate the full GP with derivative observations. The resultant sparse Gaussian process with derivatives requires only $O((t + dt^0)m^2)$ for posterior computation, which is significant better than the original complexity of $O((t + dt^0)^3)$. We have applied the proposed model in regression and Bayesian op-timization on large scale datasets. The experimental results show the effectiveness of our proposed model.

In the subsequent sections, we introduce our sparse approximation for Gaussian process with derivative observations. In Section 5.1 we briefly review the sparse Gaussian process with fully independent conditionals for function observations. Later, in Section 5.2 we derive our model to approximate Gaussian process with deriva-tive observations and investigate the usability of our method in large scale Bayesian optimization. Section 5.3 examines the effectiveness of our proposed model on approximating Gaussian process with a large number of function observations and derivatives observations. We summarize our contributions in Section 5.4.

## 5.1 Sparse Gaussian process with fully independent conditionals

We briefly review the sparse Gaussian process with fully independent conditionals (FIC) and details are referred to Chapter 3. In FIC, it introduces a set of inducing observations $D_u = \{\mathbf{M}_i, \mathbf{u}_i\}_{i=1}^{m}$ where $\mathbf{u} = \{\mathbf{u}_i\}_{i=1}^{m}$ contains values of the function at the points $\mathbf{M} = \{\mathbf{M}_i\}_{i=1}^{m}$, knowing as inducing points. Given the prior distribution $p(\mathbf{u}) = \mathcal{N}(\mathbf{u} \mid \mathbf{0}, \mathbf{K}_{m,m})$, the training conditional distribution of $\boldsymbol{f}_{1:t} = \{f(\boldsymbol{x}_i)\}_{i=1}^{t}$ given $\mathbf{u}$ can be written as

$$p(\boldsymbol{f}_{1:t} \mid \mathbf{u}) = \mathcal{N}(\boldsymbol{f}_{1:t} \mid \mathbf{K}_{t,m}\mathbf{K}_{m,m}^{-1}\mathbf{u}, \mathbf{K}_{t,t} - \mathbf{Q}_{t,t}), \tag{5.1}$$

where the shorthand notation $\boldsymbol{Q}_{a,b} = \boldsymbol{K}_{a,m}\boldsymbol{K}_{m,m}^{-1}\boldsymbol{K}_{m,b}$. The test conditional $p(f_{t+1} \mid \mathbf{u})$ is formed in the same way with Eq.(5.1) as

$$p(f_{t+1} \mid \mathbf{u}) = \mathcal{N}(f_{t+1} \mid \mathbf{K}_{t+1,m}\mathbf{K}_{m,m}^{-1}\mathbf{u}, \mathbf{K}_{t+1,t+1} - \boldsymbol{Q}_{t+1,t+1}). \tag{5.2}$$

To recover $p(\boldsymbol{f}_{1:t}, f_{t+1})$, we can simply integrate out $\mathbf{u}$ from the joint GP prior $p(\boldsymbol{f}_{1:t}, f_{t+1}, \mathbf{u})$ and obtain

$$p(\boldsymbol{f}_{1:t}, f_{t+1}) = \int p(\boldsymbol{f}_{1:t}, f_{t+1}, \mathbf{u})\, d\mathbf{u} = \int p(\boldsymbol{f}_{1:t}, f_{t+1} \mid \mathbf{u})p(\mathbf{u})d\mathbf{u}. \tag{5.3}$$

We make the assumption that $\boldsymbol{f}_{1:t}$ and $f_{t+1}$ are conditionally independent given $\mathbf{u}$, so the joint distribution of them can be written as

$$p(\boldsymbol{f}_{1:t}, f_{t+1}) = \int p(\boldsymbol{f}_{1:t} \mid \mathbf{u})p(f_{t+1} \mid \mathbf{u})p(\mathbf{u})d\mathbf{u}. \tag{5.4}$$

Then by making assumption under FIC (i.e. the training points and test points are fully independent on $\mathbf{u}$), we can derive the approximation of Eq.(5.1) as

$$p(\boldsymbol{f}_{1:t} \mid \mathbf{u}) \approx q(\boldsymbol{f}_{1:t} \mid \mathbf{u}) = \mathcal{N}(\boldsymbol{K}_{t,m}\boldsymbol{K}_{m,m}^{-1}\mathbf{u}, diag[\boldsymbol{K}_{t,t} - \mathbf{Q}_{t,t}]), \tag{5.5}$$

and the approximation of Eq.(5.2) as

$$p(f_{t+1} \mid \mathbf{u}) \approx q(f_{t+1} \mid \mathbf{u}) = \mathcal{N}\left(\boldsymbol{K}_{t+1,m}\boldsymbol{K}^{-1}_{m,m}\mathbf{u}, diag[\boldsymbol{K}_{t+1,t+1} - \mathbf{Q}_{t+1,t+1}]\right). \tag{5.6}$$

## 5.2 Proposed Framework

Full GP with derivative observations becomes prohibitive when a large number of function observations and derivative observations exist as illustrated before. We propose a sparse model to approximate the full GP with the presence of deriva-tive observations. First, we derive our model named sparse Gaussian process with derivatives (SGPD). Then, we apply the proposed sparse method as the probabilistic surrogate model in Bayesian optimization.

### 5.2.1 Sparse Gaussian Process with Derivatives for Regression

Following the idea of fully independent conditional spare Gaussian process, we also induce a set of sparse variables **u** and use the same notations as before. We can easily get the following joint distribution of function observations, derivative observations and a test point by marginalizing out **u** as

$$p(\boldsymbol{f}_{1:t}, \nabla\boldsymbol{f}_{1:t'}, f_{t+1}) = \int p(\boldsymbol{f}_{1:t}, \nabla\boldsymbol{f}_{1:t'}, f_{t+1} \mid \mathbf{u})p(\mathbf{u})d\mathbf{u}, \tag{5.7}$$

where $\nabla\boldsymbol{f}_{1:t'} = \{\nabla f(\boldsymbol{x}_i)\}_{i=1}^{t'}$. Further we make the assumption that function observations, derivative observations and a test point only depend on the inducing variables **u**. Then we can rewrite the Eq.(5.7) as

$$p(\boldsymbol{f}_{1:t}, \nabla\boldsymbol{f}_{1:t'}, f_{t+1}) = \int p(\boldsymbol{f}_{1:t} \mid \mathbf{u})p(\nabla\boldsymbol{f}_{1:t'} \mid \mathbf{u})p(f_{t+1} \mid \mathbf{u})p(\mathbf{u})d\mathbf{u}. \tag{5.8}$$

Next we show how to compute the conditional distributions $p(\boldsymbol{f}_{1:t} \mid \mathbf{u})$, $p(\nabla\boldsymbol{f}_{1:t'} \mid \mathbf{u})$ and $p(f_{t+1} \mid \mathbf{u})$ using the exact expressions. Given the prior $p(\mathbf{u}) = \mathcal{N}(\mathbf{u} \mid \mathbf{0}, \boldsymbol{K}_{m,m})$, the exact expression for $p(\boldsymbol{f}_{1:t} \mid \mathbf{u})$ can be computed as

$$p(\boldsymbol{f}_{1:t} \mid \mathbf{u}) = \mathcal{N}(\boldsymbol{f}_{1:t} \mid \boldsymbol{K}_{t,m}\mathbf{K}_{m,m}^{-1}\mathbf{u}, \mathbf{K}_{t,t} - \boldsymbol{Q}_{t,t}), \tag{5.9}$$

where we recall $\boldsymbol{Q}_{a,b} = \boldsymbol{K}_{a,m}\boldsymbol{K}_{m,m}^{-1}\boldsymbol{K}_{m,b}$. Similarly, the conditional distribution for a test point $\{\boldsymbol{x}_{t+1}, f_{t+1}\}$ is given as

$$p(f_{t+1} \mid \mathbf{u}) = \mathcal{N}(f_{t+1} \mid \mathbf{K}_{t+1,m}\boldsymbol{K}_{m,m}^{-1}\mathbf{u}, \boldsymbol{K}_{t+1,t+1} - \boldsymbol{Q}_{t+1,t+1}), \tag{5.10}$$

and the conditional distribution for derivative observations is derived as

$$p(\nabla \boldsymbol{f}_{1:t'} \mid \mathbf{u}) = \mathcal{N}(\nabla \boldsymbol{f}_{1:t'} \mid \boldsymbol{K}_{t',m}\boldsymbol{K}_{m,m}^{-1}\mathbf{u}, \boldsymbol{K}_{t',t'} - \boldsymbol{Q}_{t',t'}). \tag{5.11}$$

It is noted that $\boldsymbol{K}_{t',m}$ is the covariance matrix between derivative observations $\nabla \boldsymbol{f}_{1:t'}$ and inducing variables $\mathbf{u}$ and $\boldsymbol{K}_{t',t'}$ is the covariance matrix of derivative observa-tions. The shorthand notation $\boldsymbol{Q}_{t',t'} = \boldsymbol{K}_{t',m}\boldsymbol{K}_{m,m}^{-1}\boldsymbol{K}_{m,t'}$.

To further reduce the computational cost, we can make additional assumptions to approximate conditional distributions in Eq.(5.9, 5.10, 5.11). There are various as-sumptions in the literature such as subset of regressors (SoR) and fully independent training conditional (FITC), and we have discussed and compared these assump-tions in Chapter 2. In this framework, we use the fully independent conditional (FIC) approximation [Quinonero-Candela and Rasmussen2005] where training func-tion observations, derivative observations and test points are fully independent on $\mathbf{u}$. The FIC extends the fully independence assumption in FITC to test data and can perform a closer approximation to the prior distribution than SoR [Smola and Bartlett2001].

Now, under the FIC assumptions, we can derive the approximate expression of $p(\boldsymbol{f}_{1:t} \mid \mathbf{u})$ as

$$q(\boldsymbol{f}_{1:t} \mid \mathbf{u}) = \mathcal{N}(\boldsymbol{K}_{t,m}\boldsymbol{K}_{m,m}^{-1}\mathbf{u}, \mathit{diag}[\boldsymbol{K}_{t,t} - \boldsymbol{Q}_{t,t}]), \tag{5.12}$$

where $\mathit{diag}[\boldsymbol{K}_{t,t} - \boldsymbol{Q}_{t,t}]$ is a diagonal matrix whose elements match the diagonal of $\boldsymbol{K}_{t,t} - \boldsymbol{Q}_{t,t}$ so that we only keep the variance information of function observations themselves and ignore the covariance between function observations and inducing points. Then $\boldsymbol{f}_{1:t}$ have no any deterministic relation on $\mathbf{u}$. Likewise, the approximate expression for $p(f_{t+1} \mid \mathbf{u})$ can be written as

$$q(f_{t+1} \mid \mathbf{u}) = \mathcal{N}(\boldsymbol{K}_{t+1,m}\boldsymbol{K}_{m,m}^{-1}\mathbf{u}, diag[\boldsymbol{K}_{t+1,t+1} - \boldsymbol{Q}_{t+1,t+1}]), \tag{5.13}$$

and the approximated format for $p(\nabla\boldsymbol{f}_{1:t'} \mid \mathbf{u})$ is

$$q(\nabla\boldsymbol{f}_{1:t'} \mid \mathbf{u}) = \mathcal{N}(\boldsymbol{K}_{t',m}\boldsymbol{K}_{m,m}^{-1}\mathbf{u}, diag[\boldsymbol{K}_{t',t'} - \boldsymbol{Q}_{t',t'}]). \tag{5.14}$$

We now have approximate expressions of the three conditional distributions. Sub-stituting Eq.(5.12, 5.13, 5.14) into Eq.(5.8) results in the FIC approximate joint distribution

$$\begin{aligned} p(\boldsymbol{f}_{1:t}, \nabla\boldsymbol{f}_{1:t'}, f_{t+1}) &\approx q(\boldsymbol{f}_{1:t}, \nabla\boldsymbol{f}_{1:t'}, f_{t+1}) \\ &= \mathcal{N}\left(\mathbf{0}, \begin{bmatrix} \boldsymbol{KQ}_t & \boldsymbol{Q}_{t,t'} & \boldsymbol{Q}_{t,t+1} \\ \boldsymbol{Q}_{t',t} & \boldsymbol{KQ}_{t'} & \boldsymbol{Q}_{t',t+1} \\ \boldsymbol{Q}_{t+1,t} & \boldsymbol{Q}_{t+1,t'} & \boldsymbol{KQ}_{t+1} \end{bmatrix}\right), \end{aligned} \tag{5.15}$$

where $\boldsymbol{KQ}_t = \boldsymbol{Q}_{t,t} + diag[\boldsymbol{K}_{t,t} - \boldsymbol{Q}_{t,t}]$, $\boldsymbol{KQ}_{t'} = \boldsymbol{Q}_{t',t'} + diag[\boldsymbol{K}_{t',t'} - \boldsymbol{Q}_{t',t'}]$ and $\boldsymbol{KQ}_{t+1} = \boldsymbol{Q}_{t+1,t+1} + diag[\boldsymbol{K}_{t+1,t+1} - \boldsymbol{Q}_{t+1,t+1}]$.

The posterior distribution of the test point $\{\boldsymbol{x}_{t+1}, f_{t+1}\}$ is a Gaussian distribution

$$f(\boldsymbol{x}_{t+1}) \sim \mathcal{N}(\tilde{\mu}(\boldsymbol{x}_{t+1}), \tilde{\sigma}^2(\boldsymbol{x}_{t+1})), \tag{5.16}$$

with mean and variance as

$$\tilde{\mu}(\boldsymbol{x}_{t+1}) = \boldsymbol{Q}_{t+1,tt'}\boldsymbol{K}_{t,t'}^{FIC\,-1}[\boldsymbol{f}_{1:t}, \nabla\boldsymbol{f}_{1:t'}], \tag{5.17}$$

$$\tilde{\sigma}^2(\boldsymbol{x}_{t+1}) = \boldsymbol{K}_{t+1,t+1} - \boldsymbol{Q}_{t+1,tt'}\boldsymbol{K}_{t,t'}^{FIC\,-1}\boldsymbol{Q}_{tt',t+1}, \tag{5.18}$$

where

$$\boldsymbol{K}^{FIC}_{t,t'} = \begin{bmatrix} \boldsymbol{Q}_{t,t} + diag[\boldsymbol{K}_{t,t} - \boldsymbol{Q}_{t,t}] & \boldsymbol{Q}_{t,t'} \\ \boldsymbol{Q}_{t',t} & \boldsymbol{Q}_{t',t'} + diag[\boldsymbol{K}_{t',t'} - \boldsymbol{Q}_{t',t'}] \end{bmatrix}, \tag{5.19}$$

$\boldsymbol{Q}_{t+1,tt'} = [\boldsymbol{Q}_{t+1,t}, \boldsymbol{Q}_{t+1,t'}]$ which combines $\boldsymbol{Q}_{t+1,t}$ and $\boldsymbol{Q}_{t+1,t'}$ into one matrix and $\boldsymbol{Q}_{tt',t+1} = \boldsymbol{Q}^T_{t+1,tt'}$.

If the function observation is a noisy estimation of the actual function value as $y = f(\boldsymbol{x}) + \xi$, where $\xi \sim N(\boldsymbol{0}, \sigma_n^2)$. Then the predicted mean and variance can be computed as

$$\tilde{\mu}(\boldsymbol{x}_{t+1}) = \boldsymbol{Q}_{t+1,tt'}[\boldsymbol{K}^{FIC}_{t,t'} + \begin{bmatrix} \sigma_n^2 \\ 0 \end{bmatrix}\mathbf{I}]^{-1}[\boldsymbol{y}_{1:t}, \nabla\boldsymbol{f}_{1:t'}], \tag{5.20}$$

$$\tilde{\sigma}^2(\boldsymbol{x}_{t+1}) = \boldsymbol{K}_{t+1,t+1} - \boldsymbol{Q}_{t+1,tt'}[\boldsymbol{K}^{FIC}_{t,t'} + \begin{bmatrix} \sigma_n^2 \\ 0 \end{bmatrix}\mathbf{I}]^{-1}\boldsymbol{Q}_{tt',t+1}. \tag{5.21}$$

### 5.2.2 Optimal inducing points selection

To use this approximation in practice it remains to specify how the $m$ inducing points are selected. While we could, in principle, randomly select a subset of the $t$ training points for a desired approximation, we found that for complex problems maximizing the log marginal likelihood to select the inducing set will result in a more accurate approximation.

We denote all hyperparameters including the inducing points by $\Theta$. These can be learned by maximizing the log marginal likelihood:

$$\begin{aligned} \log p(\boldsymbol{y}|\Theta) = &-\frac{n}{2}\log(2\pi) - \frac{1}{2}\left|\mathbf{K}^{FIC}_{t,t'} + \begin{bmatrix} \sigma_n^2 \\ 0 \end{bmatrix}\mathbf{I}\right| \\ &- \frac{1}{2}[\boldsymbol{y}_{1:t}, \nabla\boldsymbol{f}_{1:t'}]^T(\mathbf{K}^{FIC}_{t,t'} + \begin{bmatrix} \sigma_n^2 \\ 0 \end{bmatrix}\mathbf{I})^{-1}[\boldsymbol{y}_{1:t}, \nabla\boldsymbol{f}_{1:t'}]. \end{aligned} \tag{5.22}$$

**Complexity analysis** In full GP with derivatives the computational complexity for training is $O((t + dt^0)^3)$ where $t$ is the number of function observations and $t^0$ is the number of the derivative observations. The proposed method which introduced a set of inducing points **u**, bring the complexity down to $O((t + dt^0)m^2)$, where $m$ is the number of inducing points and $m \leq t$.

### 5.2.3 Application to Bayesian optimization

Bayesian optimization using the standard Gaussian process as a prior can be ex-tremely costly if there are a large number of function observations and derivative observations. Therefore, we replace the modeling part with our method SGPD to maintain the scalability of BO with large number of derivative observations. The proposed algorithm is in Algorithm 5.1.

**Algorithm 5.1** Bayesian Optimization using Sparse Gaussian Process with Deriva-tives (**BOSGPD**)

Input data: $D = \{\boldsymbol{x}_i, \boldsymbol{f}_i\}_{i=1}^{t}$, $D^0 = \{\boldsymbol{x}_j, \nabla \boldsymbol{f}_j\}_{j=1}^{t^0}$ and $D_u = \{\mathbf{M}_i, \mathbf{u}_i\}_{i=1}^{m}$

2: Model SGPD using input data.

2: Find $\boldsymbol{x}_{t+1}$ by optimizing the acquisition function $EI(\boldsymbol{x}|D)$:

$$\boldsymbol{x}_{t+1} = argmax_{\boldsymbol{x}} EI(\boldsymbol{x}|D)$$

3: Evaluate the objective function: $y_{t+1} = f(\boldsymbol{x}_{t+1}) + \xi$

4: Augment the observation set $D = D \cup (\boldsymbol{x}_{t+1}, y_{t+1})$,
$D^0 = D^0 \cup \{\boldsymbol{x}_{1:t+1^0}, \boldsymbol{f}_{1:t+1^0}\}$

5: **end for**

## 5.3 Experiments

In this section we evaluate our method on regression tasks and its application of Bayesian optimization. We compare the following Gaussian process models in re-gression tasks:

- Standard Gaussian process (StdGP)

- Sparse Gaussian process (SGP)
- Gaussian process with derivatives (GPD)
- Our method 1: Sparse Gaussian process with derivatives using random subset (SGPD-Random)
- Our method 2: Spare Gaussian process with derivatives using optimal subset (SGPD-Optimal)

In Bayesian optimization of benchmark functions, we compare BO using our two models with BO using StdGP and GPD. For hyperparameter tuning experiments, we compare BO using our SGPD-Optimal (BOSGPD-Optimal) with the same setting in Chapter 4 (i.e. Standar BO and BODMM). For all GPs we use the SE kernel with the hyperparameters - the isotropic length scale $\rho_l$ = 0.8, signal variance $\sigma_f^2$ = 1 and noise variance $\sigma_{noise}^2 = (0.01)^2$. We use EI as acquisition function and DIRECT [Finkel2003] as optimizer to optimize EI.

We visualize the results for 1D function in Figure 1. We use 30 function observations ($t$ = 30) and their derivatives ($t^0$ = 30) as training data and use 450 points for testing. We randomly select 70% of $t$ (21 points) as inducing points (Derivative observations will not be used as inducing points in all of our cases), then we compare our methods with StdGP and GPD. Although the mean functions of the four methods look similar but the variance of our method SGPD-Optimal is closer to GPD than other methods.

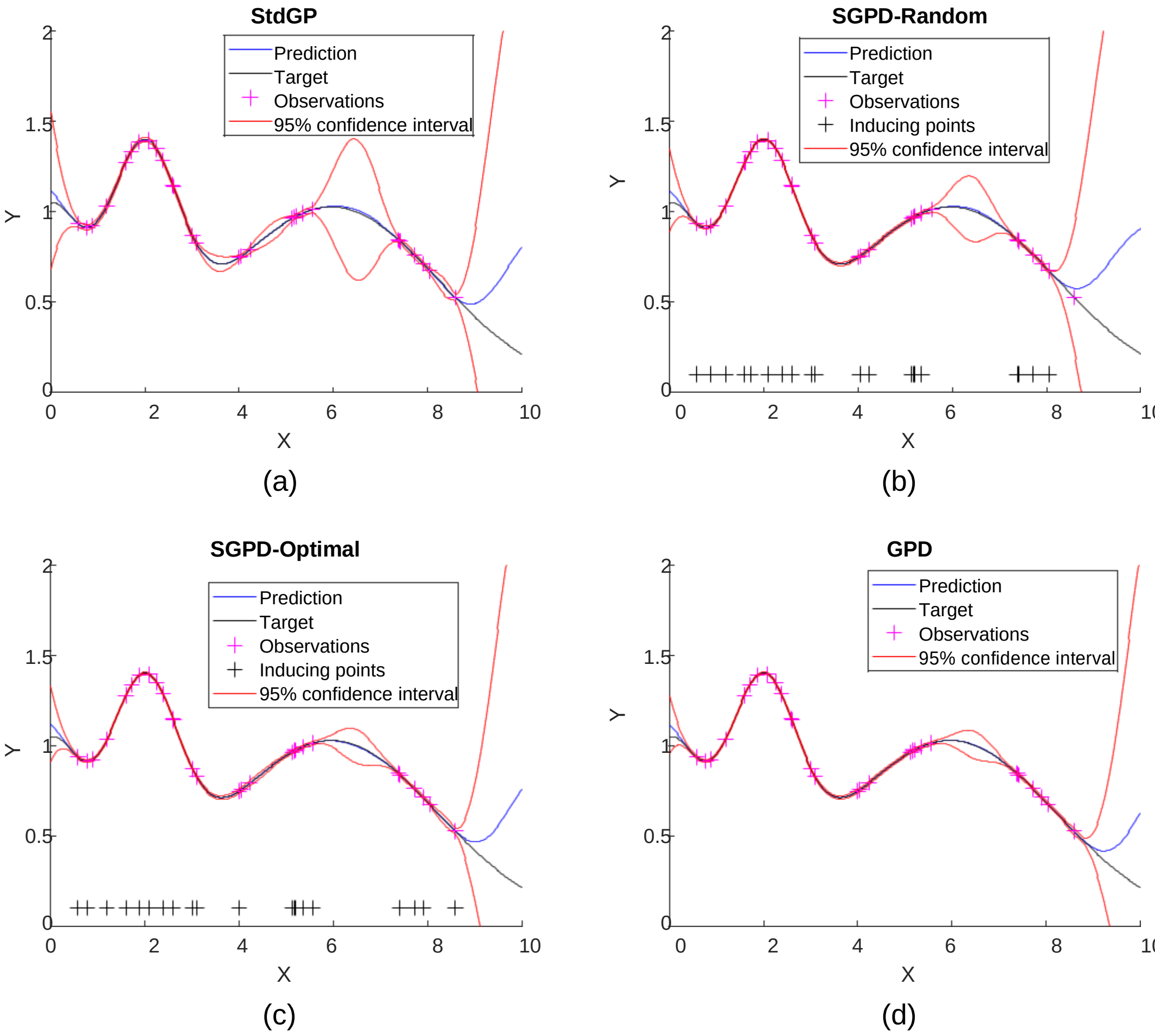

Figure 5.1: The visualisation of four methods for regression on 1*d* function (a) Stan-dard Gaussian process (b) Sparse Gaussian process with derivatives using random subset (c) Spare Gaussian process with derivatives using optimal subset and (d) Gaussian process with derivatives.

## 5.3.1 Regression on benchmark functions

In regression tasks, we run each algorithm 50 trials with different initialization and report the mean square error (MSE). Within each trial of SGP and SGPD, we run 50 times on randomly selecting subset to calculate the average MSE of each trial. We test our algorithm on five benchmark functions below:

- Function with multiple local optima-1D.

- Branin function-2D.
- Hartmann-3D.
- Hartmann-4D.
- Hartmann-6D.

For different functions, we use the number of training and test data referring to Table 1. The second column shows the number of function observations while the third column indicates the number of derivative observations $t^0$ . We use derivative observations at each dimension so that $t^0$ is equal to $d * t$, where $d$ is the number of dimensions of the input space. We set the number of inducing points $m$ as 70% of function observations ($t$) for 1D and 2D functions while setting 50% for 3D, 4D and 6D functions. We also set up an experiment to discover how our method performs if we use all function observations as inducing points ($m = t$). We summarize the MSE of each algorithm in Table 2. For all experiments, GPD has shown the best result since function observations and derivatives observations are fully used in the model. Our method SGPD closely approximates the GPD and shows a better result than SGP as well as StdGP in all regression tasks. Besides, the setting of $m = t$ with our method achieves comparable performance to GPD while requiring less com-putational complexity. For example, GPD requires 300 function observations plus 1800 derivative observations in 6D case, but SGPD ($m = t$) only incorporates 300 function observations.

Table 1. Benchmark functions vs observations for regression tasks

| | Function observations | Derivative observations | Test points |
|---|---|---|---|
| 1D | 30 | 30 | 450 |
| 2D | 200 | 400 | 800 |
| 3D | 200 | 600 | 800 |
| 4D | 300 | 1200 | 900 |
| 6D | 300 | 1800 | 900 |

Table 2. Comparison of MSE results among each method for regression tasks.

| | StdGP | SGP | GPD |
|---|---|---|---|
| 1D | 4.81e-5±1.23e-4 | 6.20e-5±1.51e-4 (0.7*t*) | **4.54e-6±1.19e-5** |
| 2D | 0.0164±0.0200 | 0.0397±0.0408 (0.7*t*) | **6.80e-4±0.0014** |
| 3D | 0.0786±0.0145 | 0.0787±0.0145 (0.5*t*) | **0.0319±0.0051** |
| 4D | 0.0453±0.0064 | 0.0473±0.0066 (0.5*t*)) | **0.0106±0.0024** |
| 6D | 0.0176±0.0038 | 0.0177±0.0026 (0.5*t*)) | **0.0053±0.0016** |

| | SGPD-Random | SGPD-Optimal | SGPD ($m = t$) |
|---|---|---|---|
| 1D | 4.68e-5±2.12e-4 (0.7*t*) | 3.92e-5±1.96e-4 (0.7*t*) | 8.00e-6±2.92e-5 |
| 2D | 0.0114±0.0145 (0.7*t*) | 0.0074±0.0068 (0.7*t*) | 0.0014±0.0022 |
| 3D | 0.0368±0.0072 (0.5*t*) | 0.0342±0.0084 (0.5*t*) | 0.0338±0.0052 |
| 4D | 0.0401±0.0079 (0.5*t*) | 0.0345±0.0073 (0.5*t*) | 0.0131±0.0030 |
| 6D | 0.0162±0.0019 (0.5*t*) | 0.0112±0.0027 (0.5*t*) | 0.0092±0.0015 |

### 5.3.2 Experiments on Bayesian optimization of benchmark functions

We apply our method on Bayesian optimization for two benchmark functions:

1. Hartmann-3D, where $\boldsymbol{x}^* = (0.1146, 0.5556, 0.8525)$ is the global minimum location with function value of $f(\boldsymbol{x}^*) = -3.8628$.
2. Hartmann-4D, where $\boldsymbol{x}^* = (0.1873, 0.1906, 0.5566, 0.2647)$ is the global mini-mum location with function value of $f(\boldsymbol{x}^*) = -3.1355$.

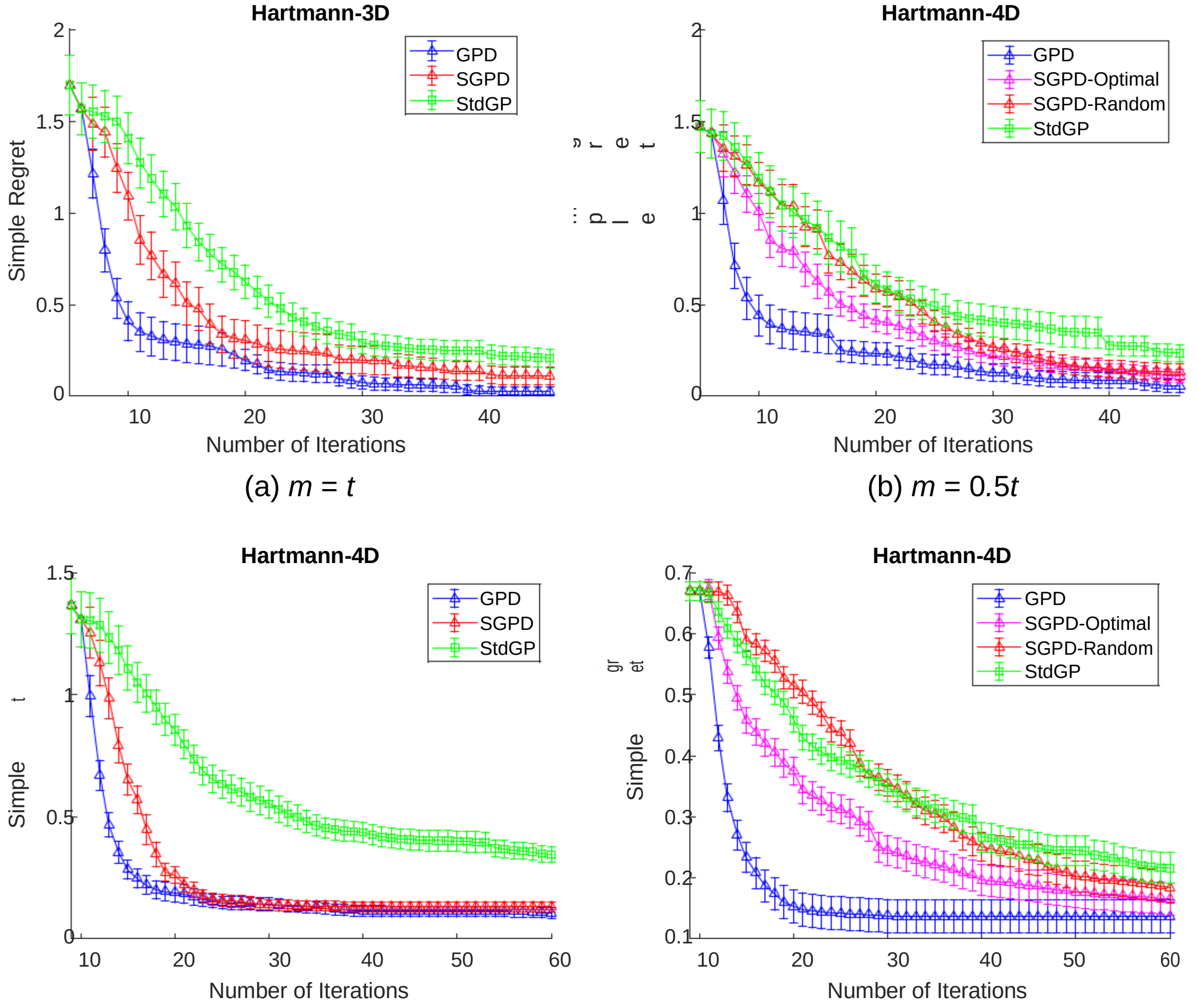

(a) $m = t$ (b) $m = 0.5t$

(c) $m = t$ (d) $m = 0.5t$

Figure 5.2: Simple regret vs iterations for (a) Hartmann-3D using 100% $t$ as inducing points (b)Hartmann-3D using 50% $t$ as inducing points (c) Hartmann-4D using 100% $t$ as inducing points(d) Hartmann-4D using 50% $t$ as inducing points.

We run each algorithm 30 trials with different initialization and report the simple regret and standard errors at the end. Simple regret is defined as $r = f(\mathbf{x}^*) - f(\mathbf{x}^+)$ where $f(\mathbf{x}^*)$ is the global optimum and $f(\mathbf{x}^+) = max_{\mathbf{x}\ \{\mathbf{x}_{1:t}\}} f(\mathbf{x})$ which is the cur-rent best value. Figure 5.2 plots the simple regret vs iterations for all experiments. BO with GPD performs the best in all three algorithms. It is easy to explain since all function observations and derivative observations have been incorporated in the algorithm. In Figure 5.2a, BO with our method SGPD using 100% function observations as inducing points outperforms BO with StdGP along the entire process. Figure 5.2b demonstrates the result of using 50% function observation as inducing points. Although our methods perform similar to StdGP at the beginning, but jump ahead after 23 iterations as more information comes into SGPD. It is emphasized that the computational cost of SGPD is less than both GPD and StdGP in this case. We also receive similar results for Hartmann-4D, and illustrate in Figure 5.2c and Figure 5.2d respectively.

### 5.3.3 Experiments on Bayesian optimization of hyperparameter tuning

We have proposed our BODMM in Chapter 4, where we can incorporate estimated derivative observations in Bayesian optimization as the real ones are not available. Here, we apply our SGPD-Optimal on BODMM for tuning hyerparameters. We experiment with three real world datasets for tuning hyperparameters of two classifiers: Support Vector Machines (SVM) and Elastic Net. In SVM we optimize two hyperparameters which are the cost parameter ($C$) and the width of the RBF kernel ($\gamma$). The search bounds for the two hyperparameters are $C = 10^{\lambda}$ where

$\lambda \in [-3, 3]$ and $\gamma = 10^{\omega}$ where $\omega \in [-5, 0]$ correspondingly. To make our search bounds manageable, we optimize for $\lambda$ and $\omega$ (i.e. we optimize in the exponent space.). In Elastic Net, the hyperparameters are the $l_1$ and $l_2$ penalty weights. The search bound for both of them is $[10^{-5}, 10^{-2}]$. We optimize in the range of expo-nents ($[-5, -2]$) and we set $m = 0.7t$ and $m = t$ respectively. For each dataset we randomly sample 70% of the data for training, and the rest as validation. All three datasets: BreastCancer, LiverDisorders and Mushrooms are publicly available from UCI data repository [Dheeru and Karra Taniskidou2017].

The results for the hyperparameter tuning of SVM and Elastic Net on the different datasets are shown in Fig 5.3 and Fig 5.4 . In the $m = 0.7t$ setting, our approach BOSGPD-Optimal performs similar with Standard BO, while performance moves closer to the BODMM in the $m = t$ setting. The latter shows that only with a frac-tion of computational cost, the sparse model has been able to provide performance similar to the full GP based BODMM model. This behavior can become more useful when BODMM in large systems becomes too big to be computationally feasible.

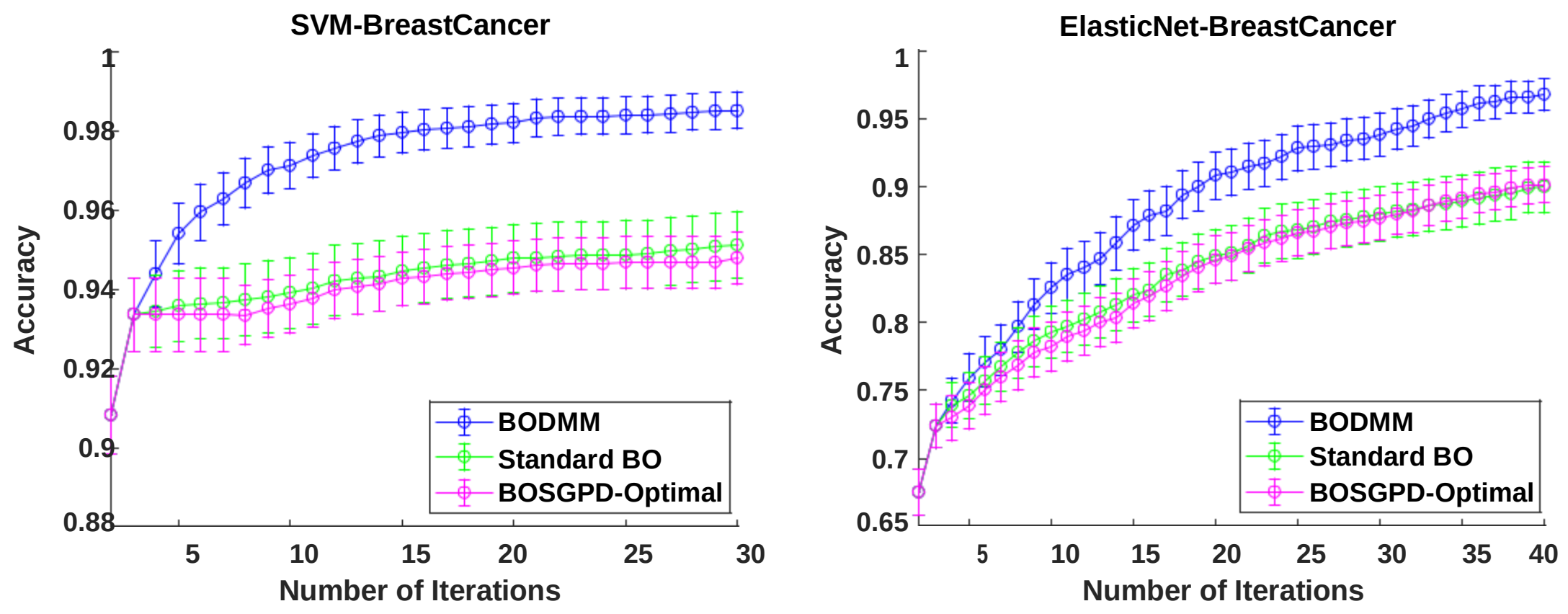

(a) BreastCancer

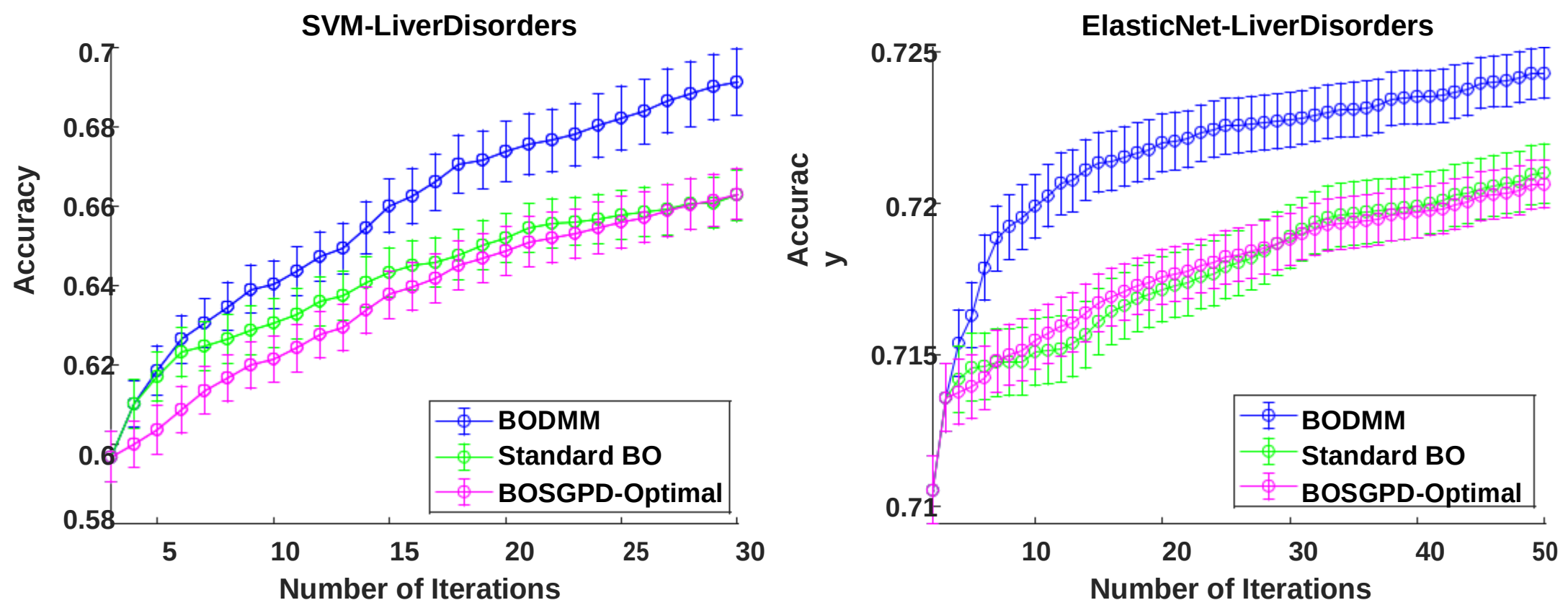

(b) LiverDisorders

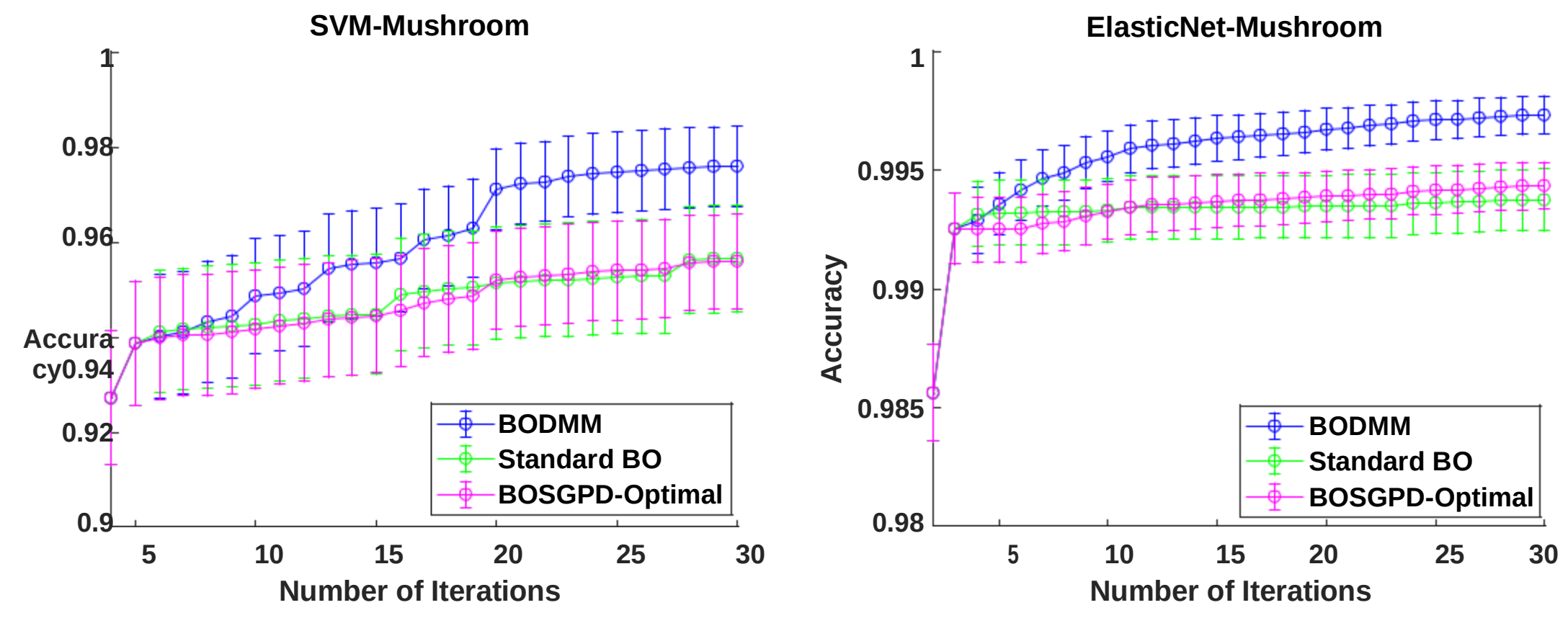

(c) Mushrooms

Figure 5.3: Accuracy vs iterations for hyperparameter tuning of SVM (left) and Elas-tic Net (right) on (a) BreastCancer, (b)LiverDisorders and (c) Mushrooms, where $m = 0.7t$.

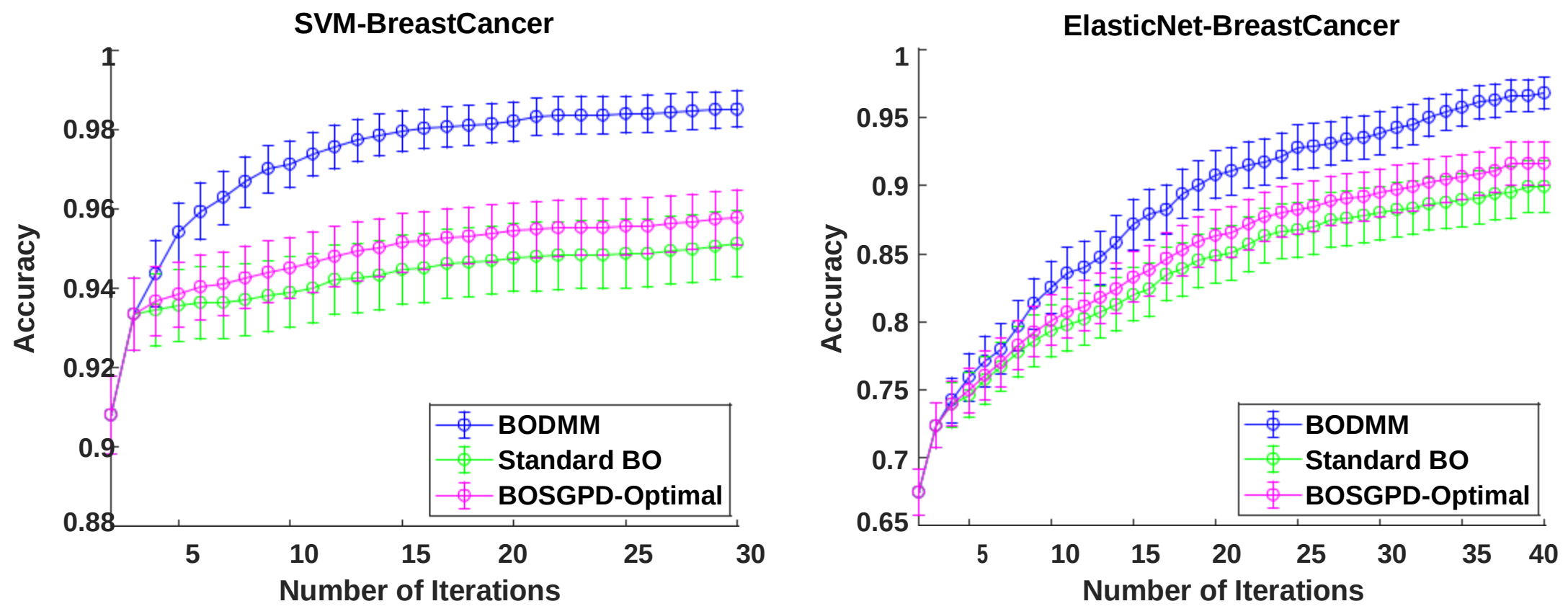

(a) BreastCancer

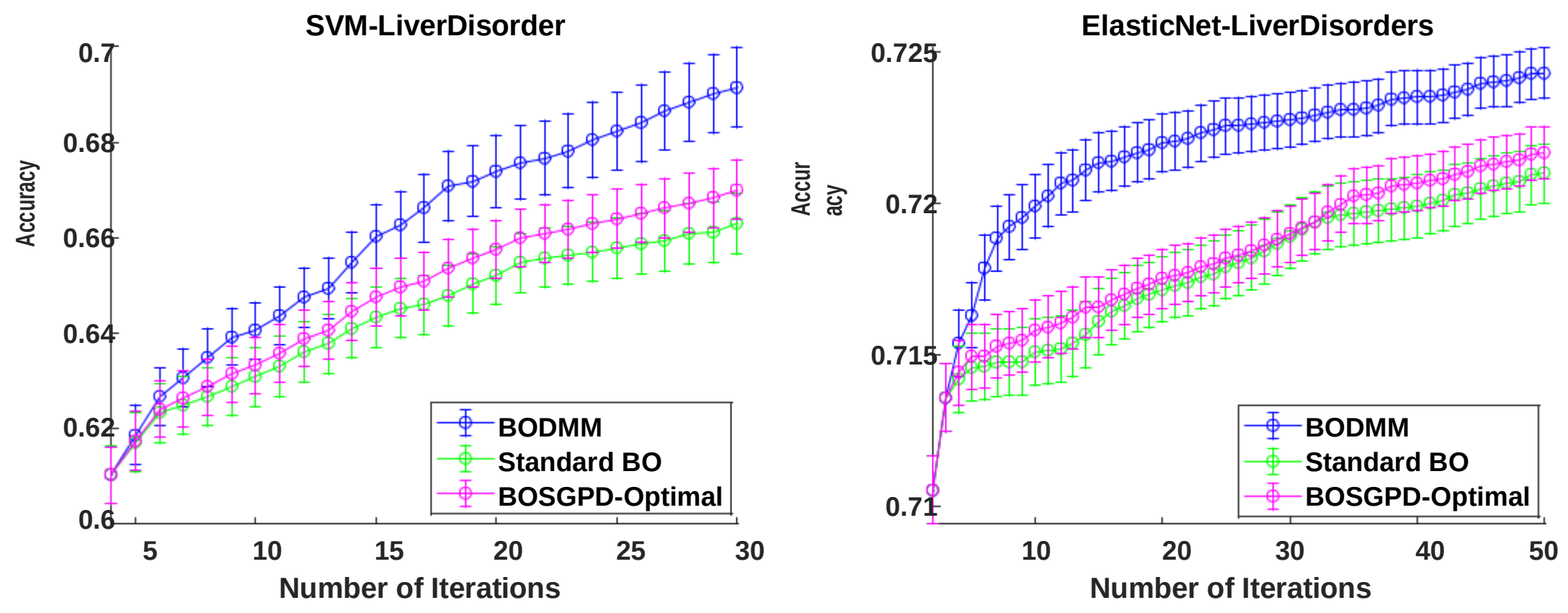

(b) LiverDisorders

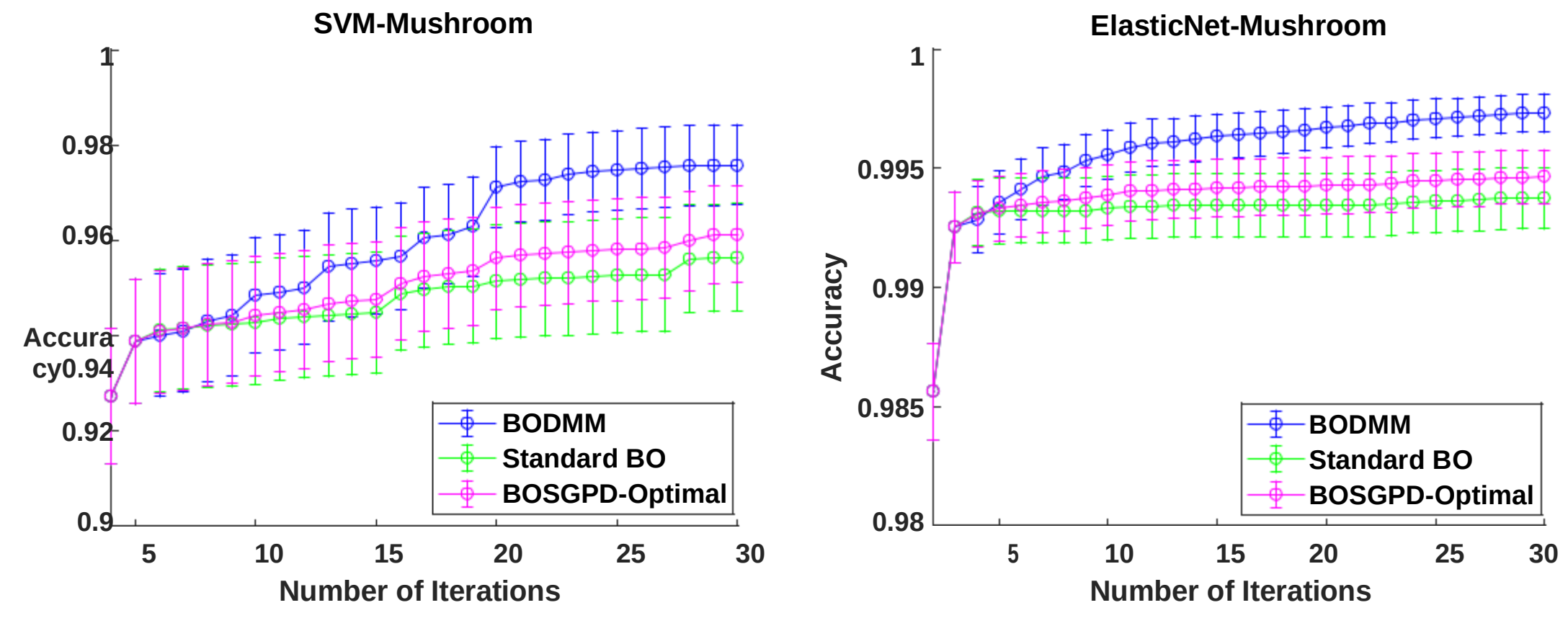

(c) Mushrooms

Figure 5.4: Accuracy vs iterations for hyperparameter tuning of SVM (left) and Elas-tic Net (right) on (a) BreastCancer, (b)LiverDisorders and (c) Mushrooms, where $m = t$.

## 5.4 Summary

In this chapter we have developed a computationally efficient approximation that takes advantages of both function observations and derivatives observations. We first demonstrate the favor of our approaches on regression tasks. Next, we investigate the usability of our method in large scale Bayesian optimization. For all experiments, our proposed approach closely approximate full Gaussian process with derivatives. It is noted that we only consider to use subset of function observations as inducing variables.

# Chapter 6

# Sparse Spectrum Gaussian Process for Bayesian Optimization

In Chapter 5 we have proposed a method to build a sparse approximation for Gaussian process in presence of large scale derivative observations. However, the sparse Gaussian process models are usually designed to target on regression tasks which fo-cus on generalization in the whole support, but these models are not always aligned with the goal of Bayesian optimization (BO) which is seeking the optimum. To be precise, existing sparse Gaussian process methods suffer from either variance under-estimation (i.e., over-confidence) [Snelson and Ghahramani2006, Lazaro Gredilla *et al.*2010] or over-estimation [Titsias2009] and thus may hamper BO as the balance between predictive mean and variance is important to the success of BO. Recently, the work of [Hensman *et al.*2017] has proposed variational Fourier features (VFF), which combines variational approximation and spectral representation of GP to-gether and plausibly can approximate both mean and variance well. However, it is difficult to extend VFF to multiple dimensional problems, since a) the number of inducing variables grows exponentially with dimensions if the Kronecker kernel is used, or b) the correlation between dimensions would be ignored if an additive kernel is used. Hence, it is important to develop a sparse GP model tailored for Bayesian optimization.

In this chapter, we propose a novel modification to the sparse spectrum Gaussian process (SSGP) [Lazaro Gredilla *et al.*2010] approach to make it more suitable for Bayesian optimization. The main intuition that drives our solution is that while being over-confident at some regions is not very critical to Bayesian optimization when those regions have both low predictive value and low predictive variance. However, being over-confident in the regions where either predictive mean or predictive vari-ance is high would be quite detrimental to Bayesian optimization. Hence, a targeted correction may be enough to make the sparse models suitable for BO. An overall measure of goodness of GP approximation for BO would be to inspect the global maximum distribution (GMD) [Hennig and Schuler2012, Hernández *et al.*2014] from the posterior GP and check its difference to that of the full GP. Alleviating over-confidence in the important regions may be sufficient to make the GMD of the sparse GP closer to that of the full GP. The based method in our work (SSGP) is notorious for under-estimate variance, which results in a less entropy of its GMD compared to that of the full GP. To correct the entropy of GMD, we add the entropy of GMD as a regularization term that is to be maximized in conjunction with the original marginal likelihood so that the derived model not only benefits for model fitting, but also fixes the over-confidence issue from the perspectives of the Bayesian optimization.

Since the GMD cannot be calculated analytically, we first provide a simple Thompson sampling approach to estimate the GMD for the sparse GP, and then propose a more efficient sequential Monte Carlo based approach. This approach provides efficiency as it offers chance to reuse samples between consecutive iterations. There-fore, many Monte Carlo samples can be reused during the optimization for the optimum frequencies as the GMD does not change drastically when frequencies are changed. Moreover, the GMD also does not change much between two consecutive iterations of Bayesian optimization as the GP differs by only one observation. Later, we empirically show that expected improvement acquisition function can be used as a proxy of the GMD with a further increase in the computational efficiency. We demonstrate our method on two synthetic functions and two real world problems one involving hyperparameter optimization in a transfer learning setting and an-other involving alloy design using a thermodynamic simulator. In all experiments our method provides superior convergence rate over the standard sparse spectrum methods. Additionally, our method also performs better than the full GP when the covariance matrix suffers from ill-conditioning due to a large number of observations placed close to each other.

We begin by outlining the details of generic sparse spectrum Gaussian process in Section 6.1 then we formulate our regularized sparse spectrum Gaussian process framework in Section 6.2. We examine Bayesian optimization convergence rate over the vanilla sparse spectrum method and over a full GP in Section 6.3. We finally conclude our contributions in Section 6.4.

## 6.1 Sparse spectrum Gaussian process

Recall from the Chapter 3 that sparse spectrum Gaussian process (SSGP) uses sparse spectrum frequencies to approximate the kernel function in spectrum domain. Briefly, any stationary covariance function can be represented as the Fourier transform of some finite measure $\sigma_f^2 p(\mathbf{s})$ with a probability density $p(\mathbf{s})$ as

$$k(\boldsymbol{x}_i, \boldsymbol{x}_j) = \int_{\mathbb{R}^D} e^{2\pi i \mathbf{s}^T (\boldsymbol{x}_i - \boldsymbol{x}_j)} \sigma_f^2 p(\mathbf{s}) d\mathbf{s}, \tag{6.1}$$

where the frequency vector $\mathbf{s}$ has the same length $D$ as the input vector $\boldsymbol{x}$. In other words, a spectral density entirely determines the properties of a stationary kernel. Furthermore, Eq.(6.1) can be computed and approximated

$$k(\boldsymbol{x}_i, \boldsymbol{x}_j) = \sigma_f^2 \mathbb{E}_{p(\mathbf{s})} \left[ e^{2\pi i \mathbf{s}^T \boldsymbol{x}_i} \left( e^{2\pi i \mathbf{s}^T \boldsymbol{x}_j} \right)^* \right] \tag{6.2}$$

$$\simeq \frac{\sigma_f^2}{m} \sum_{r=1}^{m} \cos\left[ 2\pi \mathbf{s}_r^T (\boldsymbol{x}_i - \boldsymbol{x}_j) \right] \tag{6.3}$$

$$= \frac{\sigma_f^2}{m} \varphi(\boldsymbol{x}_i)^T \varphi(\boldsymbol{x}_j). \tag{6.4}$$

The Eq.(6.4) can be obtained with the setting

$$\varphi(\boldsymbol{x}) = [\cos(2\pi \mathbf{s}_1^T \boldsymbol{x}), \sin(2\pi \mathbf{s}_1^T \boldsymbol{x}), \cdots, \cos(2\pi \mathbf{s}_m^T \boldsymbol{x}), \sin(2\pi \mathbf{s}_m^T \boldsymbol{x})]^T, \tag{6.5}$$

which is a column vector of length $2m$ containing the evaluation of the $m$ pairs of trigonometric functions at $\boldsymbol{x}$. Based on this kernel decomposition, it is straightforward to compute the posterior mean and variance of sparse spectrum Gaussian

process

$$\mu(\mathbf{x}_{t+1}) = \varphi(\mathbf{x}_{t+1})^T \mathbf{A}^{-1}\mathbf{\Phi y} \quad (6.6)$$

$$\sigma^2(\mathbf{x}_{t+1}) = \sigma_n^2 + \sigma_n^2 \varphi(\mathbf{x}_{t+1})^T \mathbf{A}^{-1}\varphi(\mathbf{x}_{t+1}), \quad (6.7)$$

where $\mathbf{\Phi} = [\varphi(\mathbf{x}_1), \ldots, \varphi(\mathbf{x}_t)] \in \mathbb{R}^{2m\times t}$ and $\mathbf{A} = \mathbf{\Phi\Phi}^T + \frac{m\sigma_n^2}{\sigma_f^2}\mathbf{I}_{2m}$. We maximize the log marginal likelihood $\log p(\mathbf{Y} | \Theta)$ to select optimal frequencies

$$\log p(\mathbf{Y} | \Theta) = -\frac{1}{2\sigma_n^2}[\mathbf{y}^T \mathbf{y} - \mathbf{y}^T \mathbf{\Phi}^T \mathbf{A}^{-1}\mathbf{\Phi y}] - \frac{1}{2}\log|\mathbf{A}| + m\log\frac{m\sigma_n^2}{\sigma_f^2} - \frac{t}{2}\log 2\pi\sigma_n^2, \quad (6.8)$$

where $\Theta$ is the set of all hyperparameters in the kernel function and the frequencies. By using $m$ optimal frequencies to approximate the full GP, SSGP holds the computational complexity $O(tm^2)$, and provides computational efficiency if $m \ll t$.

## 6.2 Bayesian optimization using regularized sparse spectrum Gaussian process

The naive SSGP can be directly used for Bayesian optimization by replacing the full GP. However, the SSGP was designed for a regression task, which means that it assigns modeling resources equally over the whole support space. Moreover, it leads to overconfidence on the GMD of interest in BO. We illustrate the overconfidence of the SSGP in Figure 6.1a-6.1b, where the GMD of SSGP in 6.1b (bottom) is narrower and sharper than the GMD of the full GP in the 6.1a.

To overcome this overconfidence, we propose a novel and scalable sparse spectrum Gaussian process model tailoring for BO. Our approach involves maximizing a new loss function to select the optimal spectrum frequencies. We design the loss function to include the marginal likelihood in the SSGP and a regularization term, which has the goal of minimizing the difference between the GMD of the full GP and that of the proposed sparse approximation. We denote our proposed sparse spectrum model as the regularized SSGP (RSSGP). For the sake of convenience, we denote the GMD of the full GP as $p(\boldsymbol{x}^*)$ and that of RSSGP as $q(\boldsymbol{x}^*)$.

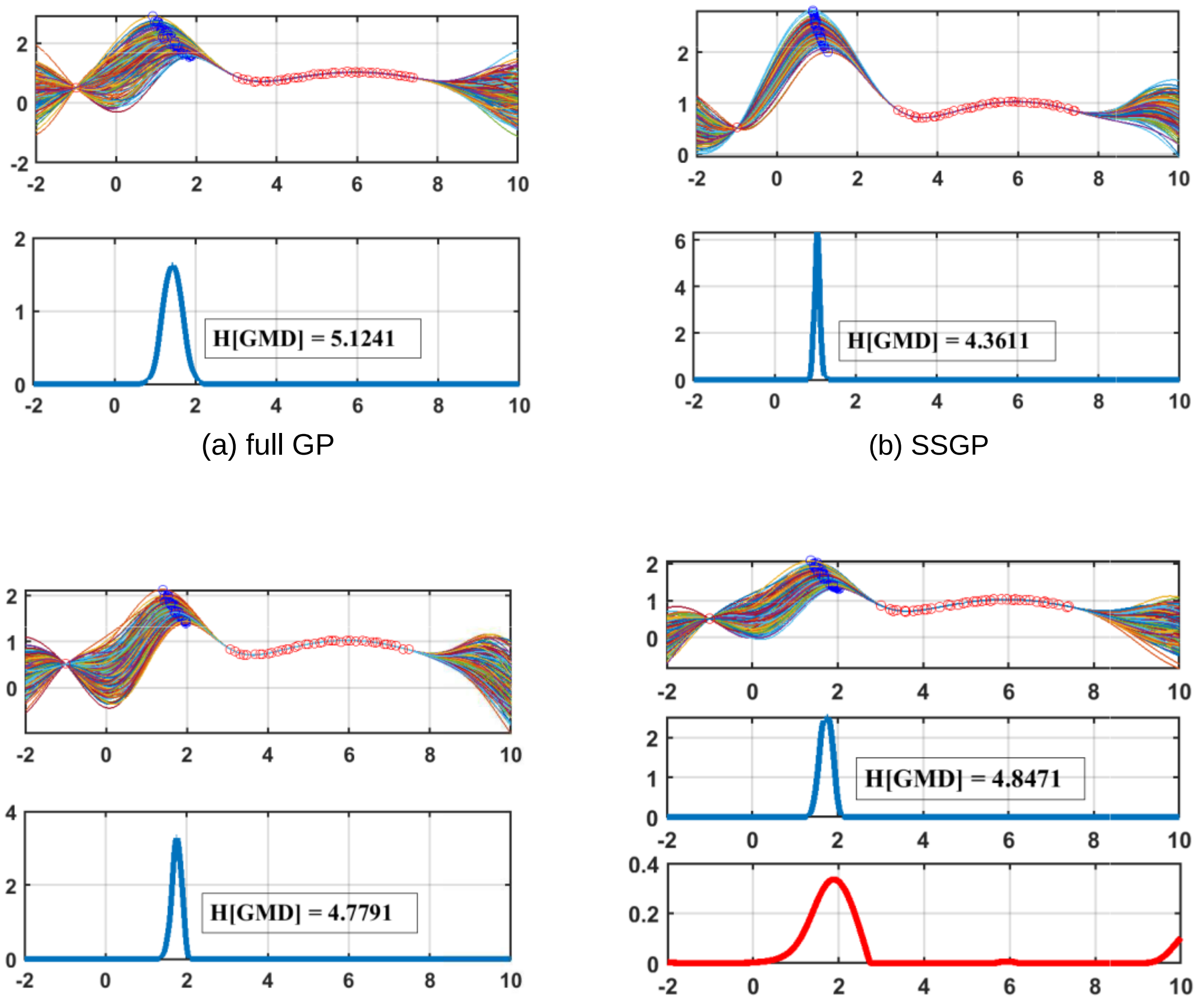

(a) full GP (b) SSGP (c) RSSGP (d) RSSGP using EI as a proxy

Figure 6.1: (a)-(c) The visualization of overconfidence of SSGP on the GMD. The *upper* graphs show 200 posterior samples of Sinc function, modeled by (a) full GP, (b) SSGP with 30 optimal frequencies, and (c) RSSGP with 30 optimal frequencies. The red circle denotes observation and the blue circle denotes the maximum location of a posterior sample. The *lower* graphs illustrate the resultant GMD respectively. The H [GMD] is the entropy of the GMD. We can see the GMD of RSSGP is closer to that of full GP than SSGP. (d) RSSGP with 30 optimal frequencies by using the EI function as a proxy to the regularization. Its GMD is at the middle and the EI function is at the bottom.

We first discuss the choice for the regularization term. Whilst the KL divergence $D_{KL}(q \,||\, p)$ seems to be the solution to measure difference between two distributions, it is not feasible in our scenario as $p(\boldsymbol{x}^*)$ is not accessible. Nevertheless, the property that the SSGP tends to over-fit [Lazaro Gredilla *et al.*2010] implies that the entropy for the GMD in SSGP would be smaller than that of the full GP. Therefore, we can use the entropy of $q(\boldsymbol{x}^*)$, or $\mathrm{H}[q(\boldsymbol{x}^*)]$ as the regularization term in the loss function that needs to be maximized. In this way, the resultant sparse GP would minimize the difference between $q(\boldsymbol{x}^*)$ and $p(\boldsymbol{x}^*)$. Formally, the loss function in RSSGP is defined as

$$\mathrm{L} = \log p(\boldsymbol{y}|\Theta) + \lambda \log \mathrm{H}\left[q(\boldsymbol{x}^*)\right], \tag{6.9}$$

where the first term is the log marginal likelihood of the naive SSGP as Eq.(6.8), the second term is the entropy for the posterior distribution of the global maximizer $q(\boldsymbol{x}^*)$ and $\lambda$ is the trade-off parameter. Now we can obtain $\Theta$ by maximizing the loss function

$$\Theta = \operatorname{argmax} \log p(\boldsymbol{y}|\Theta) + \lambda \log \mathrm{H}\left[q(\boldsymbol{x}^*)\right]. \tag{6.10}$$

The questions break down to how $q(\boldsymbol{x}^*)$ can be computed and how $q(\boldsymbol{x}^*)$ is relevant to the spectrum frequencies. Next, knowing there is no analytical form for $q(\boldsymbol{x}^*)$, we propose two methods to estimate $q(\boldsymbol{x}^*)$. One is Thompson sampling and the other is a sequential Monte Carlo approach that takes less computation. We also propose a significantly computationally-efficient approximation by using the EI acquisition function as a proxy of $q(\boldsymbol{x}^*)$.

### 6.2.1 Thompson sampling based approach

We show how to approximate $q(\boldsymbol{x}^*)$ by following the work of [Hernández *et al.*2014]. In Thompson sampling (TS), we use a linear model to approximate the function

$f(\boldsymbol{x}) = \varphi(\boldsymbol{x})^{T}\boldsymbol{\theta}$, where $\boldsymbol{\theta}$ v N $(\boldsymbol{0}, \mathbf{I})$ is a standard Gaussian. Giving observed data D $_t$, the posterior of $\boldsymbol{\theta}$ conditioning D $_t$ is a normal N $(\mathbf{A}^{-1}\boldsymbol{\Phi}^{T}\boldsymbol{y}, \sigma_n^2\mathbf{A}^{-1})$, where $\mathbf{A}$ and $\boldsymbol{\Phi}$ have already been defined in Eq.(6.6). Note that $\varphi(\boldsymbol{x})$ is a set of random Fourier features in the original TS [Hernández *et al.*2014] while it is a set of $m$ pairs of symmetric Fourier features (Eq.(6.5)) in our framework.

To estimate the GMD in RSSGP, we let $\varphi_i$ and $\boldsymbol{\theta}_i$ be a random set of $m$ pairs of features and corresponding posterior weights. Both are sampled according to the generative process above and they can be used to construct a sampled function $f_i(\boldsymbol{x}) = \varphi_i(\boldsymbol{x})^{T}\boldsymbol{\theta}_i$. We can maximize this function to obtain a sample $\boldsymbol{x}_i$. Once we have acquired sufficient samples, we use histogram based method to obtain the prob-ability mass function (PMF) over all $\boldsymbol{x}^*$, denoted as $F(\boldsymbol{x}^*)$. Then we estimate the entropy via $H[q(\boldsymbol{x}^*)] = -\sum_{i=1}^{L} F(\boldsymbol{x}^*_i) \log F(\boldsymbol{x}^*_i)$, where $L$ is the number of samples. Since our RSSGP uses Fourier features $\varphi(\boldsymbol{x})$ to approximate a stationary kernel function, and $q(\boldsymbol{x}^*)$ also changes with applying different Fourier features, therefore we can obtain the optimal features by maximizing the combined term L in Eq.(6.9). As a result, the selected optimal features in RSSGP are not only take care of pos-terior mean approximation, but also maximize the entropy of $q(\boldsymbol{x}^*)$. This is the key why we choose SSGP as our base sparse method. Sparse models like FITC and VFE are not capable with this idea since we cannot relate their sparse sets to their GMDs due to insufficient research in this area.

We illustrate the GMD of RSSGP in Figure 6.1c. We can see that it is closer to the GMD of the full GP than that of SSGP. All the GMDs in Figure 6.1 are estimated via TS.

### 6.2.2 Monte Carlo approach

The estimation of $q(\boldsymbol{x}^*)$ by TS often requires thousands of samples (e.g, $L$), each one involving the inversion of a $m \times m$ matrix. Inspired by a recent work [Bijl *et al.* 2016] employing sequential Monte Carlo algorithm to approximate the GMD, we develop an intensive approach to estimate $q(\boldsymbol{x}^*)$ in RSSGP with significantly less computation.

We start with $n_p$ particles at positions $\bar{\boldsymbol{x}}^1, \ldots, \bar{\boldsymbol{x}}^{n_p}$. Then we assign each particle a corresponding weight $\omega_1, \ldots, \omega_{n_p}$. Ultimately, these particles are supposed to converge to the maximum distribution. At each iteration, we can approximate the $q(\boldsymbol{x}^*)$ through kernel density estimation

$$q(\boldsymbol{x}^* = \boldsymbol{x}) \approx \frac{\sum_{i=1}^{n_p} \omega_i k(\boldsymbol{x}, \bar{\boldsymbol{x}}^i)}{\sum_{i=1}^{n_p} \omega_i}, \tag{6.11}$$

where $k(\boldsymbol{x}, \bar{\boldsymbol{x}}^i)$ is the approximated covariance function using $m$ features as in Eq.(6.4).

All the particles are sampled from the flat density distribution $v(\boldsymbol{x}) = \beta$ at the beginning, so that they are randomly distributed across the input space and the constant $\beta$ is nonzero. To obtain the maximum position, we will challenge existing particles. We first sample a number of $n_C$ challenger particles from a proposal dis-tribution $v^0(\boldsymbol{x})$ and denote them as $\bar{\boldsymbol{x}}_{C_1}, \ldots, \bar{\boldsymbol{x}}_{C_{n_c}}$. To challenge an existing particle e.g. $\bar{\boldsymbol{x}}^i$, we need to set up the joint distribution over $\bar{\boldsymbol{x}}^i$ and all challenger particles as

$$\begin{bmatrix} f(\bar{\boldsymbol{x}}^i) \\ f(\bar{\boldsymbol{x}}^i_{C_1}) \\ \vdots \\ f(\bar{\boldsymbol{x}}^i_{C_{n_c}}) \end{bmatrix} \sim \mathcal{N}\left( \begin{bmatrix} \mu(\bar{\boldsymbol{x}}^i) \\ \mu(\bar{\boldsymbol{x}}^i_{C_1}) \\ \vdots \\ \mu(\bar{\boldsymbol{x}}^i_{C_{n_c}}) \end{bmatrix}, \begin{bmatrix} cov(\bar{\boldsymbol{x}}^i, \bar{\boldsymbol{x}}^i) & cov(\bar{\boldsymbol{x}}^i, \bar{\boldsymbol{x}}^i_{C_1}) & \ldots & cov(\bar{\boldsymbol{x}}^i, \bar{\boldsymbol{x}}^i_{C_{n_c}}) \\ cov(\bar{\boldsymbol{x}}^i_{C_1}, \bar{\boldsymbol{x}}^i) & cov(\bar{\boldsymbol{x}}^i_{C_1}, \bar{\boldsymbol{x}}^i_{C_1}) & \ldots & cov(\bar{\boldsymbol{x}}^i_{C_1}, \bar{\boldsymbol{x}}^i_{C_{n_c}}) \\ \vdots & \vdots & \ddots & \vdots \\ cov(\bar{\boldsymbol{x}}^i_{C_{n_c}}, \bar{\boldsymbol{x}}^i) & cov(\bar{\boldsymbol{x}}^i_{C_{n_c}}, \bar{\boldsymbol{x}}^i_{C_1}) & \ldots & cov(\bar{\boldsymbol{x}}^i_{C_{n_c}}, \bar{\boldsymbol{x}}^i_{C_{n_c}}) \end{bmatrix} \right) \quad (6.12)$$

which is a multivariate Gaussian distribution [Bijl *et al.* 2016]. We can subsequently generate a sample $[\bar{f}_i, \bar{f}_{C_1}, \ldots, \bar{f}_{C_{n_c}}]^T$ from the joint distribution. We then find the maximum value in this sample. If the maximum value is great than $\bar{f}_i$, we replace $\bar{\boldsymbol{x}}^i$ with the corresponding challenger particle. Otherwise, we do nothing.

The challenger particle has an associated weight, which is often set as the ratio of the initial distribution over the proposal distribution. To speed up converge, we use the proposal distribution that is a mixture of the initial distribution and the current particle distribution. The proposal distribution $v^0(\boldsymbol{x})$ has the form as

$$v^0(\boldsymbol{x}) = (1 - \alpha)v(\boldsymbol{x}) + \alpha q(\boldsymbol{x}^* = \boldsymbol{x}) \quad (6.13)$$

where $q(\boldsymbol{x}^* = \boldsymbol{x})$ is estimated through Eq.(6.11) and $\alpha$ is a trade-off parameter (e.g., 0.5 in our experiments). To generate a challenger particle $\bar{\boldsymbol{x}}^i_{C_1}$, we first select one of the existing particles (e.g. $\bar{\boldsymbol{x}}^i$) according to the particle weights. Based on Eq.(6.13), we then can sample $\bar{\boldsymbol{x}}^i_{C_1}$ from $k(\boldsymbol{x}, \bar{\boldsymbol{x}}^i)$ with the probability $\alpha$ or from the flat density distribution $v(\boldsymbol{x})$ with the probability $1 - \alpha$. Hence, the challenger particle has a

weight as

$$\omega^i_{C_j} = \frac{v(\mathbf{x}^{-i}_{C_j})}{\alpha k(\mathbf{x}^{-i}_{C_j}, \mathbf{x}^{-k}) + (1 - \alpha) v(\mathbf{x}^{-i}_{C_j})} \quad . \tag{6.14}$$

Based on this information, we will challenge every particle once. After each round, the systematic re-sampling [Kitagawa1996] will be employed to make sure that all particles have the same weight for the next round. This process stops till sufficient rounds (In our experiments, we find that in most cases 10 rounds will be sufficient for the particles to converge to optimal ones). Thereafter, we calculate the PMF of the particles and then estimate its entropy.

The Monte Carlo (MC) approach does not require a large matrix inversion or non-linear function optimization for the purpose of $q(\mathbf{x}^*)$ approximation. Moreover, during the optimization process, $q(\mathbf{x}^*)$ does not vary a lot with the change of $\Theta$. Therefore, most of the particles can be reused in the process, significantly reducing computation cost.

We demonstrate the superiority in Figure 6.2. We denote the GMD estimated from 50,000 TS samples of a full GP posterior on a 1$d$ function as our reference $p(\mathbf{x}^*)$, showing as blue lines in Figure 6.2a and Figure 6.2b. We give the same running time (0.5$s$) to TS and MC approaches to approximate the reference $p(\mathbf{x}^*)$ respectively, showing as red lines in Figure 6.2a and Figure 6.2b. We can see that our MC approach successfully approximate the reference $p(\mathbf{x}^*)$ while the same using TS is quite off.

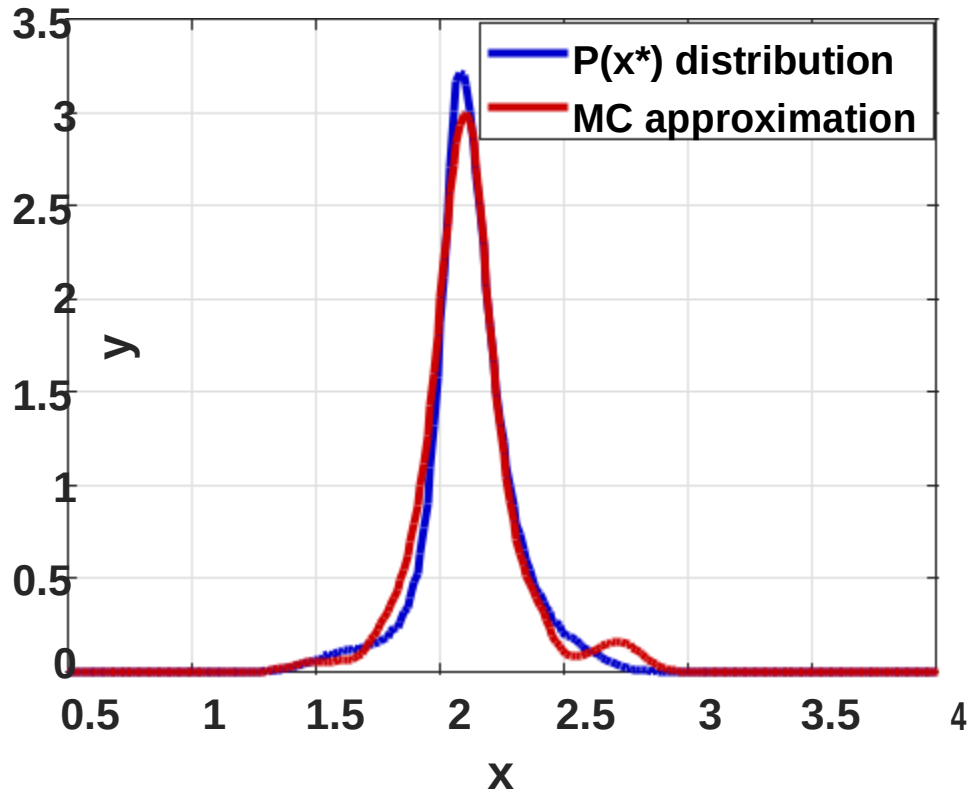

(a) MC performance

(b) TS performance

Figure 6.2: MC approach vs TS approach. The blue lines show the reference $p(\boldsymbol{x}^*)$ distribution, while the red lines illustrate the approximation using MC (a) and TS (b) given the same running time.

### 6.2.3 Expected improvement acquisition function as a proxy

To further reduce the computation, we propose to use EI function as a proxy for $q(\boldsymbol{x}^*)$. This choice is reasonable in sense that they both measure the belief about the location of the global maximum, and it can be seen from Figure 6.1d that the GMD of full GP and the EI resembles closely. We can expect that this approximation setting has a similar performance of capturing $q(\boldsymbol{x}^*)$ information as RSSGP with TS does, which is justified in Figure 6.1c-6.1d. Since EI is a function, we firstly use histogram based method to acquire the PMF of EI and then calculate the entropy. In most of the cases we find the approximation works well.

We use stochastic gradient descent to optimize Eq.(6.10). The proposed method is described in Algorithm 6.1.

**Algorithm 6.1** Regularised sparse spectrum Gaussian process for Bayesian optimization

1: **for** $n = 1, 2, ... t$ **do**
2: Optimize Eq.(6.10) to obtain hyerpararameters and optimal features,
3: Fit the data $D_t$ with RSSGP,
4: Suggest the next point $\boldsymbol{x}_{t+1}$ by maximising $\boldsymbol{x}_{t+1} = \text{argmax}\,\alpha_{EI}(\boldsymbol{x}|D_t)$,
5: Evaluate the function value $y_{t+1}$,
6: Augment the observations $D_t = D_t \cup (\boldsymbol{x}_{t+1}, y_{t+1})$.
7: **end for**

## 6.3 Experiments

In this section, we evaluate our methods on optimizing benchmark functions, an alloy design problem and hyperparameter tuning of machine learning problems using transfer learning. We compare the following probabilistic models used for Bayesian optimization:

- Full Gaussian process (**Full GP**)
- Sparse spectrum Gaussian process (**SSGP**)
- Our method 1: Regularized sparse spectrum Gaussian process with MC esti-mation for $q(\boldsymbol{x}^*)$ (**RSSGP-MC**)
- Our method 2: Regularized sparse spectrum Gaussian process with EI approx-imation for $q(\boldsymbol{x}^*)$ (**RSSGP-EI**)
- Variational Fourier features for Gaussian process using additive kernel (**VFF-AK**)
- Variational Fourier features for Gaussian process using Kronecker kernel (**VFF-KK**)

In all settings, we use EI as the acquisition function in Bayesian optimization and use the optimiser DIRECT [Finkel2003] to maximize the EI function. We include both RSSGP-MC and RSSGP-EI in synthetic experiments. We later only use RSSGP-EI due to its computational advantage and the similar performance with RSSGP-MC.

Given $d$-dimensional optimization problems and $m$ frequencies, the size of induc-ing variables would be $(2m) * d$ for VFF-AK and $(2m)^d$ for VFF-KK [Hensman *et al.*2017]. Thus, VFF-KK becomes almost prohibitively expensive for $d > 2$ and a large $m$.

### 6.3.1 Optimizing benchmark functions

We test on the following two benchmark functions:

- 2$d$ Ackley function: $f(\boldsymbol{x}^*) = 0$ and the search space is $[-10, 10]^2$;
- 6$d$ Hartmann function: $f(\boldsymbol{x}^*) = -3.32237$ and the search space is $[0, 1]^6$;

We run each method for 50 trials with different initializations and report the average simple regret along with its standard error. The simple regret is defined as $r_t = f(\boldsymbol{x}^*) - f(\boldsymbol{x}^+)$, where $f(\boldsymbol{x}^*)$ is the global maximum and $f(\boldsymbol{x}^+) = \max_{\boldsymbol{x} \{\boldsymbol{x}_{1:t}\}} f(\boldsymbol{x})$ is the best value till iteration $t$. We use the squared exponential kernel in our experiments. In terms of kernel parameters, we use the isotropic length scale, $\rho_l = 0.5, \forall l$, signal variance $\sigma_f^2 = 2$, and noise variance $\sigma_n^2 = (0.01)^2$. We empirically find that the proposed algorithms perform well when the regularization term has similar scale as that of the log marginal likelihood. Therefore, we set the trade-off parameter $\lambda = 10$ for all of our methods.

For the 2$d$ Ackley function, we start with 20 initial observations and use 20 frequencies in all sparse GP models. The experimental result is shown in Figure 6.3a. The Full GP setting performs the best, and both of our approaches (e.g., RSSGP-MC and RSSGP-EI) perform better than SSGP. RSSGP-EI performs slightly worse than RSSGP-MC since it only provides a rough approximation to the true global maximum distribution but holds simplicity. VFF-KK performs well in a low dimensional problem whilst VFF-AK performs worst. The use of additive kernel which does not capture the correlation between dimensions may result in a poor performance.

For the 6$d$ Hartmann function, we start with 150 initial observations and use 50 frequency features in all spectrum GP models. Similar results as the 2$d$ Ackley function can be seen in Figure 6.3b. We did not run VFF-KK on this case due to a huge size of inducing variables mentioned before.

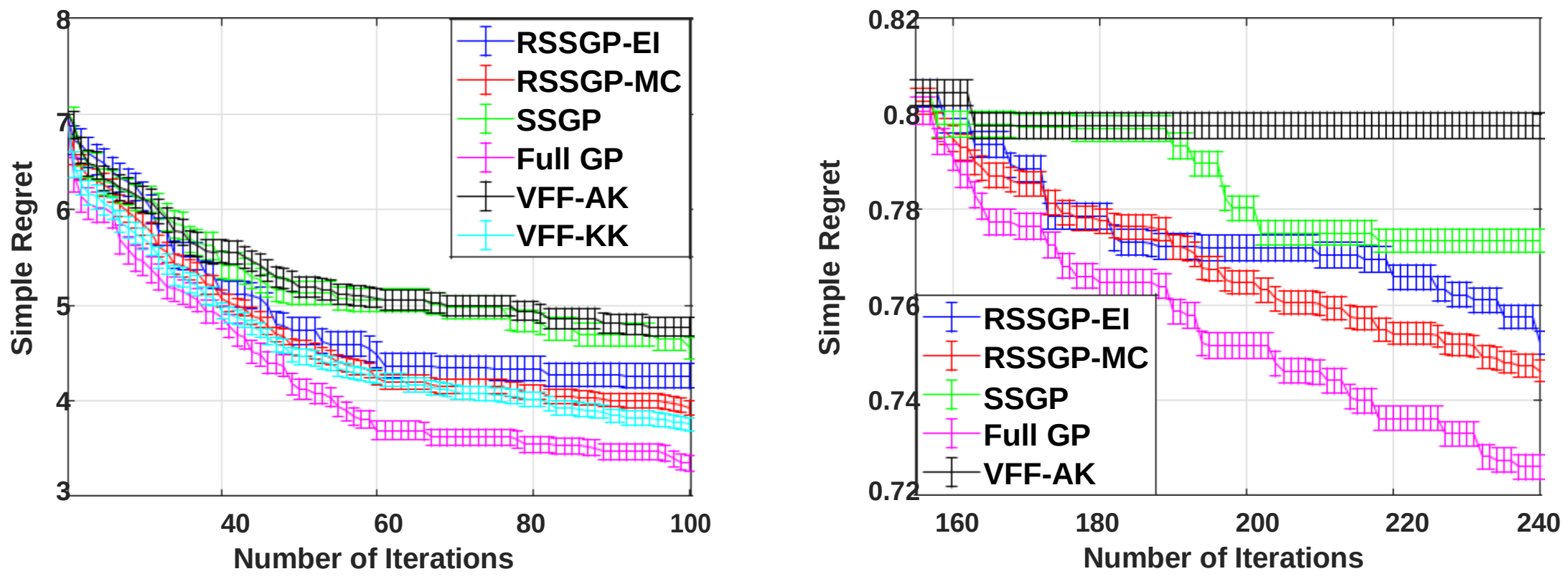

(a) 2d-Ackley (b) 6d-Hartmann

Figure 6.3: Simple regret vs iterations for the optimization of (a) 2*d* Ackley and (b) 6*d* Hartmann. The plots show the mean of minimum reached and its standard error at each iteration.

### 6.3.2 Alloy optimization

In the joint project with our metallurgist collaborators, we aim to design an alloy with a micro-structure that contains as much fraction of FCC phase as possible. FCC is a face centered cubic structure has atoms located at each of the corners and the centers of all the cubic faces. Each of the corner atoms is the corner of another cube so the corner atoms are shared among eight unit cells. We use a thermodynamic simulator called ThermoCalc [Andersson *et al.*2002]. Given a composition of an alloy, the simulator can compute thermodynamic equilibrium and predict the micro-structure of the resultant alloy using CALPHAD [Saunders and Miodownik1998] methodology. In this experiment, the search space is a 15 dimensional combination of the elements: Fe, Ni, Cr, Ti, Co, Al, Mn, Cu, Si, Nb, Mo, W, Ta, C, N. For each composition, ThermoCalc provides the amount of FCC in terms of volume fraction. The best value of volume fraction is 1. Since ThermoCalc takes around 10 minutes per composition to compute volume fraction, it fits perfectly in our notion of semi-expensive functions. We use 500 initial points and 50 frequencies and run 5 different trials with different initial points. The results in Figure 6.4 shows BO with RSSGP-EI performs the best over all three methods. We found that the covariance matrix of the full GP quickly became ill-conditioned in the presence of a large number of observations, and hence, fails to be inverted properly, being ended up harming the BO.

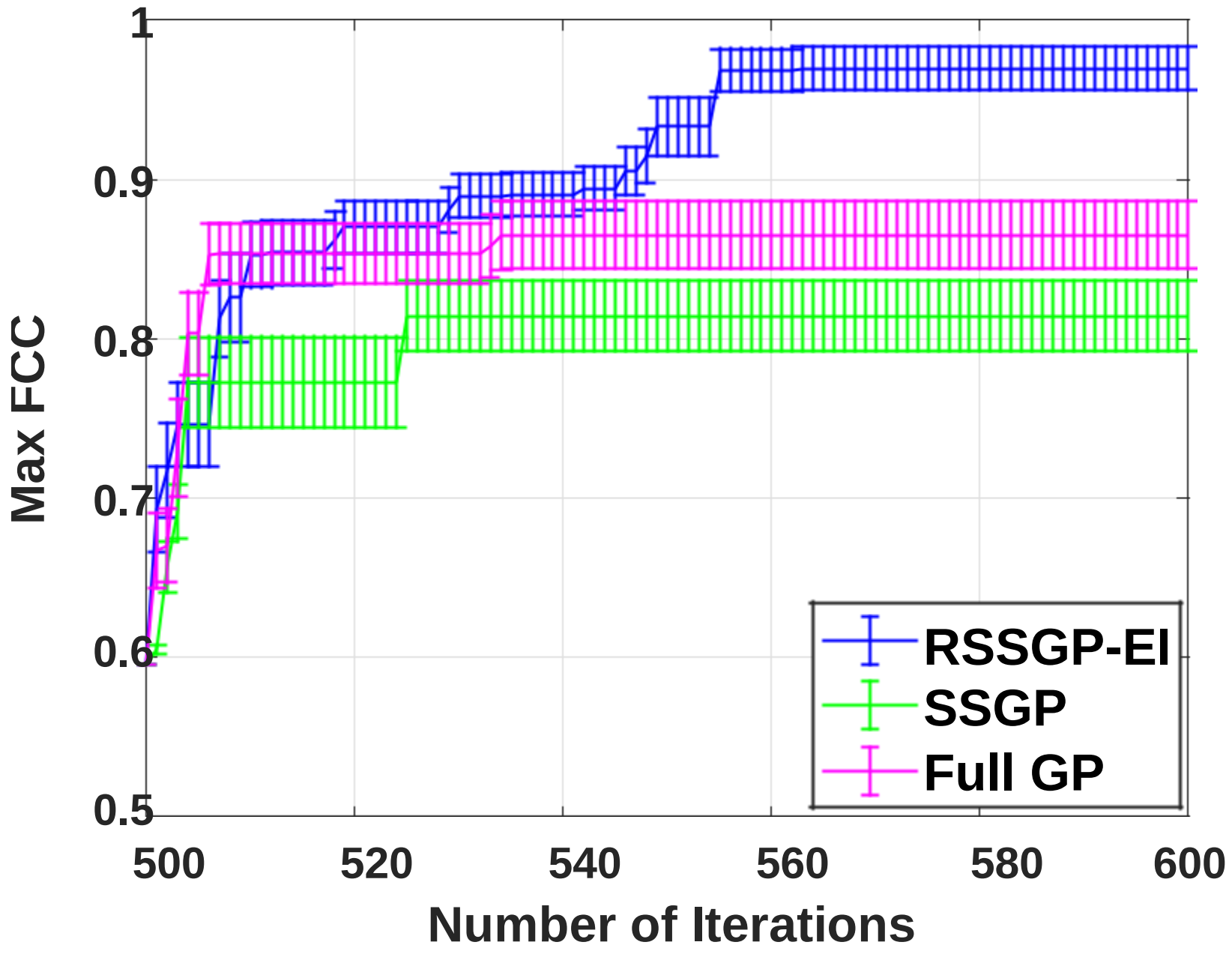

Figure 6.4: Alloy optimization-FCC at 15-dimensions.

### 6.3.3 Transfer learning on hyperparameter tuning

Transfer learning in the context of Bayesian optimization pools together observa-tions from the sources and the target to build a combined covariance matrix in the GP. In this case when the number of sources is large or/and the number of existing observations per source is large, the resultant covariance matrix can be quite huge, demanding a sparse approximation. We conduct experiments for tuning hyperpa-rameters of support vector machine (SVM) classifier in a transfer learning setting. We use the datasets: LiverDisorders, Madelon, Mushroom and BreastCancer from UCI repository [Dheeru and Karra Taniskidou2017] and construct three transfer learning scenarios. For each scenario, we use 3 out of 4 datasets as the source tasks, and the rest one as the target task. We randomly generate 900 samples of hyperpa-rameters and the corresponding accuracy from each source task. We also randomly generate 3 initial samples from the target task. As a result, we have 2703 initial observations to build the combined covariance matrix. Following the framework [Joy *et al.*2019], where the source points are considered as noisy observations for the target function, we add a higher noise variance (3 times of that in target observations) to 2700 source observations. This allows us to use the same covariance function to capture the similarity between the observations from both source and target tasks. We optimize two hyperparameters in SVM which are the cost parameter ($C$) and the width of the RBF kernel ($\gamma$). The search bounds for the two hyperparameters are $C = 10^{\lambda}$ where $\lambda \in [-3, 3]$, and $\gamma = 10^{\omega}$ with $\omega \in [-3, 0]$, respectively, and we optimize $\lambda$ and $\omega$. We run each scenario 30 trials with different initializations.

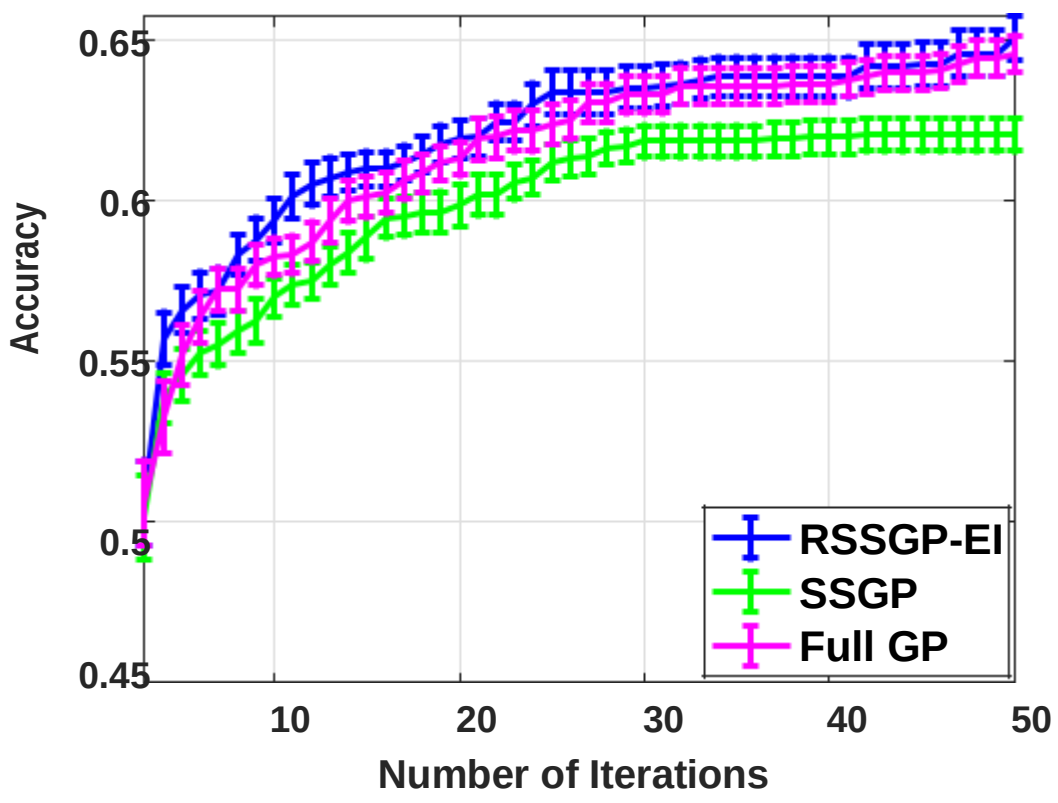

(a) LiverDisorders as Target

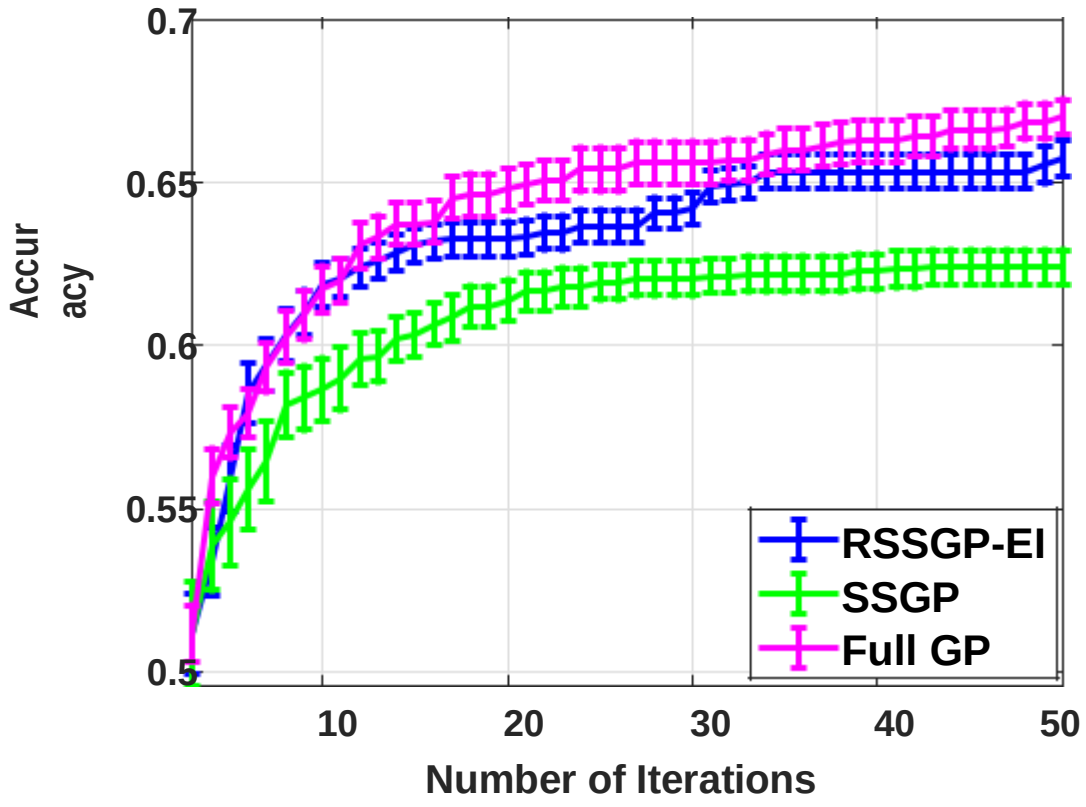

(b) Madelon as Target

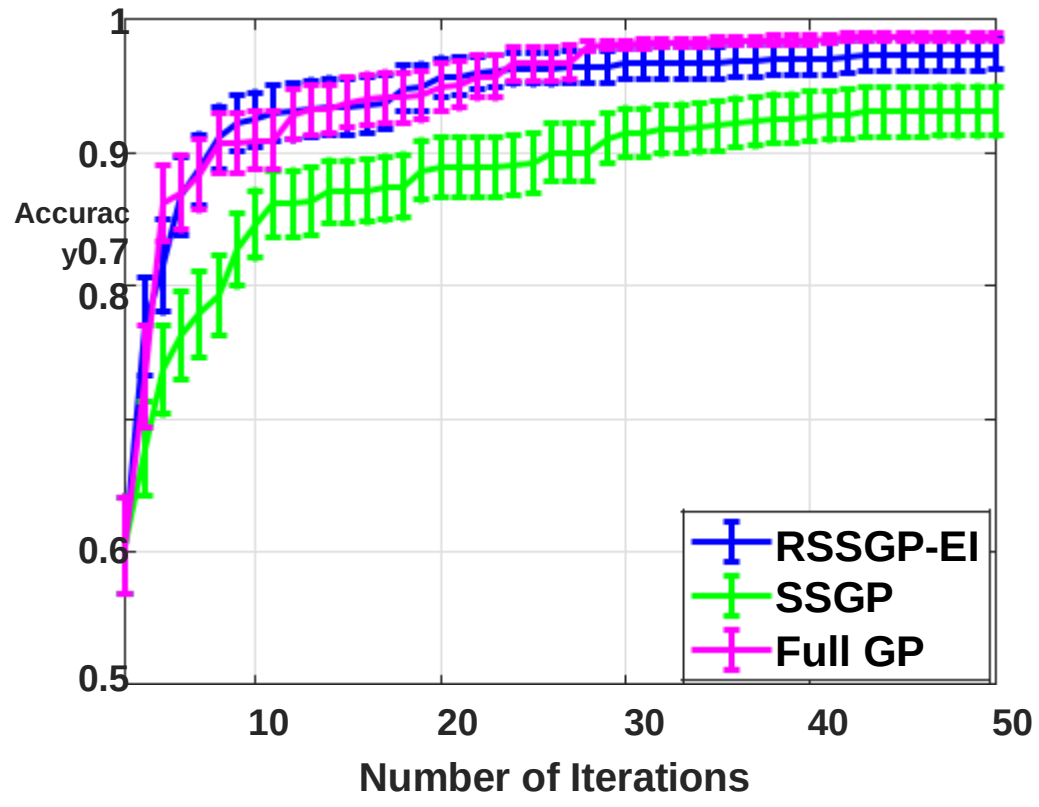

(c) Mushroom as Target

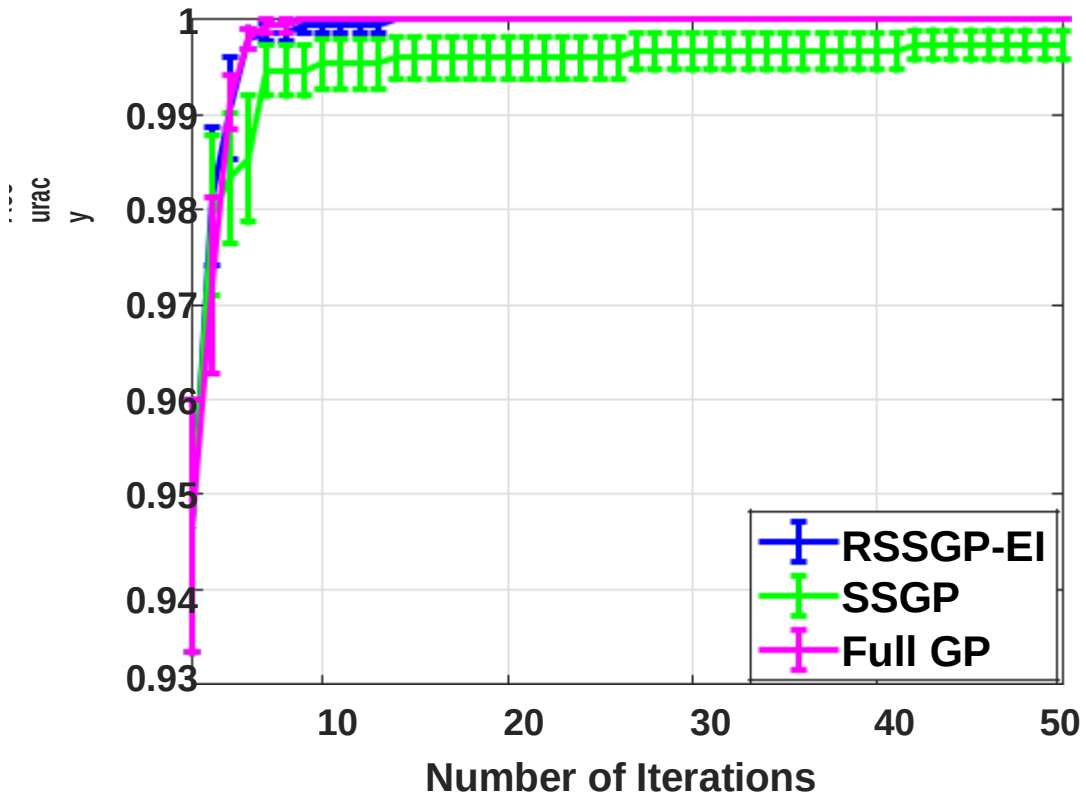

(d) BreastCancer as Target

Figure 6.5: Hyperparameters tuning for SVM classifier by transfer learning. The plots show the maximal accuracy reached till the current iteration by RSSGP-EI (blue), SSGP (green) and by FullGP (mauve). Error bar indicates the standard error.

The results are showed in Figure 6.5. We can see that in all scenarios BO with RSSGP-EI outperforms the naive SSGP. We note that the covariance matrix of full GP does not suffer from ill-conditioning since the source observations have a higher noise. Therefore, we can see the Full GP works well in this case.

## 6.4 Summary

In this chapter we propose a new regularized sparse spectrum Gaussian process method for Bayesian optimization applications. The original SSGP formulation re-sults in an overconfident GP. BO using such GP may fare poorly as the correct uncertainty prediction is crucial for the success of Bayesian optimization. We pro-pose a modification to the marginal likelihood in the original SSGP by adding the entropy of the GMD induced by the posterior GP as a regularizer. By maximizing the entropy of the GMD along with the marginal likelihood, we aim to obtain a sparse approximation which is more aligned with the goal of BO. We show that an efficient formulation can be obtained by using a sequential Monte Carlo approach to approximate the GMD. We also experimented with the expected improvement acquisition function as a proxy for the GMD. Experiments on benchmark functions and two real world problems show superiority of our approach over the vanilla SSGP method at all times and even better than the usual full GP based approach at certain scenarios.

# Chapter 7

# Conclusion

In this chapter, we summarize key contributions of this work and outline the direc-tion of future research.

## 7.1 Contributions

**An extensive review of Bayesian optimization and related works.** In Chapter 2, we present an extensive overview of Bayesian optimization, which helps us to identify open questions in Bayesian optimization. We also provide a detailed review of related works that align the research undertaken in this thesis.

**A mathematical background on Bayesian optimization components.** We explain the details of Bayesian optimization components in Chapter 3, which include the Gaussian process and various acquisition functions. We also provide mathemat-ical details of the Gaussian process with gradients and two popular sparse Gaussian process models.

**Bayesian optimization with derivative meta-model.** In Chapter 4, we propose a novel method for Bayesian optimization for well-behaved functions with small numbers of peaks. We incorporate this information through a derivative meta-model. The derivative meta-model is based on a Gaussian process with a polynomial kernel. By controlling the degree of the polynomial we control the shape of the main Gaussian process which is built using the SE kernel and the covariance matrix is computed by using both the observed function value and the derivative values sampled from the meta-model. We also provide a Bayesian way to estimate the degree of the polynomial based on a truncated geometric prior. In experiments, both on benchmark test functions and the hyperparameter tuning from popular machine learning models, our proposed model converged faster than the baselines.

**Sparse Gaussian process with derivative observations.** The computational cost in Bayesian optimization with derivative observations is significant. It is due to the inversion of the covariance matrix in the Gaussian process that requires $O((t+dt^0)^3)$, where $t$ is the number of function observations, $t^0$ is the number of deriv-ative observations and $d$ is the dimensionality of the space. Therefore, in Chapter 5, we propose an extended framework of FIC [Quinonero-Candela and Rasmussen2005] based sparse approximation to incorporate derivative observations. The novel frame-work can speed up the Gaussian process with derivatives from $O((t + dt^0)^3)$ to $O((t + dt^0)m^2)$, where $m$ is the number of inducing points and $m \leq t$. We also investigate the usability of our method in large-scale Bayesian optimization. For all experiments, our proposed approach closely approximates the full Gaussian process with derivative observations.

**Sparse spectrum Gaussian process for Bayesian optimization.** It is to be noted that 1) the classical sparse methods suffer from either variance under-estimation or over-estimation and thus may hamper Bayesian optimization as the balance between predictive mean and variance is important to the success of Bayesian optimization and 2) the state of the art method VFF [Hensman *et al.*2017], which does not seriously suffer from such misestimation, is difficult to extend to multiple dimension problems. In Chapter 6, we propose a new regularized sparse spectrum Gaussian process method to make it more suitable for Bayesian optimization applications. The original formulation results in an over-confident Gaussian process. Therefore, we propose a modification to the original marginal likelihood based estimation by adding the entropy of the global maximum distribution induced by the posterior Gaussian process as a regularizer. By maximizing the entropy of that distribution along with the marginal likelihood, we aim to obtain a sparse approximation that is more aligned with the goal of Bayesian optimization. We show that an efficient formulation can be obtained by using a sequential Monte Carlo approach to approximate the global maximum distribution. We also experimented with the expected improvement acquisition function as a proxy to the global maximum dis-tribution. Experiments on benchmark functions and two real-world problems show the superiority of our approach over the vanilla sparse spectrum Gaussian process method at all times and even better than the usual full Gaussian process based approach at certain scenarios.

## 7.2 Future work

**Inducing derivative observations.** We should point out that we only consider using a subset of function observations as inducing variables in Chapter 5, so we cannot recover full Gaussian process with derivative observations by increasing in-ducing variables. It may be more promising by considering two individual inducing sets, one for inducing function observations and the other for inducing derivative ob-servations. As a result, the model can be flexibly adapted to various circumstances. For example, one can retain the full set of function observations and use a sparse set of derivative observations (or vice versa) together to model the target functions.

**Sparse spectrum Gaussian process with derivative observations.** Although we derive a new regularized sparse spectrum Gaussian process model in Chapter 6 that tailored for Bayesian optimization, the model cannot approximate the full Gaussian process with derivative observations. Therefore, it is interesting to investigate the extension to derivative observations in sparse spectrum Gaussian process formulation and it will further advance the state of the art in Bayesian optimization.

**Learning parameters in regularized sparse spectrum Gaussian process.** In Chapter 6, we use a regularization term in the sparse spectrum Gaussian process, where a trade-off parameter $\lambda$ is empirically to control the impact of the regularizer in the model. However, it is worth understanding how the model is going to be affected by using different values of the parameter $\lambda$. Then we can optimize the $\lambda$ rather than empirically choose at a certain value. More importantly, it is also interesting to study the convergence rate of the sparse spectrum Gaussian process, so that the total number of optimal frequencies can also be learned within the model. We can expect that further investigation may result in additional, perhaps substantially improved schemes for Bayesian optimization using sparse Gaussian process models.

*The author sincerely hopes that the work in this thesis can push the research progress in large-scale Bayesian optimization, and have an impact on solving real-world problems in experimental design, machine learning and natural science.*

## Copyright Information

SPRINGER NATURE LICENSE
TERMS AND CONDITIONS

Feb 18, 2020

---

This Agreement between Ang (Leon) ("You") and Springer Nature ("Springer Nature") consists of your license details and the terms and conditions provided by Springer Nature and Copyright Clearance Center.

| | |
|---|---|
| License Number | 4771730898279 |
| License date | Feb 18, 2020 |
| Licensed Content Publisher | Springer Nature |
| Licensed Content Publication | Springer eBook |
| Licensed Content Title | Efficient Bayesian Optimisation Using Derivative Meta-model |
| Licensed Content Author | Ang Yang, Cheng Li, Santu Rana et al |
| Licensed Content Date | Jan 1, 2018 |
| Type of Use | Thesis/Dissertation |
| Requestor type | academic/university or research institute |
| Format | electronic |
| Portion | full article/chapter |
| Will you be translating? | no |
| Circulation/distribution | 1-29 |
| Author of this Springer Nature content | yes |

SPRINGER NATURE LICENSE
TERMS AND CONDITIONS

Feb 18, 2020

---

This Agreement between Ang (Leon) ("You") and Springer Nature ("Springer Nature") consists of your license details and the terms and conditions provided by Springer Nature and Copyright Clearance Center.

| | |
|---|---|
| License Number | 4771731112214 |
| License date | Feb 18, 2020 |
| Licensed Content Publisher | Springer Nature |
| Licensed Content Publication | Springer eBook |
| Licensed Content Title | Sparse Approximation for Gaussian Process with Derivative Observations |
| Licensed Content Author | Ang Yang, Cheng Li, Santu Rana et al |
| Licensed Content Date | Jan 1, 2018 |
| Type of Use | Thesis/Dissertation |
| Requestor type | academic/university or research institute |
| Format | electronic |
| Portion | full article/chapter |
| Will you be translating? | no |
| Circulation/distribution | 1-29 |
| Author of this Springer Nature content | yes |